\documentclass[journal]{IEEEtran}

\usepackage{cite}
\ifCLASSINFOpdf
  \usepackage[pdftex]{graphicx}
\else
\fi
\usepackage{amsmath}
\usepackage{amsfonts}
\usepackage{bm}
\usepackage{multirow}
\usepackage{tikz}
\usepgflibrary{arrows}
\usepackage{esvect}
\usepackage{bbm}

\usepackage{color}
\usepackage{soul}

\newcommand{\remindtext}[2]{{{#2}}}
\newcommand{\addedtext}[2]{{{#2}}}
\newcommand{\modifiedtext}[2]{{{#2}}}
\newcommand{\deletedtext}[2]{{}}

\newcommand{\newaddedtext}[2]{{{#2}}}
\newcommand{\newremindtext}[2]{{{#2}}}
\newcommand{\newmodifiedtext}[2]{{{#2}}}

\begin{document}
\title{Underactuation Design for Tendon-driven Hands via Optimization of Mechanically Realizable Manifolds in Posture and Torque Spaces}
\author{Tianjian~Chen,~\IEEEmembership{Student~Member,~IEEE,}
        Long~Wang,~\IEEEmembership{Member,~IEEE,}
		Maximilan~Haas-Heger,~\IEEEmembership{Student~Member,~IEEE,}
        and~Matei~Ciocarlie,~\IEEEmembership{Member,~IEEE}
\thanks{All authors are with the Department of Mechanical Engineering, Columbia University, New York, NY, 10027 USA.}
\thanks{E-mails: \{tianjian.chen, long.w, m.haas, matei.ciocarlie\}@columbia.edu}
\thanks{Manuscript received August 26, 2019; revised December 4, 2019.}
}

\markboth{IEEE TRANSACTIONS ON ROBOTICS}%
{Shell \MakeLowercase{\textit{Chen et al.}}: Underactuation Design for Tendon-driven Hands via Optimization of Mechanically Realizable Manifolds in Posture and Torque Spaces}

\maketitle
\begin{abstract}
Grasp synergies represent a useful idea to reduce grasping complexity without compromising versatility. Synergies describe coordination patterns between joints, either in terms of position (joint angles) or effort (joint torques). In both of these cases, a grasp synergy can be represented as a low-dimensional manifold lying in the high-dimensional joint posture or torque space. In this paper, we use the term \textit{Mechanically Realizable Manifolds} to refer to the subset of such manifolds (in either posture or torque space) that can be achieved via mechanical coupling of the joints in underactuated hands. We present a method to optimize the design parameters of an underactuated hand in order to shape the Mechanically Realizable Manifolds to fit a pre-defined set of desired grasps. Our method guarantees that the resulting synergies can be physically implemented in an underactuated hand, and will enable the resulting hand to both reach the desired grasp postures and achieve quasistatic equilibrium while loading the grasps. We demonstrate this method on three concrete design examples motivated by a real use case, and evaluate and compare their performance in practice.

\end{abstract}

\begin{IEEEkeywords}
\addedtext{1-6}{tendon-driven }underactuated hands, Mechanically Realizable Manifolds, synergies
\end{IEEEkeywords}

\section{Introduction}

\IEEEPARstart{G}{rasp} synergies refer to the correlation of multiple degrees-of-freedom (DoFs) in a hand, providing an effective way to realize versatile grasping in a simple fashion. The idea of grasp synergies originates in studies of human hands \cite{santello1998postural}, but has also been adopted in many robotic applications. For example, synergies can be used in the planning or control algorithms for robotic hands (e.g.,\cite{rosell2011autonomous, ciocarlie2009hand, wimbock2011synergy, gioioso2013mapping, meeker2018intuitive, matrone2010principal, matrone2012real, tsoli2010robot}). Synergies can also be embedded into the mechanical design of underactuated hands, moving part of ``intelligence'' to the hardware. The latter idea is the main focus of this paper.

Underactuated hands are gaining increasing attention in academia and industry. In particular, multi-fingered underactuated hands sit between two categories at opposite ends of the complexity spectrum -- parallel jaw grippers and fully-actuated dexterous hands, providing the benefits of the two sides -- simplicity and versatility. Taking advantage of synergies, multi-fingered underactuated hands can perform different grasps with only a small number of actuators and control inputs. At the same time, these hands can conform to objects by virtue of differential or breakaway mechanisms, so they do not require careful grasp synthesis and are robust to perception error.

\begin{figure}[t!]
\centering
\begin{tabular}{ccc}
    \includegraphics[width=0.25\linewidth]{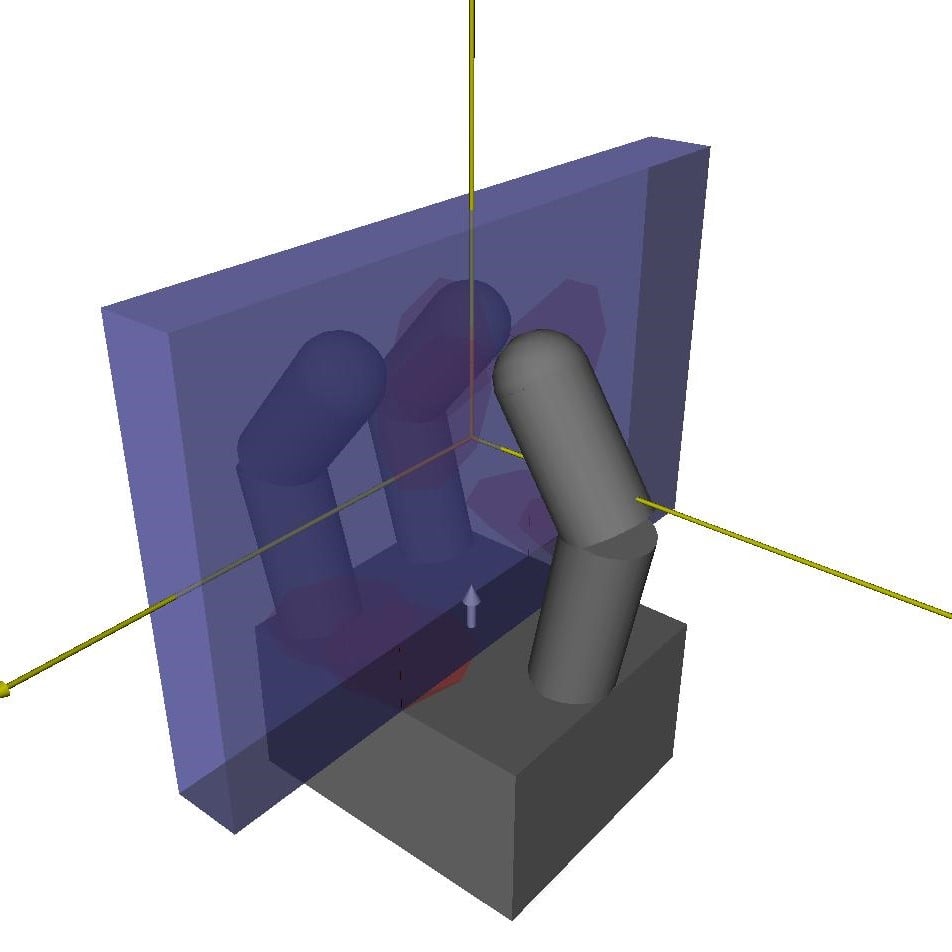}  &\hspace{-8mm} \vspace{-1mm} \includegraphics[width=0.25\linewidth]{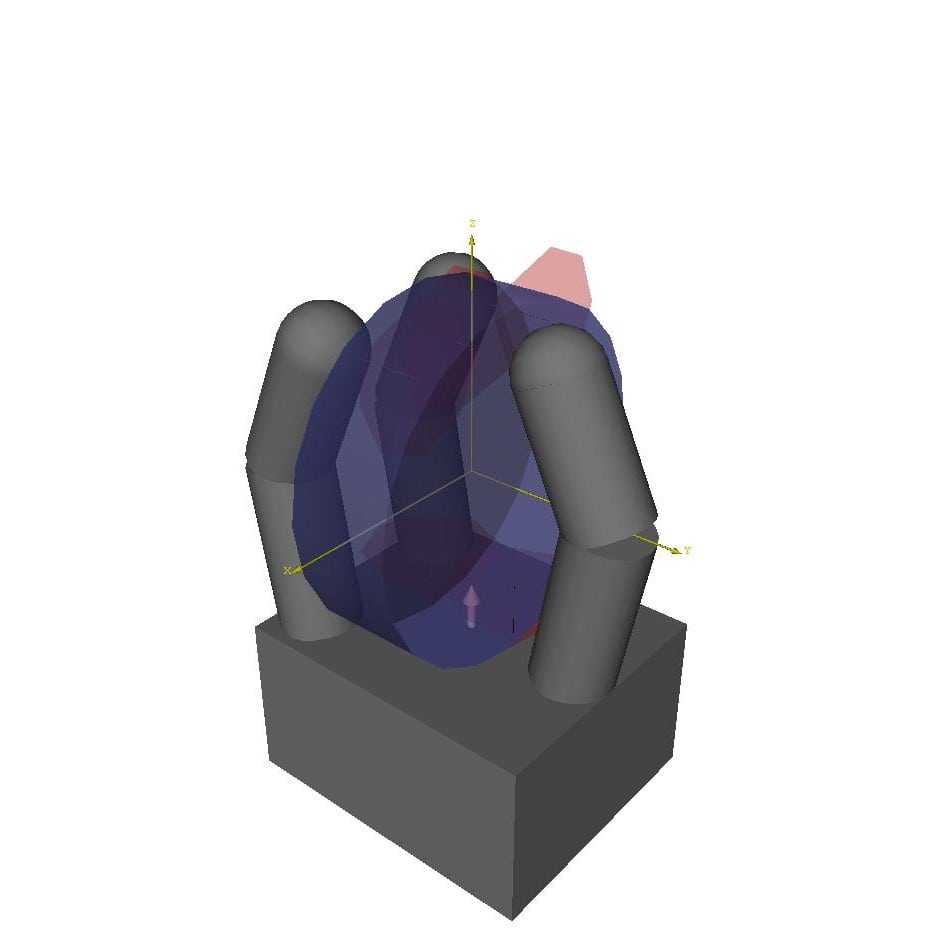} &\hspace{-8mm} \vspace{1mm} {\multirow{2}{*}[17mm]{\includegraphics[width=0.5\linewidth]{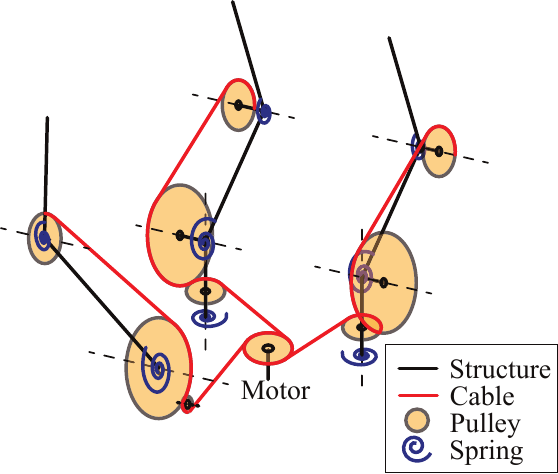}}}\\
    \includegraphics[width=0.25\linewidth]{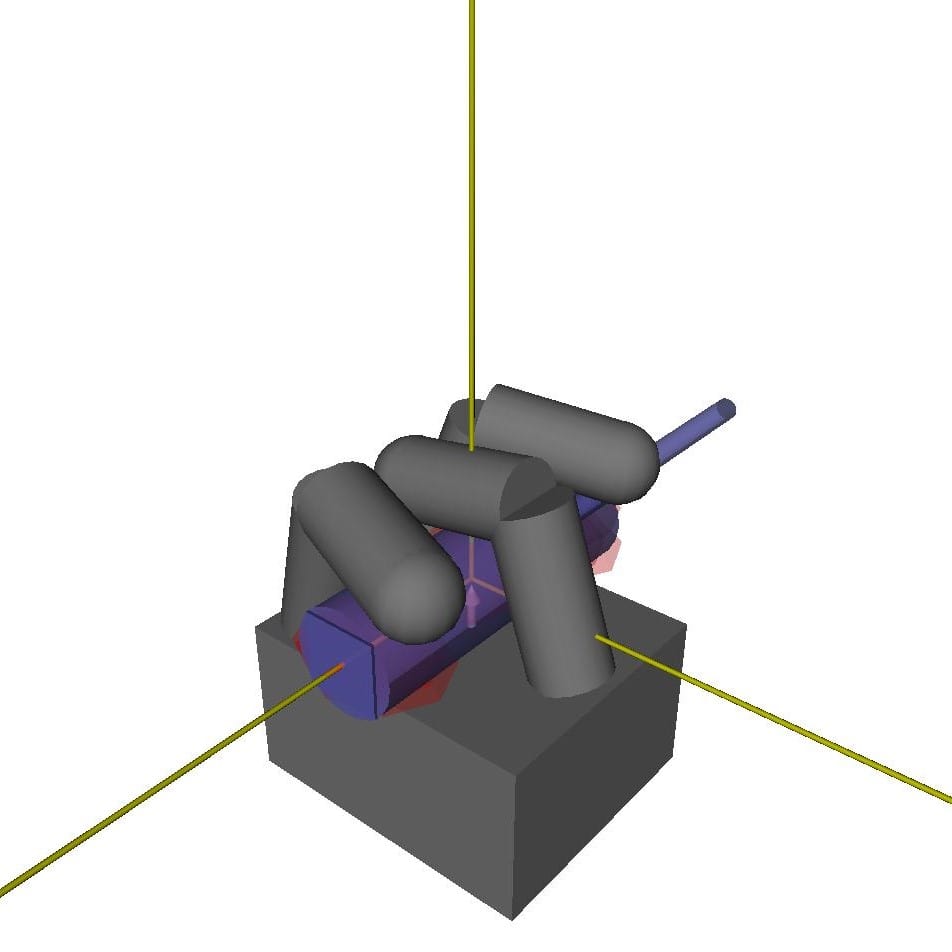}  &\hspace{-8mm} \vspace{-2mm} \includegraphics[width=0.25\linewidth]{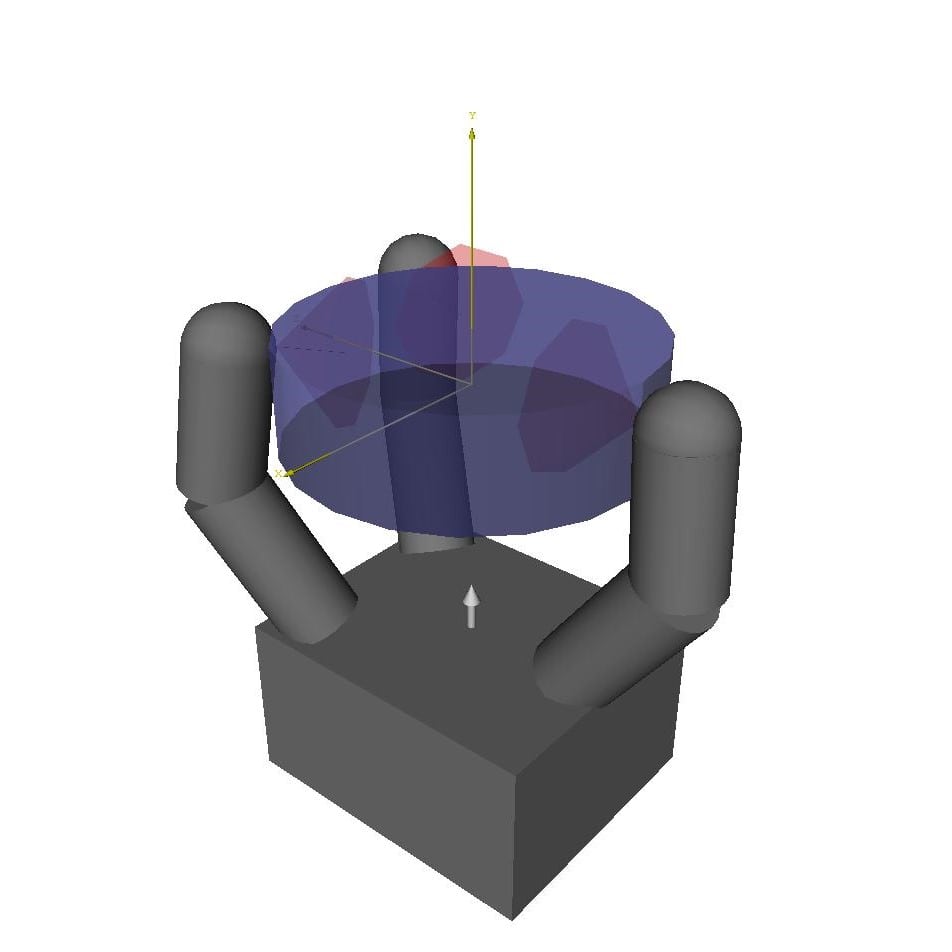} & \hspace{-8mm} \vspace{0mm} \\
    \multicolumn{2}{c}{(a)} & (b) \\
    \multicolumn{3}{c}{
        \begin{tikzpicture}[thick]
        \draw [black, line width=3mm, preaction={-triangle 60,thick,draw, line width=1mm, shorten >=-3mm}] (0, 4mm) -- (0, 0) node [right] {};
        \end{tikzpicture}}\\
\end{tabular}

\begin{tabular}{ccc}
    \hspace{-4mm}{\includegraphics[width=0.4\linewidth]{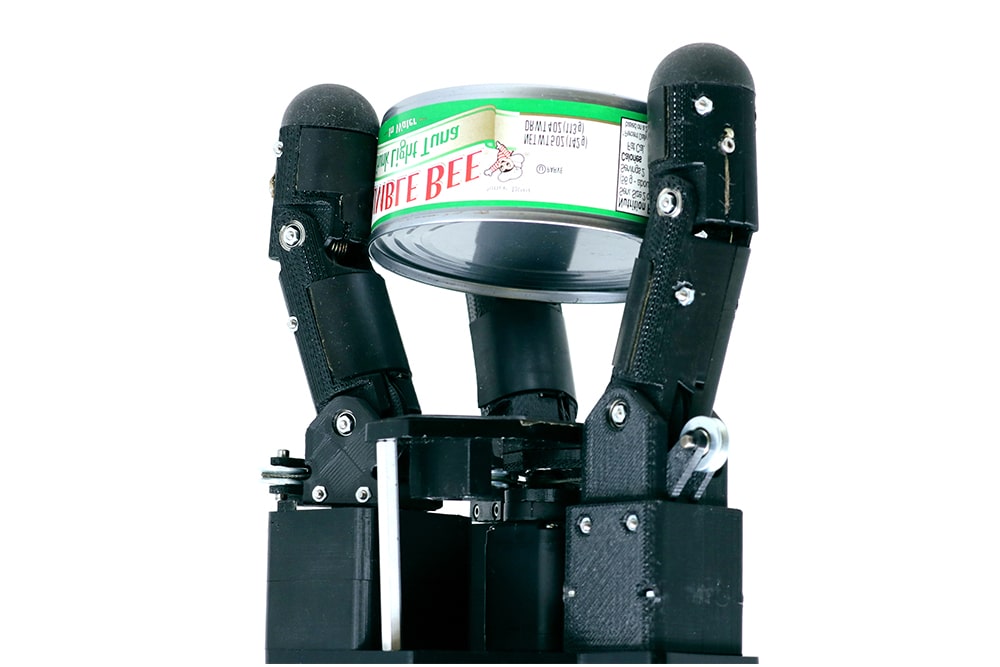}} &\hspace{-12mm} {\includegraphics[width=0.4\linewidth]{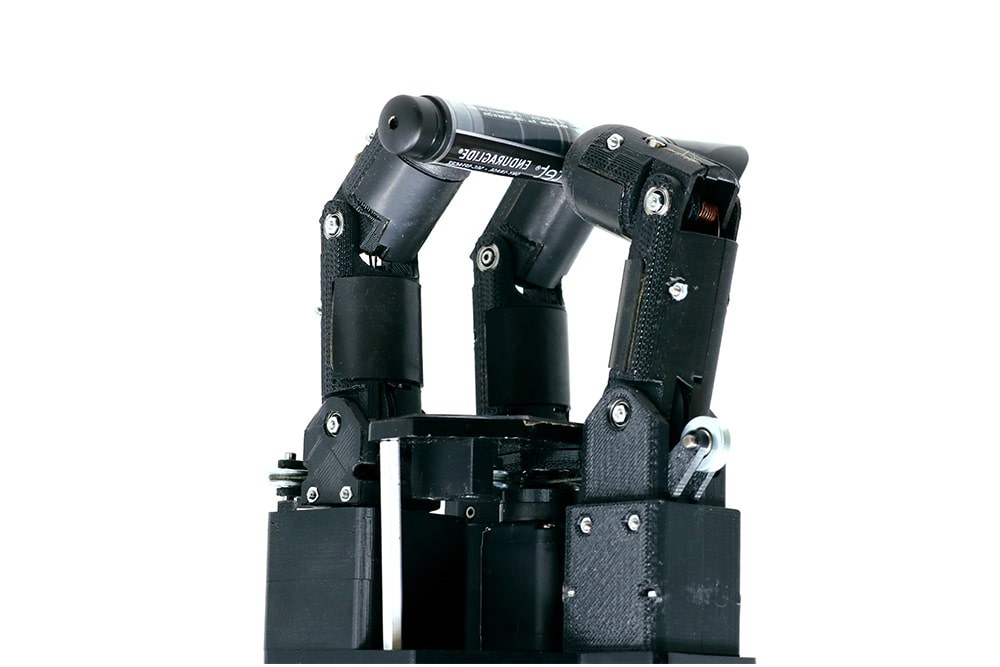}} &\hspace{-12mm} {\includegraphics[width=0.4\linewidth]{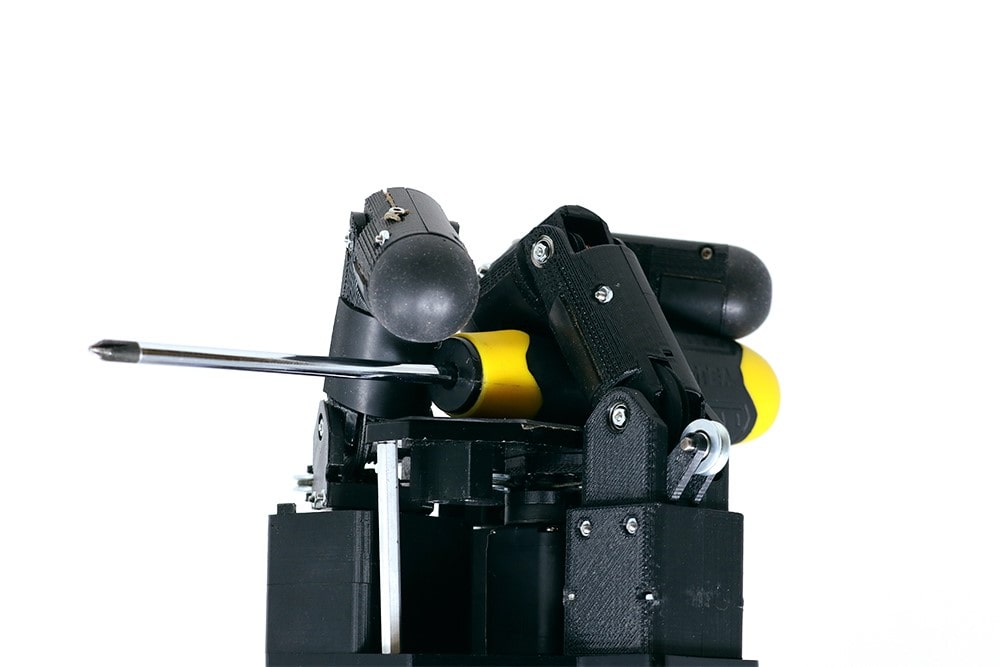}} \\
    \hspace{-4mm}(c) & \hspace{-12mm} (d) & \hspace{-12mm} (e)\\

\end{tabular}

\caption{An overview of our method. We aim to (approximately) realize a set of dexterous grasps with a highly underactuated hand (i.e. the number of actuators is much smaller than the number of joints). We start from (a) a set of simulated desired grasps without considering underactuation (assuming independent joints) and (b) an actuation scheme (desired number of motors, and tendon connectivity pattern). We propose a method that fits the Mechanically Realizable Manifolds to these grasps in both joint posture and torque spaces, ensuring that a highly underactuated hand with the given morphology and kinematics can be physically constructed to perform these desired grasps stably.}
\label{fig:cover}
\end{figure}
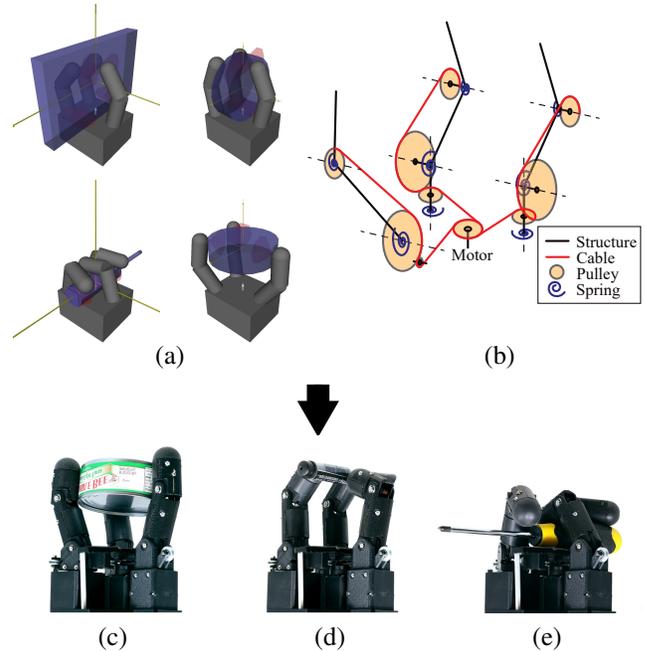

In joint posture space, a synergy of a (fully- or underactuated) hand can be represented by a low-dimensional manifold, along which a point representing a hand configuration can slide. In joint torque space, a low-dimensional torque coordination scheme also corresponds to a manifold where the torques can take values. 

We introduce the term \textit{Mechanically Realizable Manifolds} to refer to the aforementioned manifolds in either joint angle or torque space \textit{that can be physically realized by underactuated mechanisms}. These manifolds are parameterized based on mechanical specifications, and can be altered by changing values of the design parameters to exhibit different shapes corresponding to different hand behaviors.

Realizing an arbitrary manifold using mechanisms under physical constraints is a non-trivial task. One possible way is by using additional mechanical transmissions (further reviewed in the next section), which usually leads to a complex and bulky design. Another option, which is the direction we take, is to carefully design the must-have actuation mechanisms (such as pulleys and springs) to exhibit the desired behavior. However, even though certain mechanical parameters are allowed to vary, the design choices of hand structures still limit the range of manifolds that can be obtained. For example, circular pulley-tendon mechanisms always impose linear relationships between different joint torques on one tendon, and thus cannot be made to represent a nonlinear target manifold. In addition, design parameters are also limited due to real-world constraints, so the shape of the corresponding manifold can only vary in a limited range. These are examples of issues that we attempt to address in this paper. 

The key question we want to answer is the following: \textit{How can we design highly underactuated hands to perform a given set of grasps?} Concretely, we formulate our problem as follows. We start from: (i) a hand model with a known morphology and kinematic configuration (e.g., finger number and dimensions, tendon connectivity patterns, etc.) but \textit{undetermined actuation parameters} (e.g. tendon routing parameters, restoring spring parameters, etc.), and (ii) a set of desired grasps created using the aforementioned hand model but \textit{without accounting for underactuation} (assuming independent joints). Our goals are to (i) design the \textit{Mechanically Realizable Posture Manifold} to approach the desired grasp postures, and (ii) design the \textit{Mechanically Realizable Torque Manifold} to generate grasp forces as close to equilibrium as possible. This idea is illustrated in Fig. \ref{fig:cover}. 

We achieve our goal of obtaining desired Mechanically Realizable Manifolds via optimization of design parameters of the underactuation mechanisms. We propose a dual-layer framework combining a stochastic global search to select parameters in the outer layer and a convex optimization to calculate scores in the inner layer.

\remindtext{3-11}{We contrast our method against two traditional approaches for obtaining low-dimensional manifolds (or synergies)}, further reviewed in the next section. One is to obtain a manifold by simply fitting a function (e.g. linear fit through Principal Component Analysis (PCA)) to a desired set of data (e.g. target grasp postures). However, these methods \textit{have no guarantee that the resulting manifold can be implemented in an underactuated mechanism} under design constraints; thus, the corresponding synergies can often be implemented only at a software level. Furthermore, operating exclusively in posture space does not guarantee grasp stability -- a force equilibrium problem. The second traditional approach is to implicitly define a manifold by designing a hand empirically, but no method exists that can \textit{make the resulting manifold fit a specific set of desired grasps}. We aim to combine the advantages of both methods: by fitting Mechanically Realizable Manifolds to target grasps in both posture and torque spaces, we ensure that the results are realizable in practice via underactuation and suitable for a set of grasps that the resulting hand can execute in a stable fashion. 

Our main contribution in this paper can be summarized as follows: To the best of our knowledge, \textit{we are the first to propose a method to determine both (i) pre-contact postural synergies and (ii) post-contact joint torque coordination schemes, which are guaranteed to be mechanically realizable in tendon-driven underactuated hands, and which enable both pre- and post-contact equilibrium for a set of desired grasps.} We also present an evaluation of our method on three concrete hand design cases with different hand kinematics, grounded in the needs of the real use case of a free-flying assistive manipulator in the International Space Station.

\section{Related Work}

A common place to apply the idea of postural synergies for robotic hands is in the planning or control for fully-actuated dexterous hands. Various studies are presented in this category, though the term for synergies may be different, e.g. ``eigengrasps" or ``eigenpostures". Planning in the low-dimensional subspace can significantly reduce the computation complexity of the grasp search. For example, Rosell et al. \cite{rosell2011autonomous} studied the motion planning problem of a hand-arm system in a reduced-dimensional synergy space. Ciocarlie and Allen \cite{ciocarlie2009hand} discussed the use of low-dimensional postural subspace in the automated grasp synthesis, and proposed a planner which takes advantage of reduced dimensionality and can be fast enough to run in real-time. Moreover, the technique of synergy is also adopted in the (real-time) control of robotic hands, meaning the joints are commanded in a coupled fashion. For example, the work from Wimbock et al. \cite{wimbock2011synergy} showed a synergy-level impedance controller for a multi-finger hand. One specific idea in this category of robot hand control is to use low-dimensional synergies as a human-robot interface for teleoperation, learning by demonstration, and prosthetics, in order to reduce the required communication bandwidth between the human and the robot. For teleoperation, this method has been shown in the studies from Gioioso et al. \cite{gioioso2013mapping} and Meeker et al. \cite{meeker2018intuitive}. For hand prosthetics, Matrone et al. \cite{matrone2010principal} \cite{matrone2012real} as well as Tsoli and Jenkins \cite{tsoli2010robot} presented the aforementioned idea and developed working prototypes. However, these studies mostly consider postural control without accounting for grasping force equilibrium. \deletedtext {1-7}{Besides, most of these studies are limited to anthropomorphic hands in which the synergies can be extracted from human data.}

In contrast, since the methods above implement synergies in software, we can also transfer this idea by coupling joints in hardware. This leads to the mechanical realization of pre-defined synergies. For example, Brown and Asada \cite{brown2007inter} designed a mechanical implementation of PCA results for a hand using pulley-slider systems to realize inter-finger coordination. Xu et al. \cite{xu2014design} \cite{xu2019composed}, Li et al. \cite{li2014design}, Chen et al. \cite{chen2015mechanical}, Xiong et al. \cite{xiong2016design} also proposed several studies to enable hardware synergies for anthropomorphic hands, based on different types of mechanisms such as gears, continuum mechanisms, cams, and sliders. These studies only consider postural behaviors without the notion of force, so they do not have a guarantee for grasp stability. To deal with this issue, a series of works from Gabiccini et al. \cite{gabiccini2011role}, Prattichizzo et al. \cite{prattichizzo2013motion}, Grioli et al. \cite{grioli2012adaptive}, and Catalano et al. \cite{catalano2014adaptive}, presented the concept of ``soft synergies" and ``adaptive synergies" and provided the models and tools to account for force generation and force equilibrium. They also used this theory in the design and control of a multi-finger dexterous hand (the Pisa/IIT hand). Though all studies above present feasible ways to implement synergies mechanically, they are limited to anthropomorphic hands for which the synergies can be extracted from human data, whereas we try to discover synergies for a broader range of hands (both anthropomorphic and non-anthropomorphic). We also aim to implement the synergistic behavior only by altering tendon routes and spring parameters, without the need for additional mechanisms such as sliders, gears, differentials, etc.

To fulfill certain requirements of an underactuated hand design, one common way is to select parameters via optimization. There is a lot of literature in this category: for example, Birglen et al. \cite{birglen2007underactuated} presented an optimization study in detail for linkage-driven underactuated hands. Dollar and Howe \cite{dollar2011joint} optimized the kinematic configuration and joint stiffnesses to maximize the successful grasp range and minimize contact forces. Ciocarlie and Allen \cite{ciocarlie2011constrained} formulated the parameter design problem of a certain underactuated gripper as a globally convex quadratic programming. Saliba and Silva \cite{saliba2016quasi} used a quasi-dynamic analysis for the optimization of an underactuated gripper. Dong et. al \cite{dong2018geometric} optimizes the dimensions and tendon routes of a tendon-driven hand using the Genetic Algorithm. Ciocarlie et al. \cite{ciocarlie2014velo} optimized the Velo gripper to achieve both fingertip grasp and enveloping grasp using a single actuator. Compared to existing design optimization methods, the uniqueness of this work stems from the idea of optimizing the mechanisms of a highly underactuated hand in order to satisfy a specific set of desired grasps. Using an analogy to data fitting problems, the ``data points" here are the target grasps, and the ``fitting function" is the Mechanically Realizable Manifold. We minimize the deviation between the ``data points" and the ``fitting function".

The preliminary work of this study is shown in \cite{chen2018underactuated}, which uses Mechanically Realizable Manifolds in posture space. In this paper, we extend the concept of Mechanically Realizable Manifold also to the domain of joint torques, and present a more generic optimization framework, a more comprehensive design procedure, more efficient algorithms, more design examples, and in-depth analysis and discussions.

\section{Problem Formulation}

Our overall goal is to design highly underactuated hands able to perform a set of versatile grasps in a stable fashion. We formalize our problem as follows.

First of all, since the design space of a robot hand is very large, we narrow down our problem by requiring a hand model with pre-defined morphology and kinematics; \remindtext{3-3-1}{\textit{the unknowns are the (under-)actuation parameters} which determine the hand motion and force generation behavior, i.e. grasping behavior, when driven by motors.} The given hand kinematics specifies the number of fingers and links, the shape and dimensions of the fingers, the tendon connectivity pattern (specifying which tendons drive which joints), etc. The unknown parameters include tendon moment arms, restoring spring parameters, etc. We point out that we are not presenting a method to conduct the initial design for hand dimensions or kinematics, but only the (under-)actuation parameters that determine hand behavior. However, for different pre-specified kinematic design options, the results computed from our optimization algorithms can be used as performance metrics to compare between them, as we will discuss in Section \ref{sec:discussion}.

In addition, we assume we have collected a set of stable simulated grasps as data points of desired grasps (such as the ones shown in Fig. \ref{fig:cover} (a) and Fig. \ref{fig:des_grasps}, using the hand model with pre-defined kinematics. These grasps do \textit{not} account for underactuation, i.e. the joints are considered independent and the hand can exhibit arbitrary poses as needed for grasping a certain object. Meanwhile, all of these grasps are required to have force-closure, i.e. there exists a set of contact forces that produce a zero resultant wrench on the object while satisfying friction constraints. \newmodifiedtext{2-4}{In summary, each data point of a desired grasp contains the information about (i) the hand pose and (ii) a set of contacts on which the force equilibrium in such a hand pose is built.}

Our goals are as follows. We aim to design the underactuation mechanism for a hand to (approximately) conduct a set of desired grasps using \modifiedtext{1-8}{(much)} fewer actuators than joints. This creates two simultaneous requirements: even with constraints due to underactuation, the hand needs to both (i) reach the desired postures and (ii) apply the needed forces to load the grasps stably. Using the concept of underactuation manifolds, we thus aim to optimize the actuation parameters under real-world constraints, in order to (i) shape the \textit{Mechanically Realizable Posture Manifold} to approach the desired grasping poses in the pre-contact phase, and also (ii) optimize the \textit{Mechanically Realizable Torque Manifold} to provide the joint torques as close to equilibrium as possible in the post-contact phase.

\begin{figure}[t!]
\centering
\begin{tabular}{cc}
\includegraphics[width=0.5\linewidth]{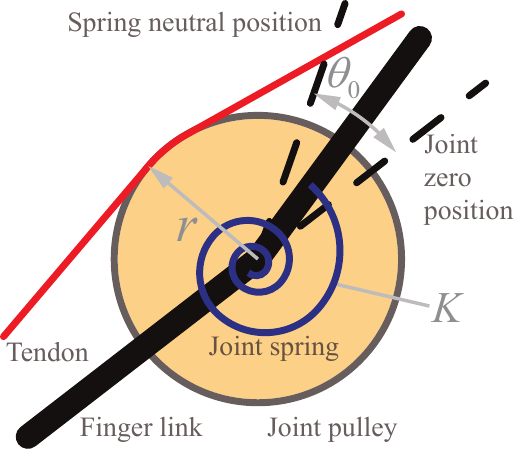} \hspace{-5mm} & \includegraphics[width=0.5\linewidth]{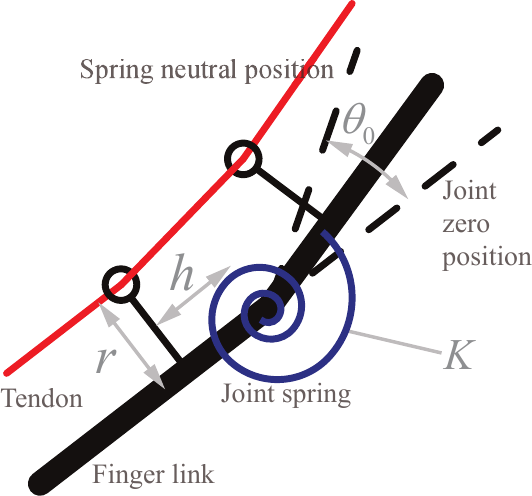} \\
(a) & (b) \\
\end{tabular}
\caption{(Under-)actuation parameters in a joint. (a) shows the parameterization for the design using pulleys and (b) shows the design with via-points.}
\label{fig:joint_params}
\end{figure}

In our case, the actuation parameters we wish to optimize are: the tendon moment arms (or the pulley radii) in the joints, the restoring spring stiffnesses, and the spring preloads (defined as the spring flexion when joint angles are zeros). Fig. \ref{fig:joint_params} shows the parameterization of two different cases where (a) shows the design with pulleys and (b) shows the design with tendon via-points. 

The optimization needs to obey certain constraints. For example, restoring spring stiffnesses are constrained by the physical dimensions of the allowed mounting space, and they are only available in a discrete series of values offered by the manufacturer. 
The ability to deal with real-world constraints is one of the advantages of our method compared to the direct implementation of synergies in \cite{brown2007inter} -- \cite{xiong2016design}. 

\section{Design Method}

\subsection{Problem Decomposition}

The aforementioned optimization is a multi-objective problem in the same design space: we need to optimize for both pre-contact kinematic behaviors and post-contact force generation behaviors, and both behaviors are related to all actuation parameters listed above. While formulating some weighted combination of these objectives is possible, it would require arbitrarily assigned weights, which we would like to avoid. Besides, it is also a high-dimensional optimization if we search all parameters simultaneously. It would thus be beneficial if we could split the optimization appropriately.

Our insight is that we can solve our problem by decomposing it into three parts: 
\begin{itemize}
	\item The optimization of Mechanically Realizable Torque Manifold
	\item The optimization of Mechanically Realizable Posture Manifold for \textit{inter}-tendon behaviors
	\item The optimization of Mechanically Realizable Posture Manifold for \textit{intra}-tendon behaviors
\end{itemize}
We explain this decomposition as follows.

To begin with, we can split the problem into the optimizations for pre-contact and post-contact behaviors, by \modifiedtext{1-3}{requiring} that the hand is in (or close to) equilibrium in both of these phases so the there is no (or negligible) post-contact movement. \newmodifiedtext{2-2}{\addedtext{1-3-1}{We emphasize that these equilibrium conditions are not assumptions but the goals of our optimization (as we will show in Subsection \ref{sec:method:torque})}}. \newremindtext{2-2-2}{\remindtext{1-3}{If the result indicates this goal cannot be achieved for any grasp in our set, we have measures to exclude such grasps and conduct the optimization again.}}

For a joint (either single DoF or multi-DoF such as a universal joint), equilibrium just before the contacts are made can be expressed as (\ref{eq:eq_before_touch}), which models the general case that multiple tendons are connecting to a joint. $\bm{r$}, $\bm{t}_{pre}$, $\bm{\theta}$, $\bm{\tau}_s$ are the vectors of tendon moment arms, pre-contact tensions, joint values and joint spring torques. $p$ is the number of tendons connected to this joint.

\begin{equation} \label{eq:eq_before_touch}
\sum_{j=1}^{p} \bm{t}_{pre,j}\times\bm{r}_j + \bm{\tau}_s(\bm{\theta}) = \bm{0}
\end{equation}

Once additional torque is applied and the grasp is loaded, equilibrium can be expressed as (\ref{eq:eq_after_touch}), where $\bm{t}_{post}$ is the post-contact tendon tension. 

\begin{equation} \label{eq:eq_after_touch}
\sum_{j=1}^{p} \bm{t}_{post,j}\times\bm{r}_j + \bm{\tau}_s(\bm{\theta}) = \bm{\tau}_{net}
\end{equation}

Combining (\ref{eq:eq_before_touch}) and (\ref{eq:eq_after_touch}) shows that the net joint torque is only affected by the moment arms but not spring parameters, shown in (\ref{eq:net_trq}).

\begin{equation} \label{eq:net_trq}
\bm{\tau}_{net} = \sum_{j=1}^{p} {(\bm{t}_{post,j}-\bm{t}_{pre,j})} \times \bm{r}_j = \sum_{j=1}^{p} \bm{t}_{net,j} \times \bm{r}_j
\end{equation}

Therefore, we can optimize for the \textit{post-contact} grasp equilibrium first (which results in zero post-contact movements as the contact forces are increased), and then optimize the rest of the parameters for the \textit{pre-contact} kinematic behaviors.

\begin{figure}[t!]
\centering
\includegraphics[width=\linewidth]{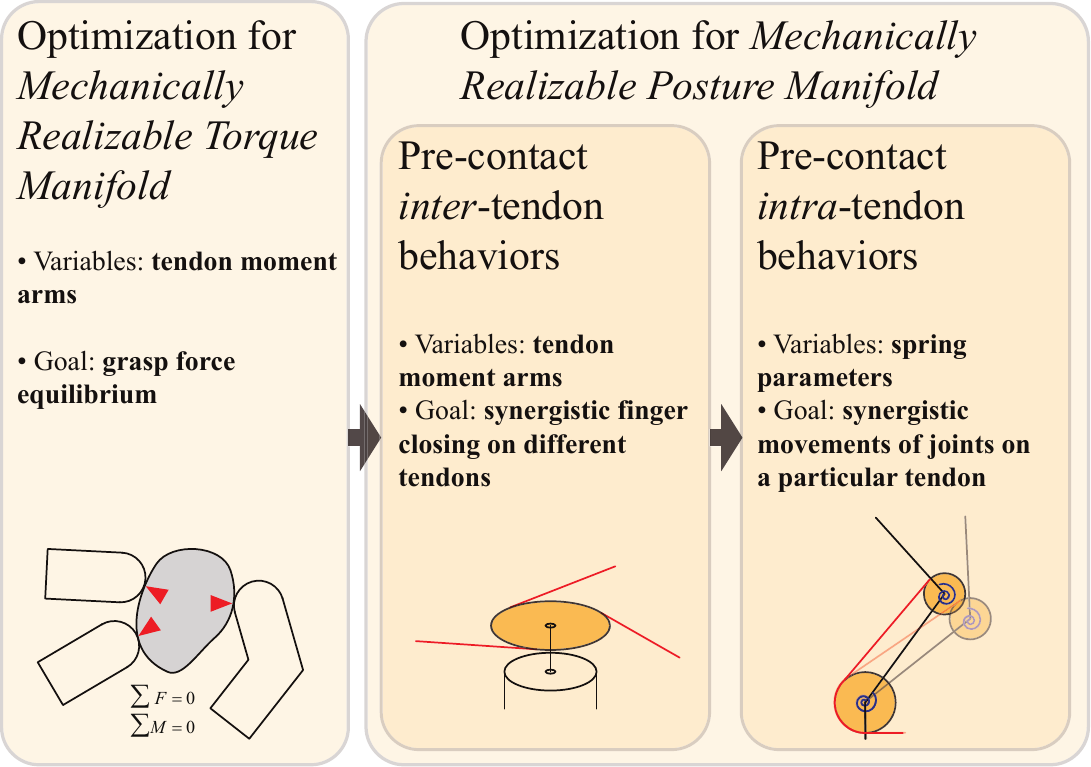}
\caption{Illustration of the optimization decomposition. The hand optimization is divided into three steps as above, and each step aim to optimize different aspects of the hand behaviors.}
\label{fig:opt_decomposition}
\end{figure}

Furthermore, the optimization for the pre-contact stage has two aspects: the \textit{inter}-tendon kinematic behavior and the \textit{intra}-tendon kinematic behavior. In the case of inter-tendon kinematics, different tendons are driven by the same motor, exhibiting no differential behavior: if one tendon stops (due to contact on a driven link), then all tendons connected to the same motor stop as well (as we do not use differential mechanisms such as floating pulleys~\cite{dollar2011joint} \cite{ciocarlie2011constrained}). In contrast, in the case of intra-tendon kinematics, multiple joints are driven by the same tendon, exhibiting a differential-like behavior: assuming a proximal joint is stopped, more distal joints can continue to move as tendons can slide on the via points or around idlers on the blocked joints. Thus, the inter-tendon optimization ensures the coordination between the kinematic chains on different tendons is as desired, and the intra-tendon optimization ensures the correlation of individual joints connected along one tendon is as desired. We select different groups of design parameters in different optimizations \addedtext{1-9}{mentioned above}: the \modifiedtext{1-9}{inter-tendon optimization} determines the tendon moment arms, and the \modifiedtext{1-9}{intra-tendon optimization} determines the spring parameters.

The decomposition of the problem is summarized in Fig. \ref{fig:opt_decomposition}. In the first step (subsection \ref{sec:method:torque}), we alter the tendon moment arms to optimize the Mechanically Realizable Torque Manifold to create post-contact equilibrium. As the optimization in this step has an infinite number of minima, which we will explain in the subsection \ref{sec:method:torque}, we pick one that also solves the second step: the optimization of Mechanically Realizable Posture Manifold for inter-tendon behaviors (subsection \ref{sec:method:inter}). At this point, we can fully determine the optimal tendon moment arms. After that, we optimize the spring parameters (stiffnesses and preloads) to make sure that the Mechanically Realizable Posture Manifold for each tendon passes closely to the desired grasps, in subsection \ref{sec:method:intra}.


\subsection{Optimization of Mechanically Realizable Torque Manifold}
\label{sec:method:torque}
The objective of this optimization is to have the underactuated hand and object as close to equilibrium as possible in the post-contact phase. The variables we alter are the tendon moment arms in all joints. We will explain at the end of this subsection that there are infinite solutions, so the optimal tendon moment arms will be determined in the next step with an extra objective. \addedtext{3-8}{We highlight that, since we aim to optimize the grasps to be in equilibrium, we do not consider dynamics in the loop, i.e. all analysis and algorithms are quasi-static.}

\subsubsection{Grasp Analysis}

To form an objective for the optimization, an evaluation of grasp stability is needed. Here we employ a well-established grasp analysis formulation \cite{ciocarlie2011constrained}. We use the (linearized) model of point-contact with friction.  For contact $k$, contact wrench $\bm{c}_k$ can be expressed as linear combinations of $\bm{\beta}_k$ --- the amplitudes of the frictional and normal components, related by a matrix $\bm{D}_k$, as shown in (\ref{eq:contact_wrench}). The matrix $\bm{D}_k$ and the vector $\bm{\beta}_k$ are also known as friction pyramid ``generator'' and ``weights'' in \cite{prattichizzo2008grasping}. Equations (\ref{eq:local_friction_constraint}) and (\ref{eq:beta_k_bound}) model the effect that contact force must be constrained inside the friction cone (or pyramid). The details about the construction of matrices $\bm{D}_k$ and $\bm{F}_k$ can be found in \cite{prattichizzo2008grasping}.

\begin{equation} \label{eq:contact_wrench}
\bm{c}_k = \bm{D}_k \bm{\beta}_k
\end{equation}

\vspace{-2mm}

\begin{equation} \label{eq:local_friction_constraint}
\bm{F}_k \bm{\beta}_k \leq \bm{0}
\end{equation}

\vspace{-2mm}

\begin{equation} \label{eq:beta_k_bound}
\bm{\beta}_k \geq \bm{0}
\end{equation}

In general, a grasp is stable if the following conditions are satisfied:
\begin{itemize}
\item
\textit{Hand equilibrium}: the active joint torques are balanced by contact forces. 
\begin{equation} \label{eq:hand_eq}
\bm{J}^T\bm{c} = \bm{J}^T\bm{D\beta} = \bm{\tau}_{eq}
\end{equation}

\item
\textit{Object equilibrium}: the resultant object wrench is zero.
\begin{equation} \label{eq:obj_eq}
\bm{Gc} = \bm{GD\beta} = \bm{0}
\end{equation}

\item
\textit{Friction constraints}: the contact forces are constrained inside the friction cone (or pyramid).
\begin{equation} \label{eq:friction_constraint}
\bm{F \beta} \leq \bm{0}
\end{equation}

\vspace{-2mm}

\begin{equation} \label{eq:beta_bounds}
\bm{\beta} \geq \bm{0}
\end{equation}

\end{itemize}

In these formulations, $\bm{J}$ is the contact Jacobian, $\bm{G}$ is the Grasp Map matrix, $\bm{D}$ and $\bm{F}$ are block-diagonal matrices constructed by $\bm{D}_k$ and $\bm{F}_k$ respectively, $\bm{c}$ and $\bm{\beta}$ are stacked vectors constructed by $\bm{c}_k$ and $\bm{\beta}_k$ respectively, and $\bm{\tau}_{eq}$ is the desired joint torque vector to create hand equilibrium.

\addedtext{1-2-1}{For a force-closure grasp with a given pose (as the desired grasps in this paper), there may be an infinite number of equilibrium torque combinations that satisfy (\ref{eq:hand_eq}) -- (\ref{eq:beta_bounds})}. To select one, this formulation can be turned into an optimization problem by switching any one of the conditions to an objective. We further discuss the optimization formulation in the next subsection.

\subsubsection{Optimization Formulation}

The overview of our formulation is as follows: we incorporate a dual-layer optimization framework, in which the inner layer is an optimization to calculate a quality metric of a specific grasp given a certain set of design parameters, and the outer layer is a search over the parameters for all considered grasps.

In our problem, since the hand is underactuated, the joint torques are not independent. Instead, the actually generated net torque $\bm{\tau}^{gen}_{net}$ follows the relationship:
\begin{equation} \label{eq:actuation}
\bm{\tau}^{gen}_{net} = \bm{A}\bm{t}_{net}
\end{equation}
where $\bm{A}$ is the Actuation Matrix, which is a function of the tendon moment arms $r_1, r_2, \cdots, r_m$ ($m$ is the number of DoFs), and may or may not be configuration-dependent. $\bm{t}_{net}$ is the net tendon tension vector compared to the tension just before touch.

For a set of given tendon moment arms and a given grasp pose, we wish to find the contact force magnitudes $\bm{\beta}$ and the net tension in each tendon $\bm{t}_{net}$ which solve (\ref{eq:hand_eq})--(\ref{eq:actuation}). We change the search for the exact solution to an optimization problem by turning hand equilibrium (\ref{eq:hand_eq}) from a constraint into an objective function. The unbalanced joint torque vector $\Delta\bm{\tau}_{post}$ is shown in (\ref{eq:unbalanced_trq_grasping}) (subscript $post$ meaning ``post-contact"). We use the norm of this vector $\rVert \Delta\bm{\tau}_{post} \rVert$ as the stability metric, and a lower value is considered better. 

\begin{equation} \label{eq:unbalanced_trq_grasping}
\Delta\bm{\tau}_{post} = \bm{\tau}_{eq} - \bm{\tau}^{gen}_{net} = \bm{J}^T\bm{D\beta} - \bm{At}_{net}
\end{equation}

\newremindtext{2-2-1}{\addedtext{1-3-2}{For any grasp, the value of this objective after optimization is completed might be (i) exactly zero, (ii) a small value (below an empirically determined threshold), or (iii) a large value (exceeding the same threshold). For case (i), no post-contact movement will occur. For case (ii), we assume that the small unbalanced torques will be compensated by either friction or a negligible reconfiguration of the hand which will not affect stability. For case (iii), we have measures to exclude such grasps and conduct the optimization again, as we will discuss at the end of this subsection.}}

The inner layer problem -- to find the minimal norm of unbalanced torques for one given grasp -- is a convex Quadratic Program (QP) as follows, shown in (\ref{eq:qp1_find}) -- (\ref{eq:qp1_st4}).
\vspace{2mm}

\text{\textit{find:}}
\begin{equation} \label{eq:qp1_find}
\bm{x} = \begin{bmatrix}\bm{\beta} \\ \bm{t}_{net}\end{bmatrix}
\end{equation}

\text{\textit{minimize:}}
\begin{equation} \label{eq:qp1_min}
\lVert \Delta\bm{\tau}_{post} \rVert^2 = \lVert \bm{Qx} \rVert^2 = \bm{x}^T\bm{Q}^T\bm{Qx}
\end{equation}
$$ \text{where }  \bm{Q} = \begin{bmatrix}\bm{J}^T\bm{D} & \bm{-A}\end{bmatrix}$$

\vspace{2mm}
\text{\textit{subject to:}}
\begin{equation} \label{eq:qp1_st1}
\begin{bmatrix}\bm{GD} & \bm{O}\end{bmatrix}\bm{x} = \bm{0}
\end{equation}

\vspace{-4mm}

\begin{equation} \label{eq:qp1_st2}
\begin{bmatrix}\bm{F} & \bm{O}\end{bmatrix}\bm{x} \leq \bm{0}
\end{equation}

\vspace{-4mm}

\begin{equation} \label{eq:qp1_st3}
\bm{x} \geq \bm{0}
\end{equation}

\vspace{-4mm}

\begin{equation} \label{eq:qp1_st4}
\begin{bmatrix}1 \cdots 1\end{bmatrix}\begin{bmatrix}\bm{J}^T\bm{D} & \bm{O}\end{bmatrix}\bm{x} = 1
\end{equation}

The constraints (\ref{eq:qp1_st1}) -- (\ref{eq:qp1_st3}) are extended versions of (\ref{eq:obj_eq}) -- (\ref{eq:beta_bounds}) and the last one (\ref{eq:qp1_st4}) prevents the trivial solution where all contact forces and joint torques are zeros, by constraining the sum of joint torques equal to one. In this way, the calculated grasp stability metric $\lVert \Delta\bm{\tau}_{post} \rVert$ is a normalized unitless torque.

Since the aforementioned inner layer can give a stability metric for a specific desired grasp, the outer-layer is a global search over the tendon moment arms for all considered grasps using the inner layer results. The objective function for the global search is an overall metric using the root of squared sum of all grasps' quality metrics, shown in (\ref{eq:force_optim_min}). The outer layer global optimization is formulated as:

\vspace{2mm}
\text{\textit{search:}}
\begin{equation} \label{eq:force_optim_search}
r_1, r_2, \cdots, r_{m}
\end{equation}

\text{\textit{minimize:}}
\begin{equation} \label{eq:force_optim_min}
\begin{aligned}
{f}_{trq}(r_1, r_2,  \cdots, & r_{m}) = \left(\sum_{i=1}^{n} \lVert \Delta\bm{\tau}_{post, i} \rVert^2 \right) ^ \frac{1}{2}
\end{aligned}
\end{equation}

\text{\textit{subject to:}}
\begin{equation} \label{eq:force_optim_st}
r_i \in [r_{lb}, r_{ub}], ~ i=1,2,\cdots,m
\end{equation}

\text{\textit{save:}}
\begin{equation} \label{eq:force_optim_save}
{f}_{trq}^{min} = min({f}_{trq}(r_1, r_2, \cdots, r_{m}))
\end{equation}
where $\lVert \Delta\bm{\tau}_{post, i} \rVert$ is the individual stability metrics computed by the QP in (\ref{eq:qp1_find}) -- (\ref{eq:qp1_st4}) for each simulated desired grasp, $n$ is the number of grasps, $r_{lb}$ and $r_{ub}$ are the lower and upper bounds of tendon moment arms.

Although the inner-layer is convex, the outer-layer is not, and is not trivial to be reformulated as a convex problem. Therefore we decided to use a stochastic global search. The optimizer we chose is the Covariance Matrix Adaptation – Evolutionary Strategy (CMA-ES) \cite{hansen2001completely} \cite{hansen2003reducing}. It is a stochastic, derivative-free algorithm for black-box global optimization, in which the covariance matrix of the distribution of the candidate solutions is updated adaptively in each generation. This method learns a stochastic second-order approximation of the objective, and drives the candidate solutions to the optimum, even when the function is ill-conditioned. 

We aim to find a Mechanically Realizable Manifold that minimizes the unbalanced torque $\lVert \Delta\bm{\tau}_{post, i} \rVert$ for all grasps. In practice, we exclude from this optimization the \addedtext{1-11}{difficult-to-achieve grasps} where the unbalanced torque is found to be \modifiedtext{1-11}{larger than an empirically determined threshold}, and then we iterate again and solve the optimization problem for the rest of grasps. In this way, we can design the hand to perform as well as possible on the grasps possible to create, instead of attempting to also satisfy equilibrium for impossible grasps. Moreover, the number of grasps excluded in this way is an important metric of the hand's overall capabilities to achieve our design goals. All our results, presented later in the paper, will thus report both the number of excluded grasps, and the values of all optimization objectives (computed over the grasps that have been kept).

Here we explain why the minimum is not unique. Let us assume we have found a set of optimal tendon moment arms, with a certain $\bm{x}$ (thus a certain set of $\bm{t}_{net}$) in (\ref{eq:qp1_find}). As long as the QP (\ref{eq:qp1_find}) -- (\ref{eq:qp1_st4}) can find an $\bm{x}$ (and thus $\bm{t}_{net}$) that can keep $\bm{A}\bm{t}_{net}$ the same when we alter the tendon moment arms (one possible way is to scale $\bm{x}$ (or $\bm{t}$) while scaling entries in $\bm{A}$), the metric of the inner layer (the QP (\ref{eq:qp1_find}) -- (\ref{eq:qp1_st4})) remains minimal, then the objective function value of the outer layer (the global search (\ref{eq:force_optim_min}) (\ref{eq:force_optim_st})) remains minimal. Therefore there are other optimal solutions for the tendon moment arms. 

\begin{figure}[t!]
\centering
\includegraphics[width=\linewidth]{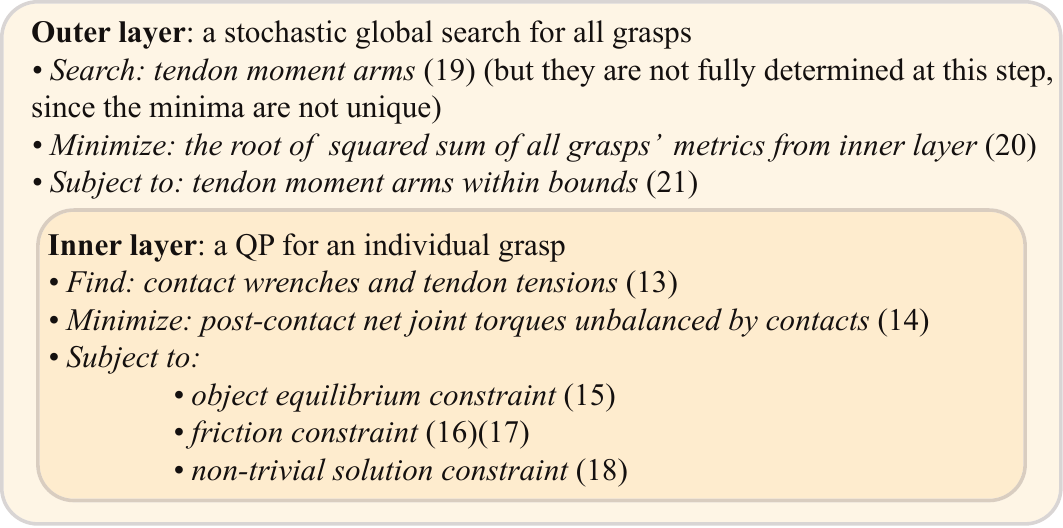}
\caption{Summary of the optimization for Mechanically Realizable Torque Manifolds. The algorithm consists of two layers: the outer layer optimization search over tendon moment arms while the inner layer optimization gives a stability metric for the outer layer.}
\label{fig:force_opt_summary}
\end{figure}

The optimization of Mechanically Realizable Torque Manifold is summarized as Fig. \ref{fig:force_opt_summary}. After this step, we can find a Mechanically Realizable Torque Manifold, on which the joint torque distribution, determined by the tendon moment arms, is as close to equilibrium as possible for all the considered grasps. We will pick the one set of values for tendon moment arms from the non-unique and equally good solutions with an extra objective in the next subsection.

\subsection{Optimization of the Mechanically Realizable Posture Manifold for Inter-tendon Kinematic Behaviors}
\label{sec:method:inter}
In this step, we aim to get a coordinated finger movement among different tendons. As the hand moves from a reference starting pose into a grasp pose, each tendon must travel a specific amount, dictated by the movement of each joint as well as the tendon's moment arm around each joint. (This assumes that tendons are inextensible, and also are not allowed to become slack as they lose force transmission abilities.) However, tendon travel must also be in accordance with the travel of the motor that the tendon is connected to, a problem which becomes non-trivial for the case of multiple tendons connected to the same motor.

We again translate this requirement into an optimization problem. For each of the desired grasp poses, we minimize the differences between required and actual tendon travel for all tendons connected to the same motor, with motor travel angle(s) as a free variable. 

This optimization is only related to the tendon moment arms, which are picked from the set of equally good solutions in the previous optimization, and can be fully determined with the extra objective in this step.

We note that the pool of desired grasps is a superset of the one in the previous subsection: in addition to desired grasps, we also include the fully open configuration, \addedtext{1-4-1}{in which all joints are opened to their mechanical limits.} For each grasping pose in the original grasp pool, we add an opening pose, such that opening and closing poses are always in pairs. In this way, we can ensure the hand can actually open, instead of moving between desired grasping poses.

We also follow the dual-layer optimization framework as in the previous subsection. The inner layer minimizes the norm of an error vector whose entries are the difference between the tendon travel the motor collects and the one the grasp requires. The outer layer is also a stochastic global optimization to minimize the overall metric for all grasps, with a constraint that the objective function in the previous optimization needs to reach its minimum.

In the inner layer, the error vector can be expressed as (\ref{eq:tendon_travel}), where the matrix $\bm{M}$ is a motor-tendon connection matrix whose entries can be either motor pulley radius $r_1$, $r_2$, $\cdots$, $r_m$ (meaning the corresponding tendon is connected to the corresponding motor) or zero (meaning not connected), $\bm{\theta}_{mot}$ is a vector of angles that the motors moved, and $\bm{s}$ is the tendon travel vector whose entries are the travel of the corresponding tendon required by the joint values in the desired grasp (compared to zero positions).

\begin{equation} \label{eq:tendon_travel}
\bm{e} = \bm{M}\bm{\theta}_{mot} - \bm{s}
\end{equation}

Minimizing the norm of the error vector $\lVert \bm{e} \rVert$ is a QP over the angles the motors moved $\bm{\theta}_{mot}$ , with no constraints. It is shown in the QP (\ref{eq:qp2_find}) (\ref{eq:qp2_min}) below.

\vspace{2mm}
\text{\textit{find:}}
\begin{equation} \label{eq:qp2_find}
\bm{\theta}_{mot}
\end{equation}

\text{\textit{minimize:}}
\begin{equation} \label{eq:qp2_min}
\begin{aligned}
\lVert \bm{e} \lVert^2 = \bm{\theta}_{mot}^T\bm{M}^T\bm{M}\bm{\theta}_{mot} -  2\bm{s}^T\bm{M}\bm{\theta}_{mot} + \bm{s}^T\bm{s}
\end{aligned}
\end{equation}

The outer layer, which is a global optimization over the tendon moment arms to minimize the overall metric ${f}_{inter} (r_1, r_2, \cdots,  r_{m})$(the root of squared sum of inner layer results), has to obey the constraint that the objective of previous optimization (of Mechanically Realizable Torque Manifold) needs to take its minimum value, shown in (\ref{eq:inter_tendon_opt_st}) (\ref{eq:inter_tendon_opt_st2}). \newaddedtext{2-3}{With this constraint, the tendon moment arms can only take values which minimize the objective in the previous optimization. Then we can find a solution to both the previous and the current optimization.} The constraint is handled in a soft way by giving a high penalty if the constraint is violated, and more penalty if the candidate solutions are farther away from the feasible region. The outer layer optimization is as follows:

\text{\textit{find:}}
\begin{equation} \label{eq:inter_tendon_opt_find}
r_1, r_2, \cdots, r_{m}
\end{equation}

\text{\textit{minimize:}}
\begin{equation} \label{eq:inter_tendon_opt_min}
\begin{aligned}
{f}_{inter} (r_1, r_2, \cdots,  r_{m}) = \left(\sum_{i=1}^{n} \lVert \bm{e}_{i} \lVert^2 \right) ^ \frac{1}{2}
\end{aligned}
\end{equation}

\text{\textit{subject to:}}
\begin{equation} \label{eq:inter_tendon_opt_st}
{f}_{trq}(r_1, r_2, \cdots, r_{m}) = f_{trq}^{min}
\end{equation}
\begin{equation} \label{eq:inter_tendon_opt_st2}
r_i \in [r_{lb}, r_{ub}], ~ i=1,2,\cdots,m
\end{equation}
where $\lVert \bm{e}_{i} \lVert$ is the individual inner layer metric for each grasp calculated from the QP (\ref{eq:qp2_find}) (\ref{eq:qp2_min}), $f_{trq}^{min}$ is the minimal function value saved from the previous step, and $r_{lb}$ and $r_{ub}$ are the lower and upper bounds of the pulley radii $r$. We also incorporate CMA-ES for the global search.

\begin{figure}[t!]
\center
\includegraphics[width=\linewidth]{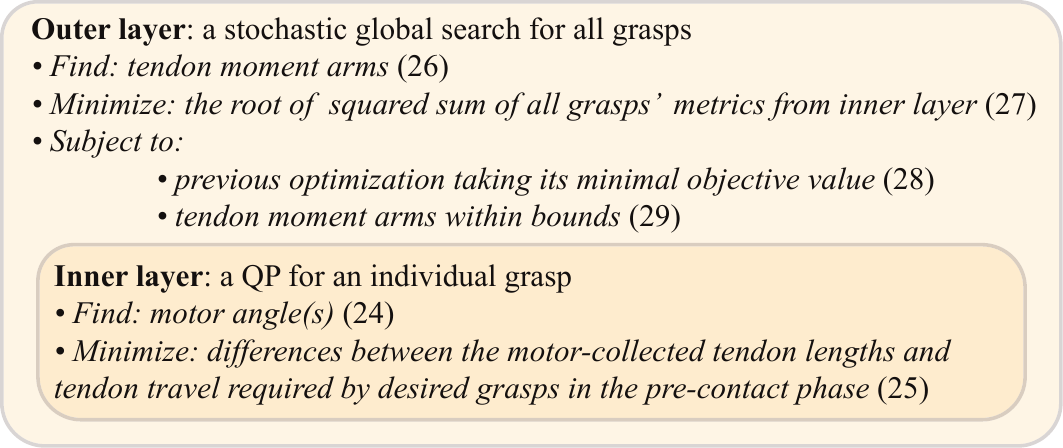}
\caption{Summary of the optimization for  Mechanically Realizable Posture Manifold for inter-tendon kinematic behaviors. The algorithm consists of two layers: the outer layer optimization search over tendon moment arms while the inner layer optimization gives a tendon travel error metric for the outer layer.}
\label{fig:inter_tendon_opt_summary}
\end{figure}

At this point, the tendon moment arms $r_1, r_2, \cdots, r_{m}$ are fully determined. The optimization of Mechanically Realizable Posture Manifold for inter-tendon kinematic behaviors can be summarized as Fig. \ref{fig:inter_tendon_opt_summary}.

\subsection{Optimization of Mechanically Realizable Posture Manifold for Intra-tendon Kinematic Behaviors}
\label{sec:method:intra}
The goal of this optimization is to coordinate the movements of different joints along one certain tendon to have the Mechanically Realizable Posture Manifold close to the desired grasp poses. We import the tendon moment arms from the previous optimization, and the remaining parameters to optimize are: the spring stiffnesses and the spring preloads.

We translate the goal of quasi-statically reaching desired grasps to the one of minimizing the pre-contact unbalanced spring torques if the hand is posed in the desired grasp configurations. Lower unbalanced torque means the Mechanically Realizable Posture Manifold is closer to the desired grasp. We emphasize that in this part we only consider the equilibrium of the hand itself, without the object. 

The unbalanced spring torque vector can be calculated as (\ref{eq:unbalanced_trq_free_motion}), where the matrix $\bm{A}$ is the aforementioned Actuation Matrix, and $\bm{\tau}_{spr}$ is a vector of spring torques calculated by given spring parameters and given poses (shown in (\ref{eq:spring_trq})) . We note that here the $\bm{t}$ is the absolute tendon tension, which is different from the $\bm{t}_{net}$

\begin{equation} \label{eq:unbalanced_trq_free_motion} 
\Delta\bm{\tau}_{pre}  = \bm{A}\bm{t} - \bm{\tau}_{spr}
\end{equation}

\begin{equation} \label{eq:spring_trq}
\bm{\tau}_{spr} = [K_1(\theta_1 + \theta_{01}), \cdots,K_m(\theta_m + \theta_{0m})]^T
\end{equation}

We wish to find the $\bm{t}$ vector resulting in a minimum norm of unbalanced joint torques, which is also a convex QP as shown below.

\vspace{2mm}
\text{\textit{find:}}
\begin{equation} \label{eq:qp3_find}
\bm{t}
\end{equation}

\text{\textit{minimize:}}
\begin{equation} \label{eq:qp3_min}
\begin{aligned}
\lVert\Delta\bm{\tau}_{pre}\rVert^2 = \bm{t}^T\bm{A}^T\bm{At} - 2\bm{\tau}_{spr}^T\bm{At} + \bm{\tau}_{spr}^T\bm{\tau}_{spr}
\end{aligned}
\end{equation}

\text{\textit{subject to:}}
\begin{equation} \label{eq:qp3_st}
\bm{t} \geq \bm{0}
\end{equation}

\begin{figure}[t!]
\center
\includegraphics[width=\linewidth]{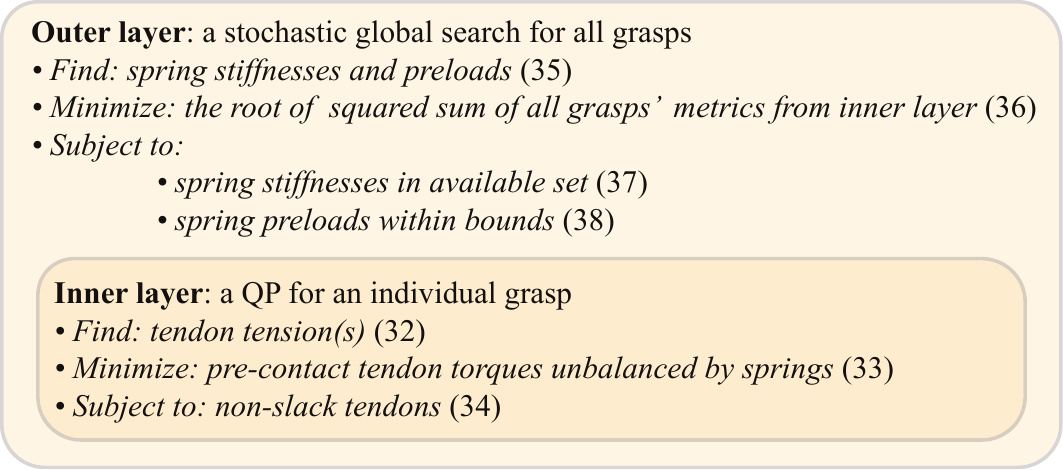}
\caption{Summary of the optimization for Mechanically Realizable Posture Manifold for intra-tendon kinematic behaviors. The algorithm consists of two layers: the outer layer optimization search over spring parameters while the inner layer optimization gives a metric for the outer layer.}
\label{fig:intra_tendon_opt_summary}
\end{figure}

We still incorporate the dual-layer optimization. In the outer layer, we use CMA-ES to find the global optimum of the overall metric ${f}_{intra} (K_1, \cdots, K_{m}, \theta_{01}, \cdots, \theta_{0m})$, as shown in  (\ref{eq:intra_tendon_opt_find}) -- (\ref{eq:intra_tendon_opt_st}). In practice, since the intra-tendon kinematic behaviors are only related to the joint along that tendon, we can perform optimization only for those joints separately at a time, in order to reduce the search dimensions. It is also needed to note that the spring stiffnesses can only take discrete values offered by the manufacturer, so we incorporate the integer handling in CMA-ES.

\vspace{2mm}
\text{\textit{find:}}
\begin{equation} \label{eq:intra_tendon_opt_find}
K_1, K_2, \cdots, K_{m}, \theta_{01}, \theta_{02}, \cdots, \theta_{0m}
\end{equation}

\text{\textit{minimize:}}
\begin{equation} \label{eq:intra_tendon_opt_min}
\begin{aligned}
{f}_{intra} & (K_1, \cdots, K_{m}, \theta_{01}, \cdots, \theta_{0m}) \\
& = \left(\sum_{i=1}^{n} \lVert \Delta  \bm{\tau}_{pre, i}\rVert^2 \right) ^ \frac{1}{2}
\end{aligned}
\end{equation}

\text{\textit{subject to:}}
\begin{equation} \label{eq:intra_tendon_opt_st}
K_i \in \mathbbm{K}, ~ i=1,2,\cdots,m
\end{equation}
\begin{equation} \label{eq:intra_tendon_opt_st2}
\theta_{0i} \in [\theta_{0lb}, \theta_{0ub}], ~ i=1,2,\cdots,m
\end{equation}
where the $\lVert\Delta\bm{\tau}_{pre,i}\rVert$ is the individual metric for each grasp from the inner layer optimization (\ref{eq:qp3_find}) -- (\ref{eq:qp3_st}), $\mathbbm{K}$ is the set of discrete spring stiffnesses provided by the manufacturer, and $\theta_{0lb}$ and $\theta_{0ub}$ are the lower and upper bounds of the spring preload angle $\theta_0$ .

The optimization of Mechanically Realizable Posture Manifold for intra-tendon kinematic behaviors can be summarized as Fig. \ref{fig:intra_tendon_opt_summary}.


\subsection{Summary}
\begin{figure}[t!]
\center
\includegraphics[width=\linewidth]{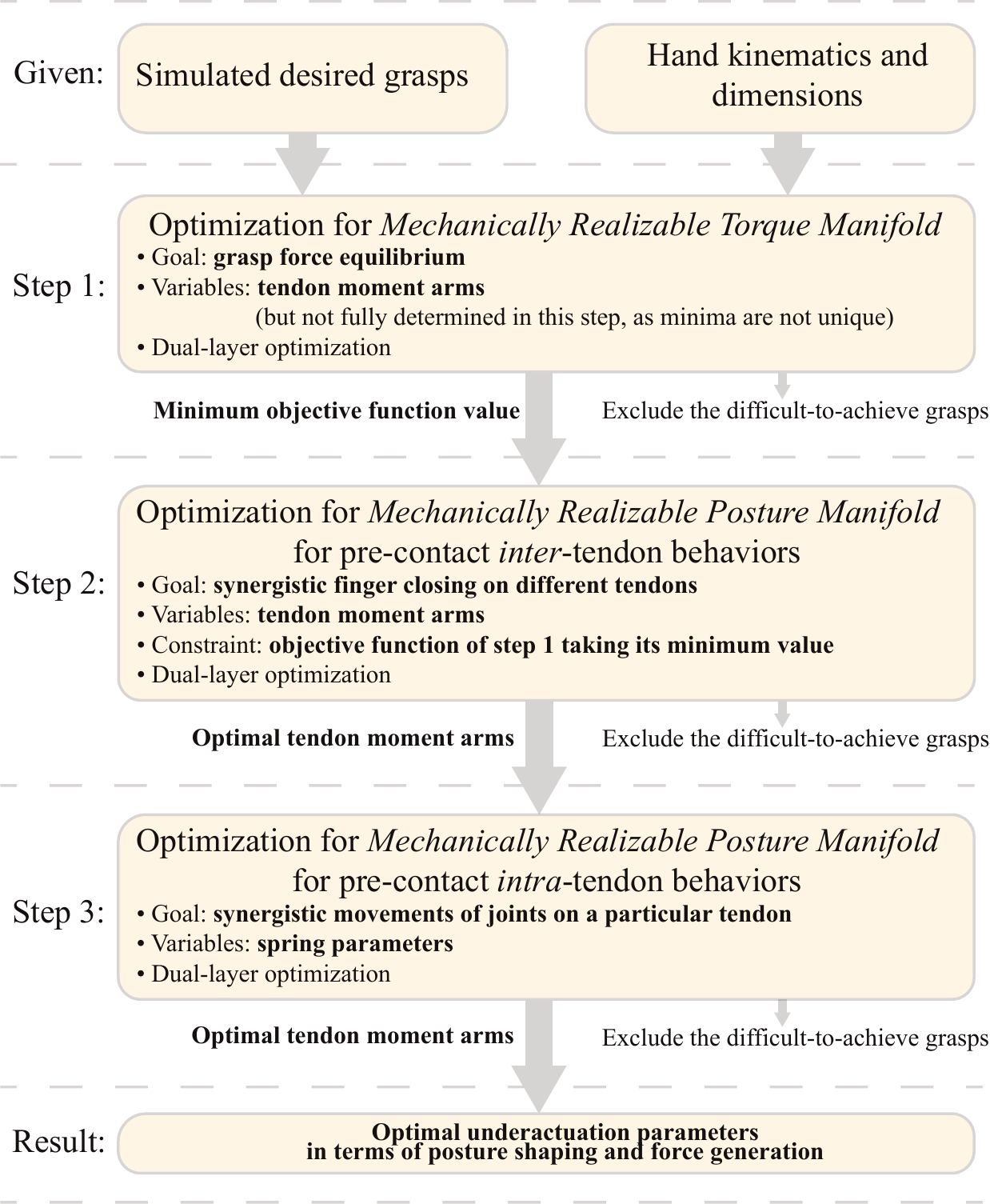}
\caption{\modifiedtext{1-12}{Summary of the proposed three-step optimization of Mechanically Realizable Manifolds.}}
\label{fig:algorithm_summary}
\end{figure}

Fig. \ref{fig:algorithm_summary} is the recap of our method. The design process starts from pre-specified desired grasps and hand kinematics, then goes through the aforementioned three steps for different aspects of hand behaviors: the optimization of Mechanically Realizable Torque Manifold, as well as the Mechanically Realizable Posture Manifold for inter- and intra-tendon behaviors. Finally, it results in a set of optimal actuation parameters for both posture shaping and force generation.

\section{Design Cases and Evaluations}

We completed three concrete design examples of three-finger underactuated hands for use in the \textit{Astrobee} robot \cite{bualat2015astrobee} in the International Space Station (ISS). The \textit{Astrobee} is a cube-shaped free-flying assistive robot, designed to help the astronauts for in-cabin monitoring and many other tasks. We aim to enable it to do object retrieval by mounting a simple arm and a versatile hand to its payload bay.

Our method is suitable for this design task for several reasons. First of all, a highly synergistic underactuated but versatile hand is needed for this application because of the limited onboard space and control signals. Second, the objects in the ISS are known and relatively unchanged, which means we have a given set of objects, and including them in our simulated grasp set may have a good chance of good performance in practice. Third, \remindtext{3-3-2}{the available room to store the hand inside the Astrobee robot is given, so the dimensions of the hand can be specified beforehand} as required by our method.


\subsection{Design Case I: Single-motor Hand with Roll-pitch Fingers}

\begin{figure}[t!]
\centering
\begin{tabular}{cc}
\includegraphics[width=0.5\linewidth]{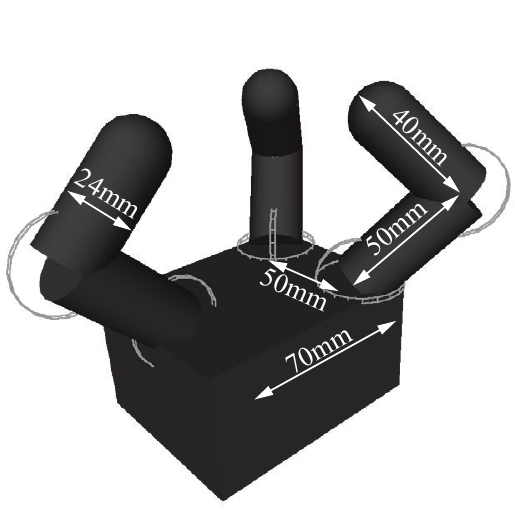} \hspace{-6mm} & \includegraphics[width=0.5\linewidth]{images/actuation1.pdf} \\
(a) & (b)
\end{tabular}
\caption{(a) Hand model with pre-defined kinematics and (b) actuation scheme of Design Case I}
\label{fig:hand_and_actuation1}
\end{figure}

The first design case is a three-finger single-motor hand with roll-pitch finger configuration (we define rolling axes perpendicular to the palm). The hand has altogether eight joints: two joints (proximal and distal joints) on the thumb, and three joints (the finger roll joint, proximal joint and distal joint) on the opposing two fingers. The hand models including kinematic configuration and tendon connectivity pattern are shown in Fig. \ref{fig:hand_and_actuation1}.

\subsubsection{Grasp Collection}

\begin{figure}[t!]
\begin{tabular}{cccccc}
 \hspace{-5mm} \includegraphics[width=0.22\linewidth]{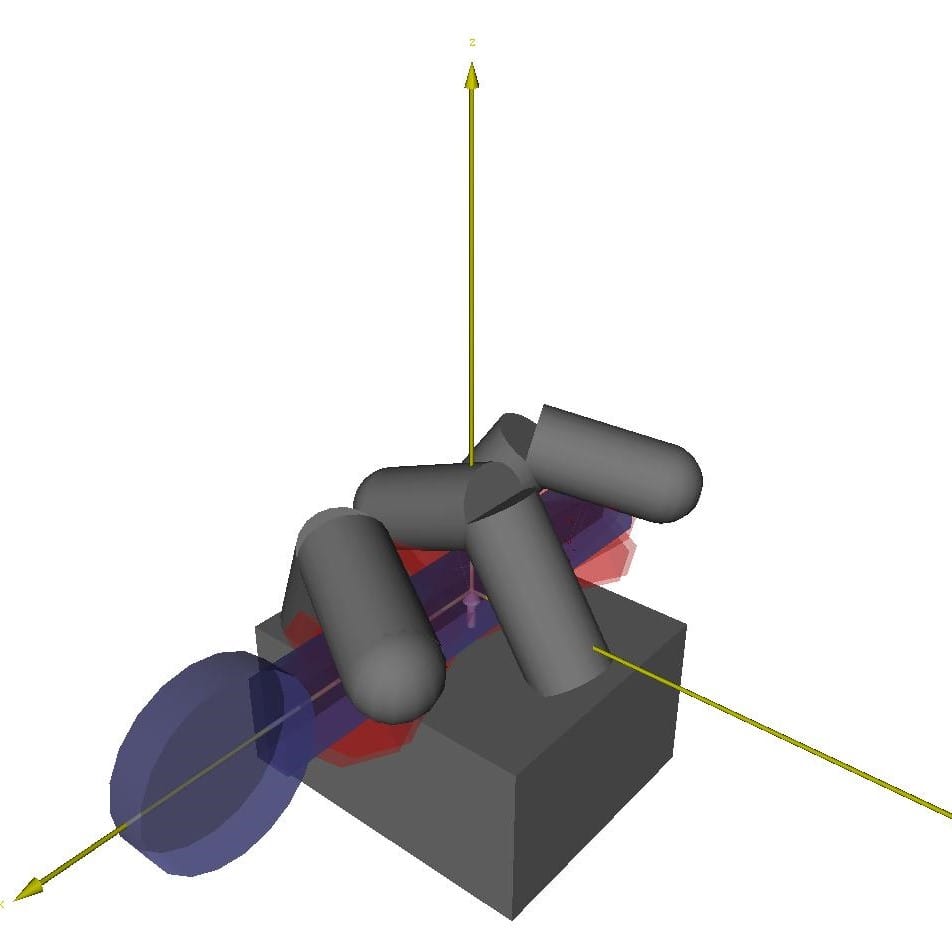} 
&\hspace{-10mm} \includegraphics[width=0.22\linewidth]{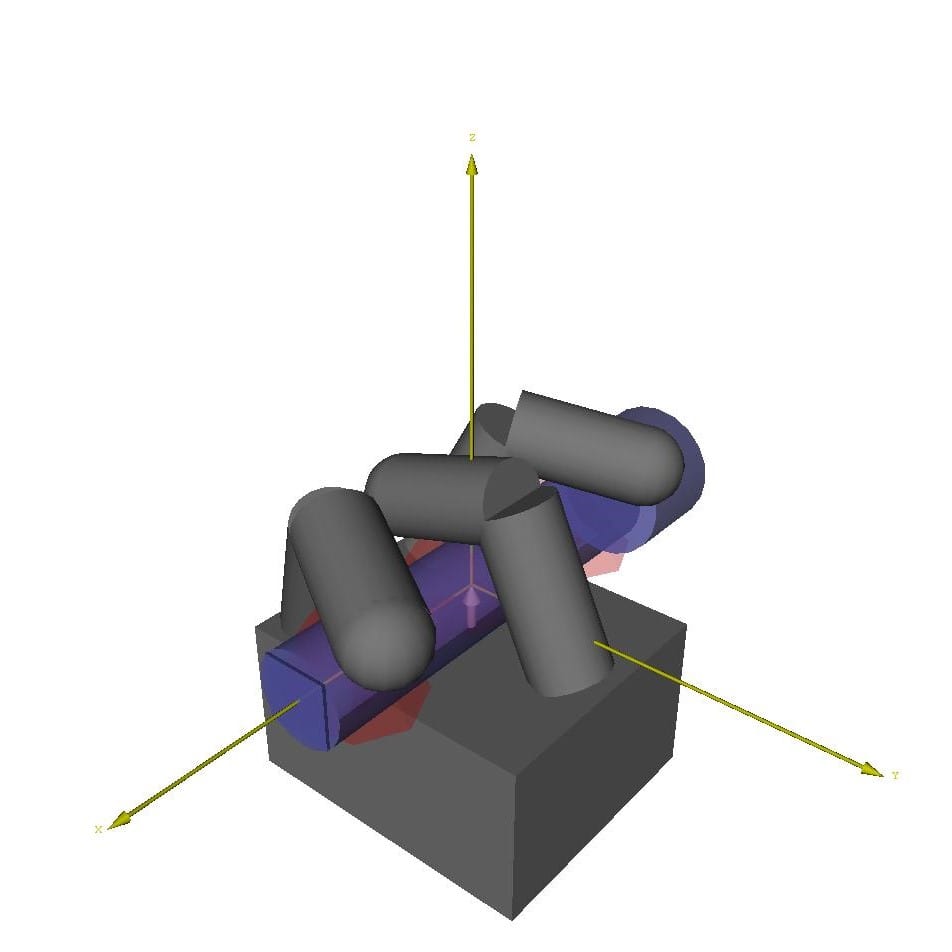}
&\hspace{-10mm} \includegraphics[width=0.22\linewidth]{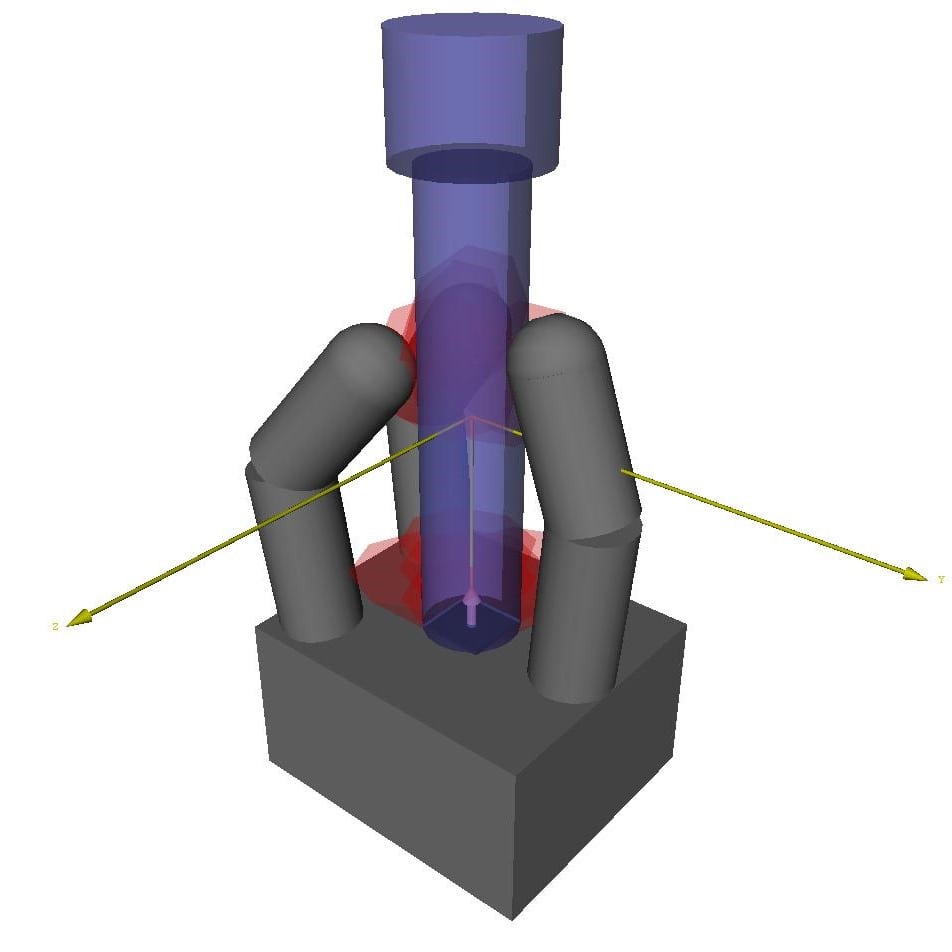}
&\hspace{-10mm} \includegraphics[width=0.22\linewidth]{images/des_grasp4.jpg}
&\hspace{-10mm} \includegraphics[width=0.22\linewidth]{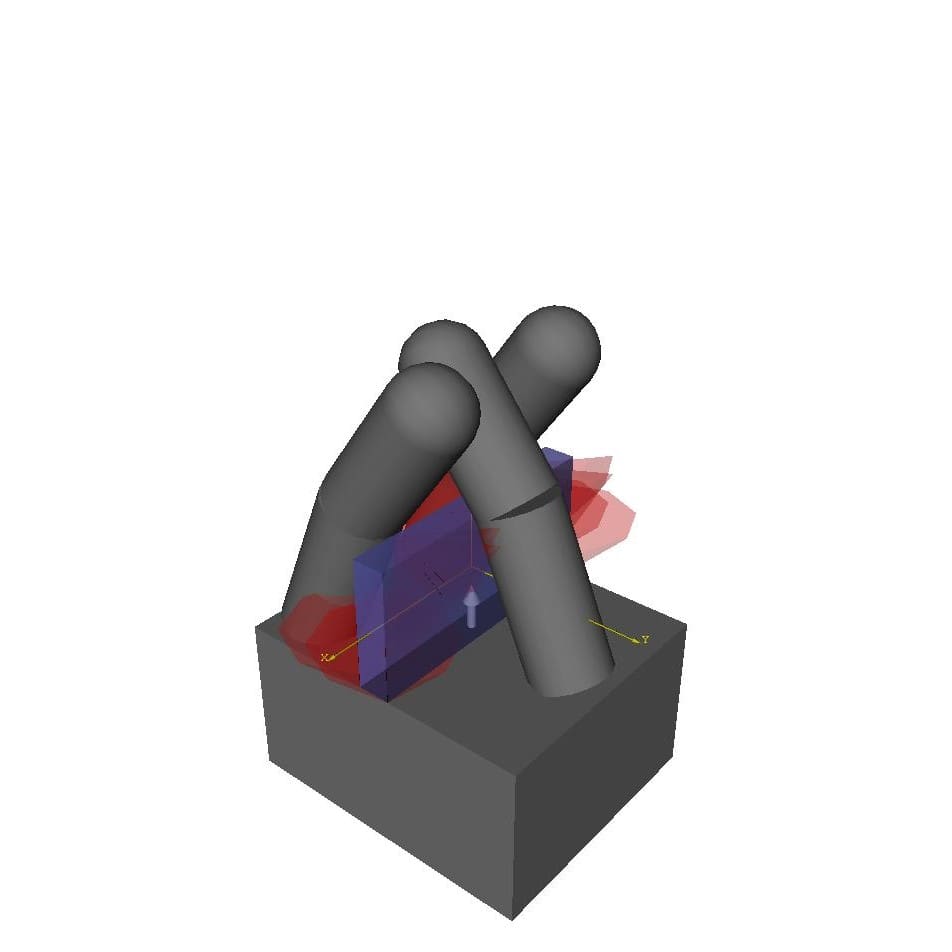}
&\hspace{-10mm} \includegraphics[width=0.22\linewidth]{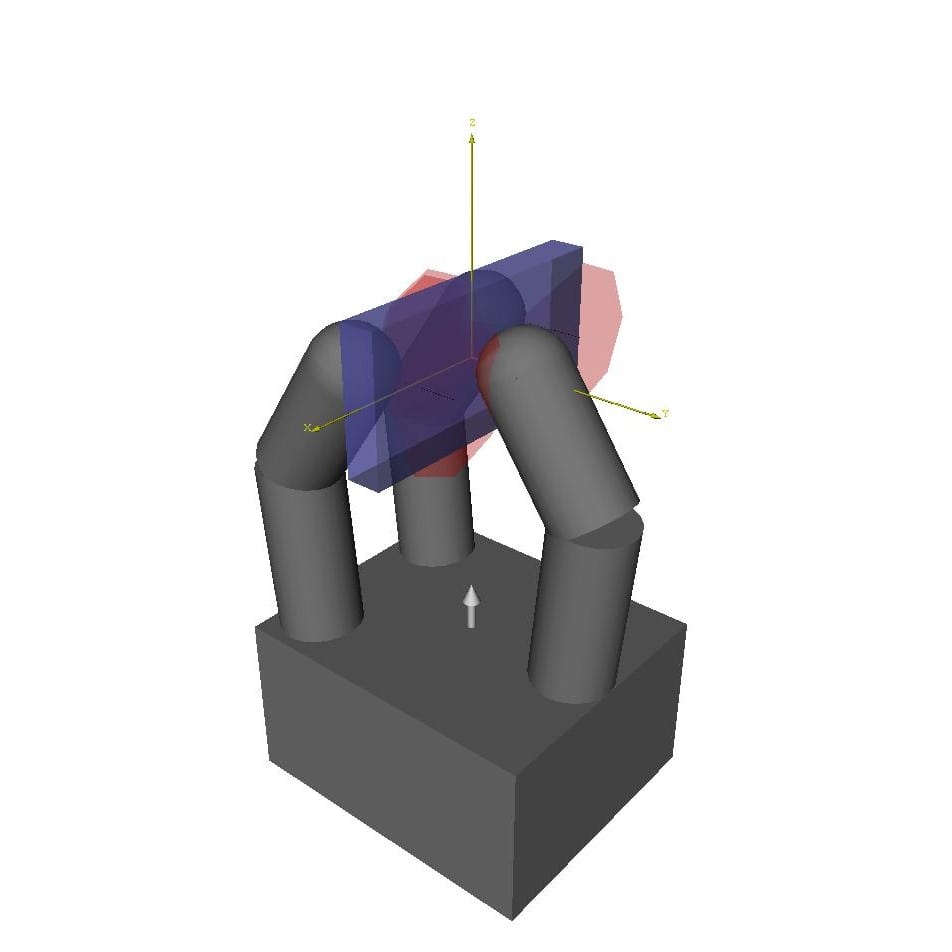}
 \vspace{-1mm}
\\
 \hspace{-5mm} \includegraphics[width=0.22\linewidth]{images/des_grasp7.jpg}
&\hspace{-10mm} \includegraphics[width=0.22\linewidth]{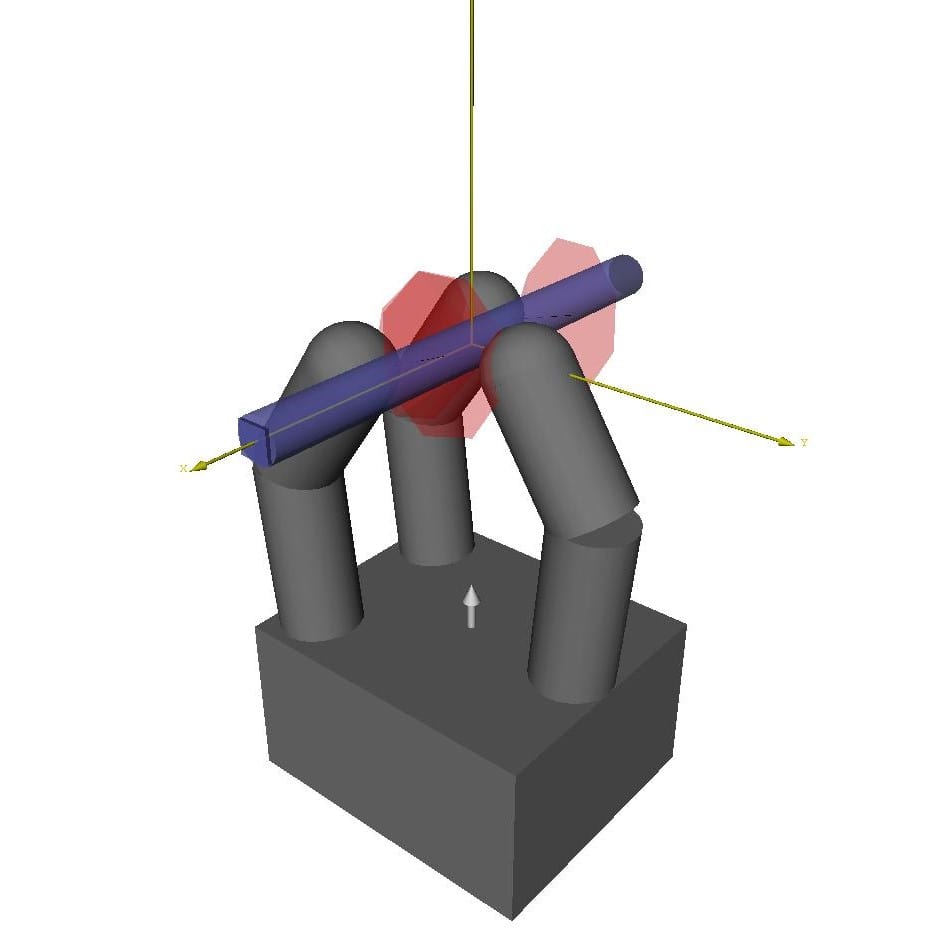}
&\hspace{-10mm} \includegraphics[width=0.22\linewidth]{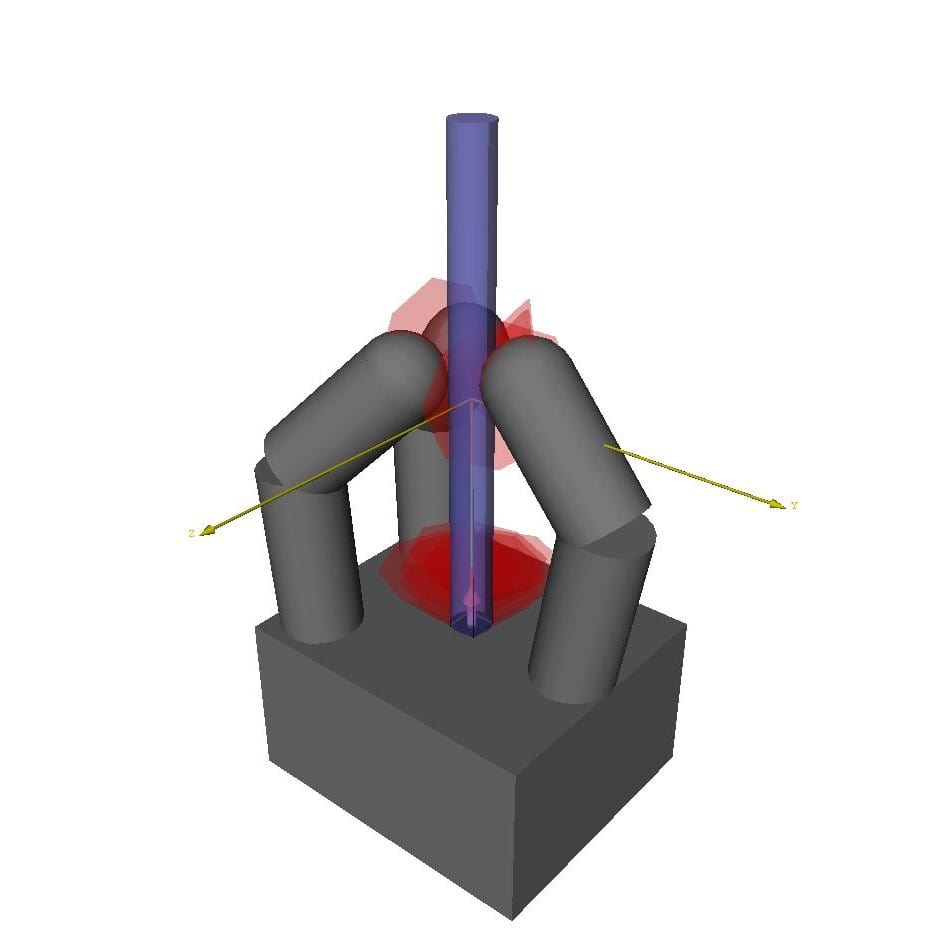}
&\hspace{-10mm} \includegraphics[width=0.22\linewidth]{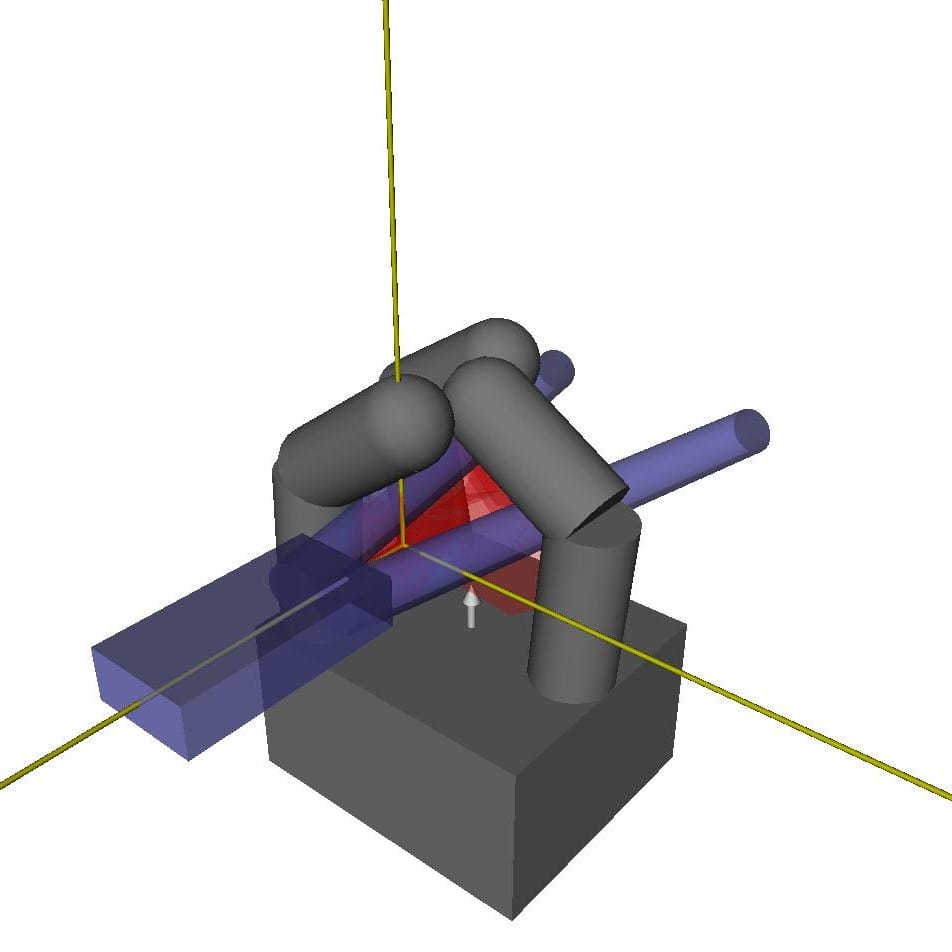}
&\hspace{-10mm} \includegraphics[width=0.22\linewidth]{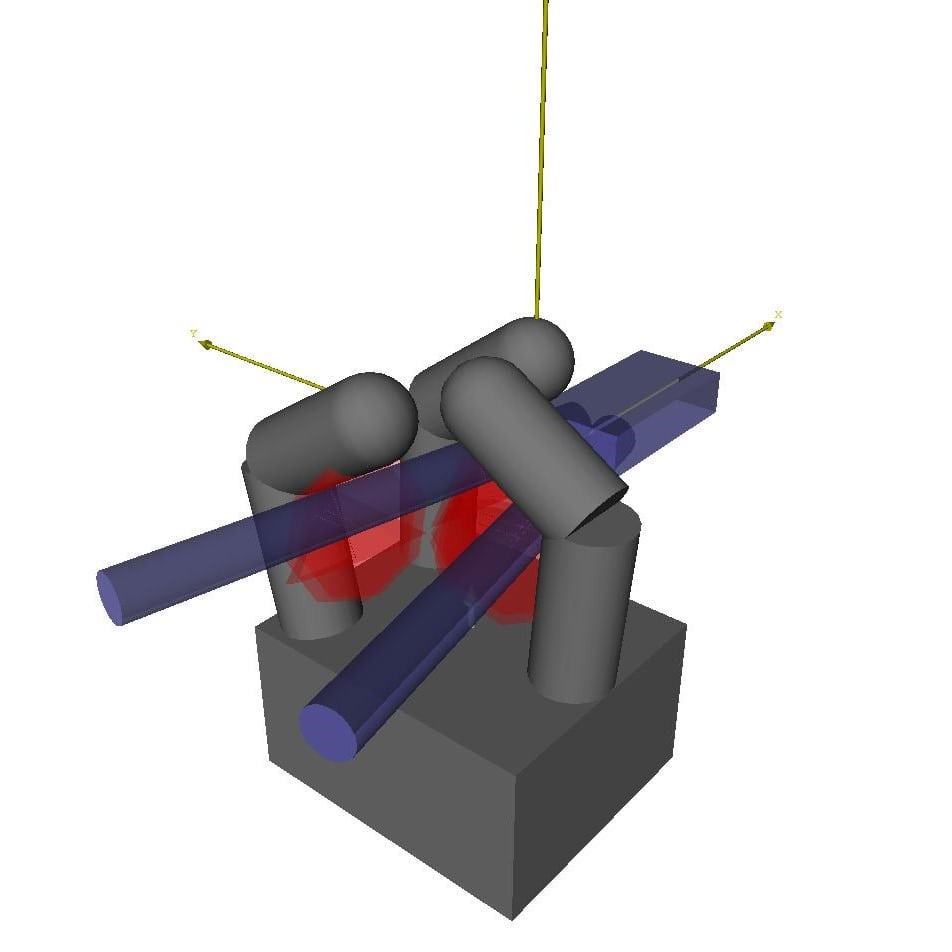}
&\hspace{-10mm} \includegraphics[width=0.22\linewidth]{images/des_grasp12.jpg}
 \vspace{-1mm}
\\
 \hspace{-5mm} \includegraphics[width=0.22\linewidth]{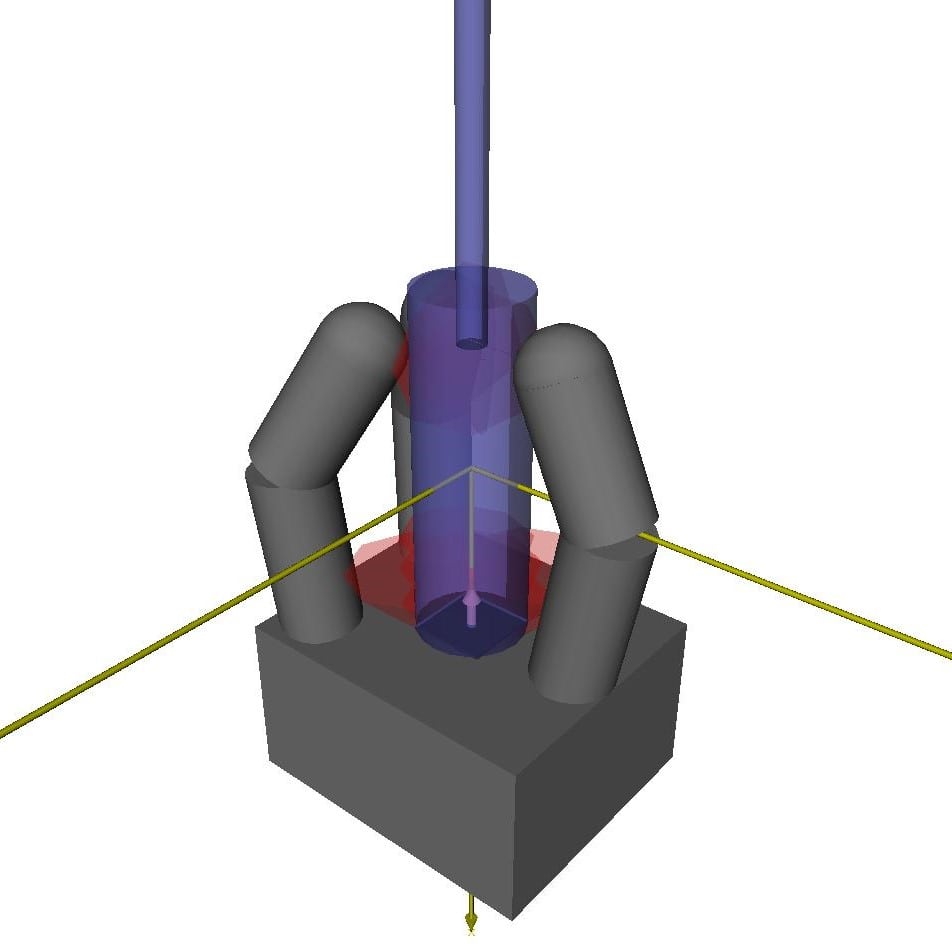}
&\hspace{-10mm} \includegraphics[width=0.22\linewidth]{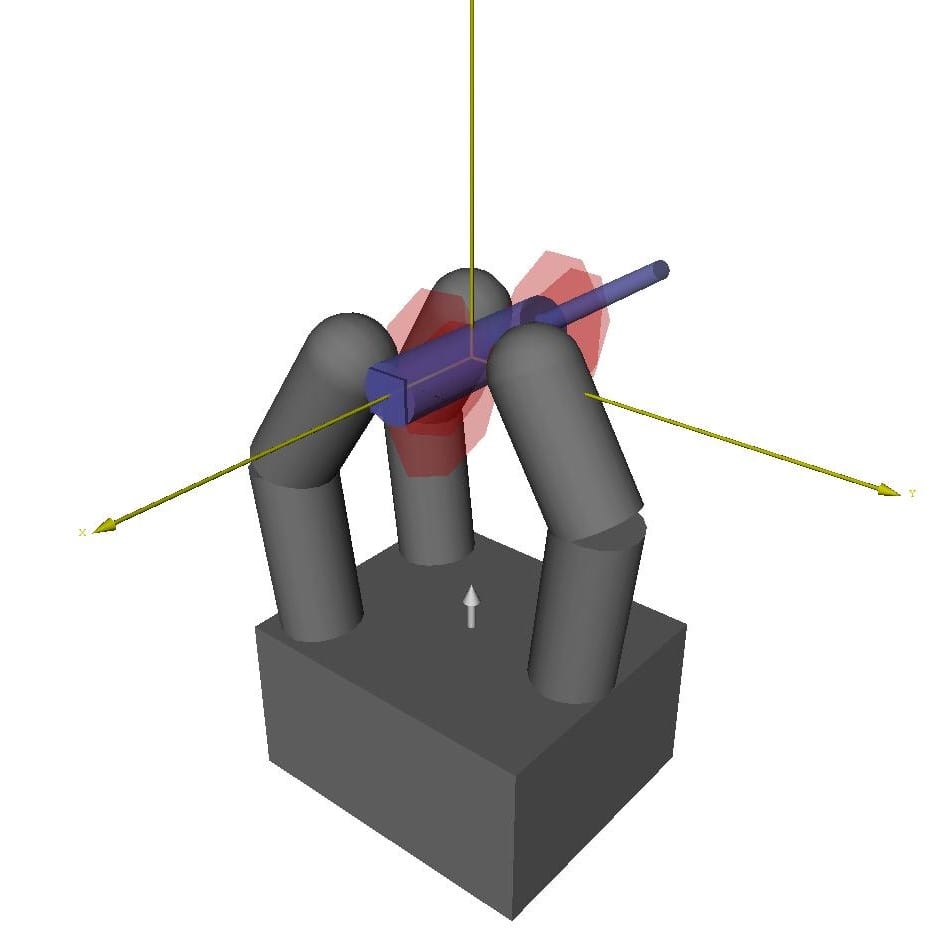}
&\hspace{-10mm} \includegraphics[width=0.22\linewidth]{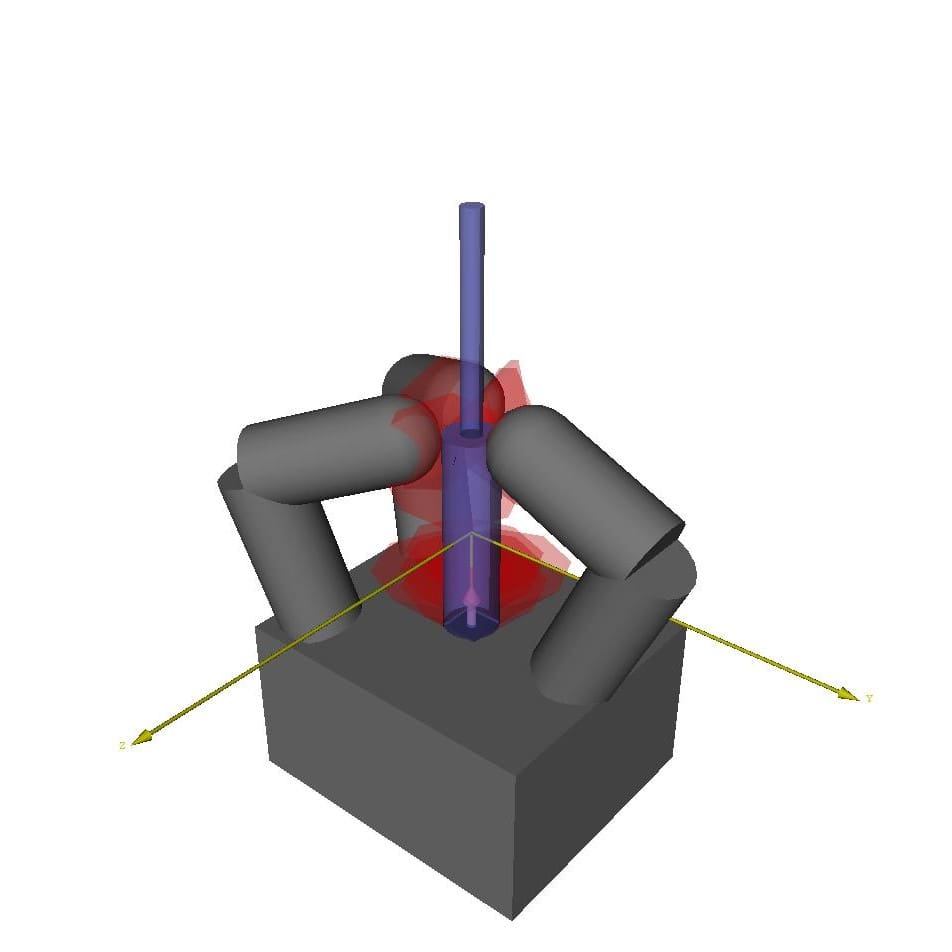}
&\hspace{-10mm} \includegraphics[width=0.22\linewidth]{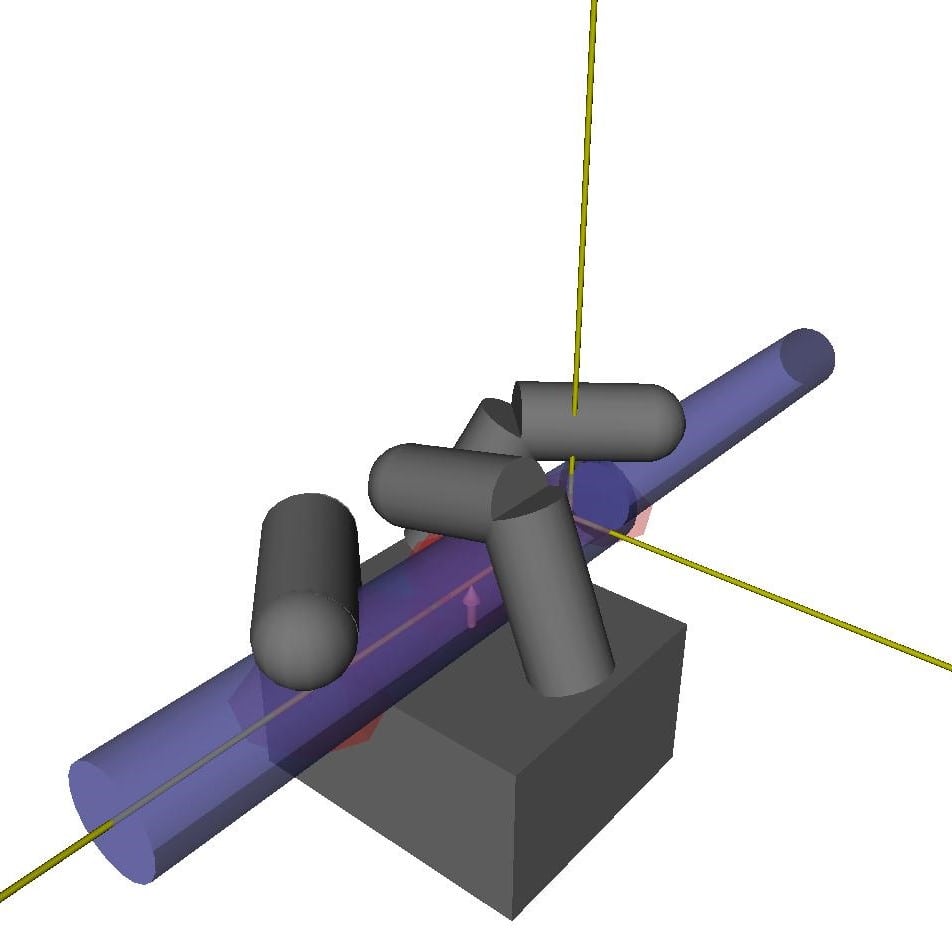}
&\hspace{-10mm} \includegraphics[width=0.22\linewidth]{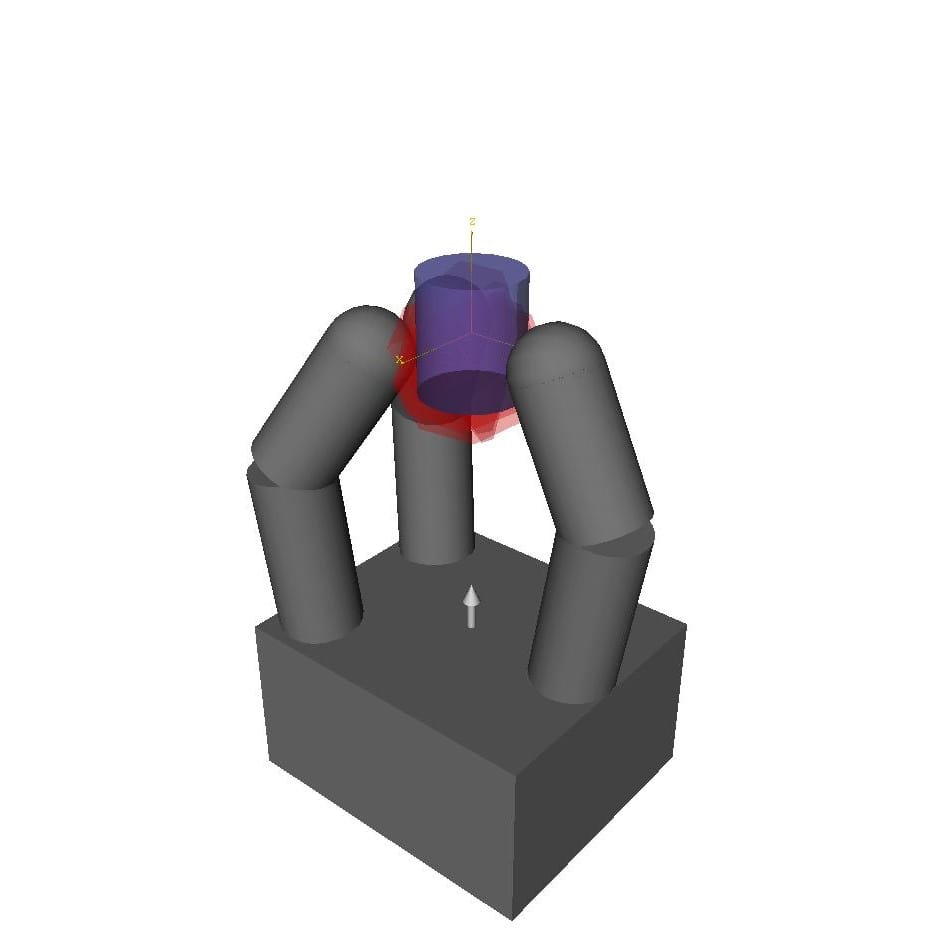}
&\hspace{-10mm} \includegraphics[width=0.22\linewidth]{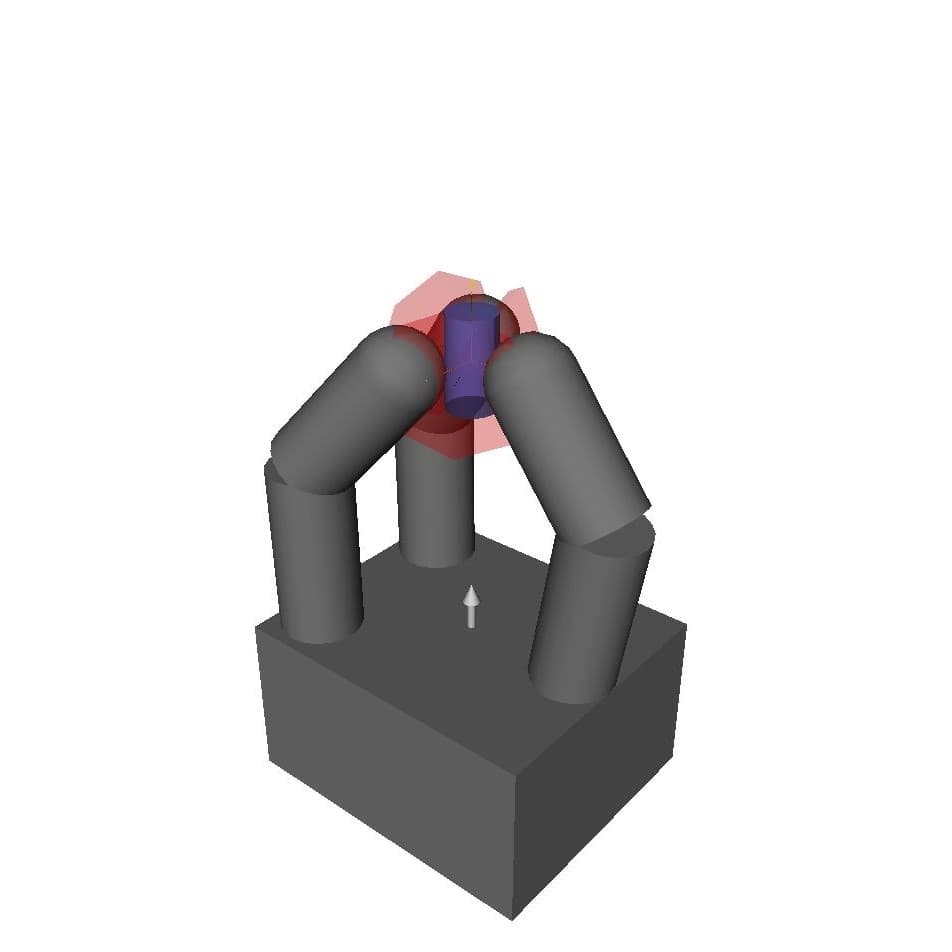}
 \vspace{-1mm}
\\
 \hspace{-5mm} \includegraphics[width=0.22\linewidth]{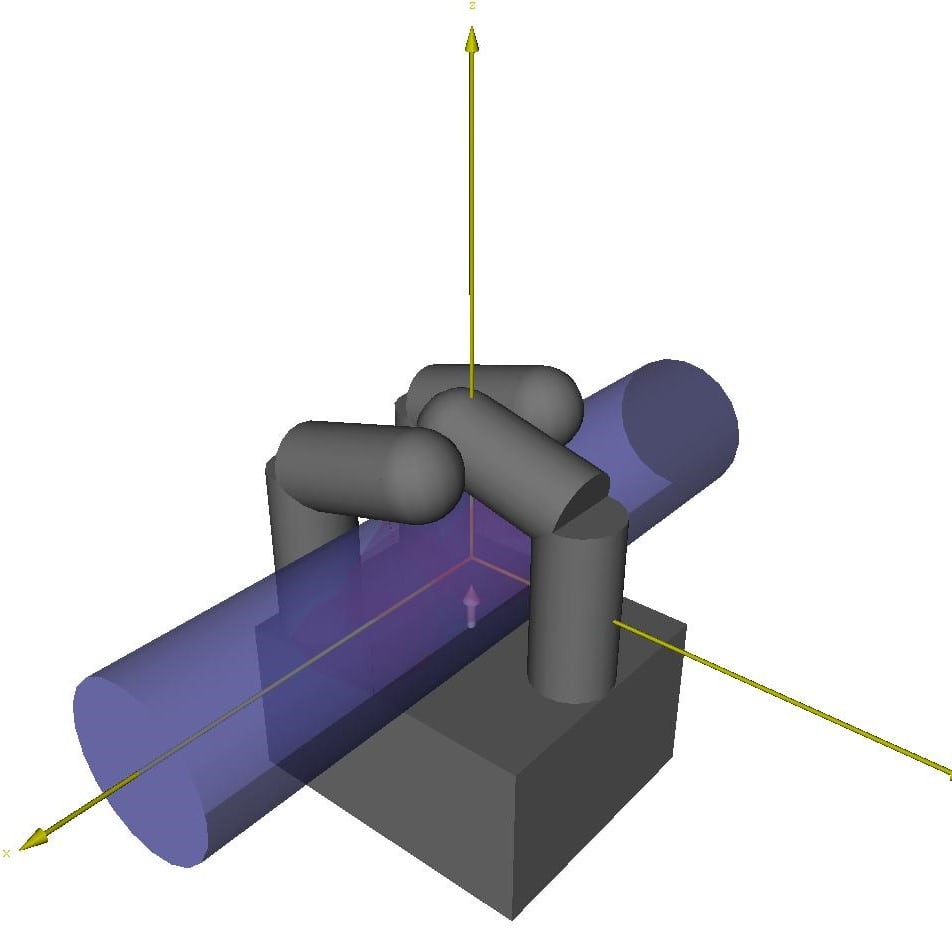}
&\hspace{-10mm} \includegraphics[width=0.22\linewidth]{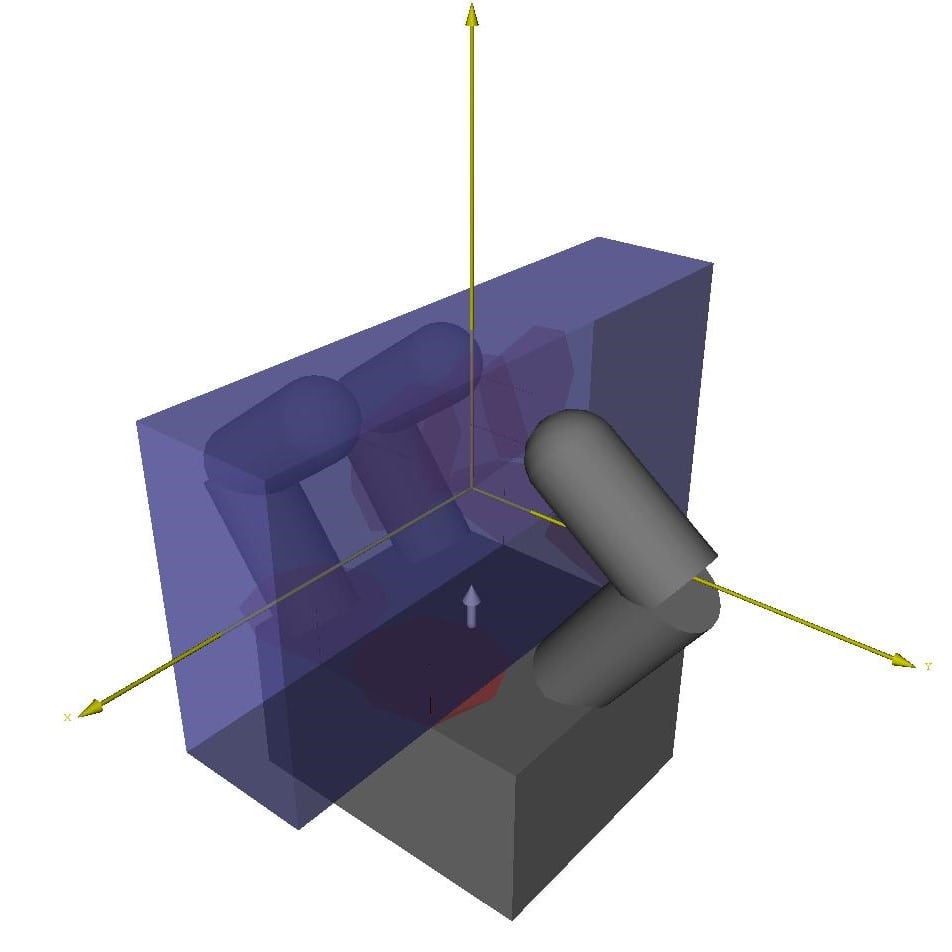}
&\hspace{-10mm} \includegraphics[width=0.22\linewidth]{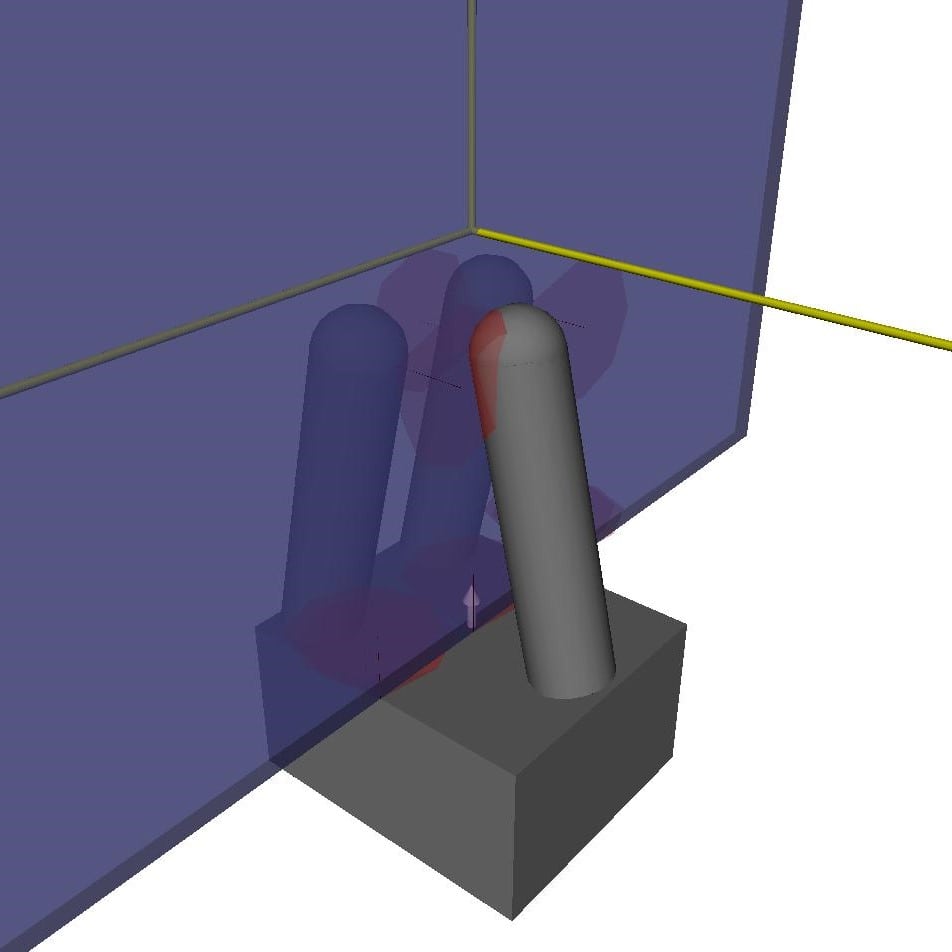}
&\hspace{-10mm} \includegraphics[width=0.22\linewidth]{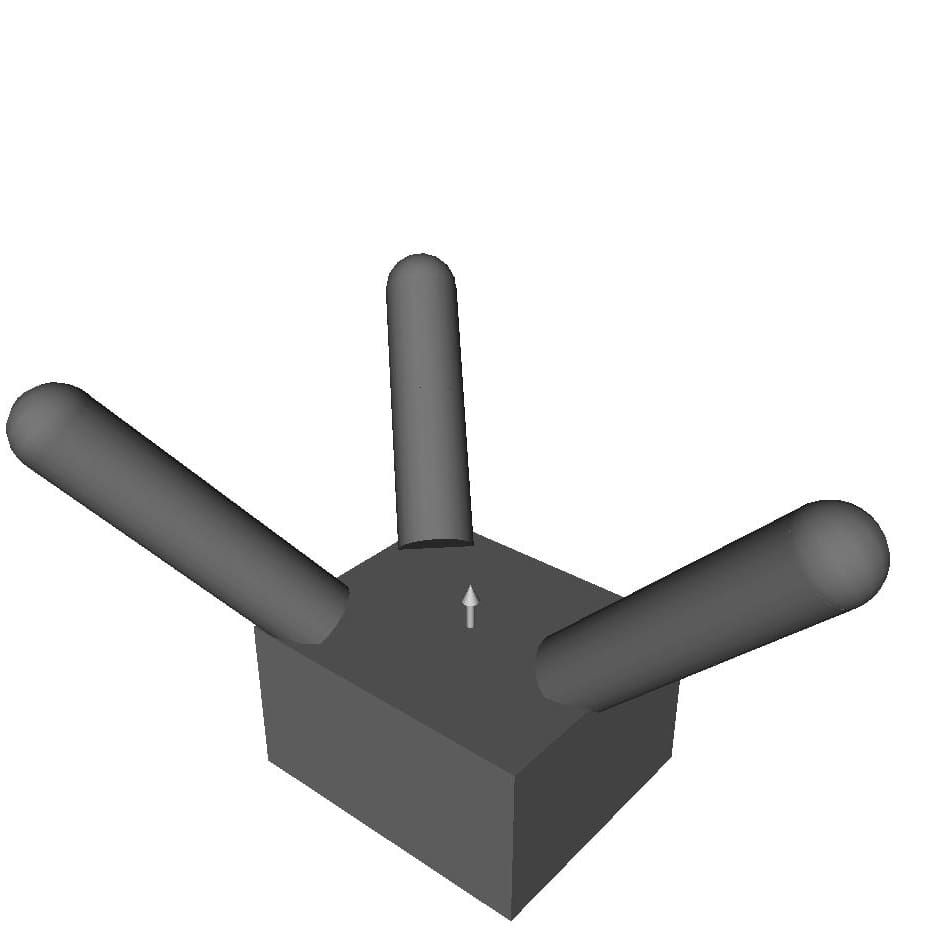}
\\
\end{tabular}
\caption{All desired grasps (the first 21 images) and \addedtext{1-4-2}{the opening configuration (the last one)} created in \textit{GraspIt!} simulator for Design Case I. The hands are shown in grey, the objects are shown in blue, and contacts (friction cones) are shown in red. All grasps do not consider underactuation and have force-closure property.}
\label{fig:des_grasps}
\end{figure}

The hand models are created in \textit{GraspIt!} Simulator \cite{miller2004graspit} without considering underactuation, and 21 desired grasps are created for 15 commonly-used ISS objects selected \remindtext{1-14 \& O3-4}{according to a case study with a domain expert, mainly including food (such as cans) and tools (such as screwdrivers)}. All grasps have force-closure property, checked with the Ferrari-Canny  $\epsilon$-metric \cite{ferrari1992planning} by $\epsilon > 0$. \addedtext{3-5}{The grasp types are determined empirically, and if multiple types of grasps as possible for a certain object they are all included}. All of the desired grasps for Design Case I are shown in Fig. \ref{fig:des_grasps}. The mechanical limits are $-\pi/4$ to $\pi/4$ radians for roll joints and proximal joints, and $0$ to $\pi/2$ radians for distal joints, defining the zero-position as the pose in which all links are perpendicular to the palm and all proximal and distal joint axes are parallel.

\newaddedtext{2-4}{The ``raw'' grasp data contains the joint angles, the contact locations and the friction coefficients. Then we can calculate the grasp Jacobian $\bm{J}$, the grasp map $\bm{G}$, the friction constraint matrix $\bm{F}$ and the friction pyramid generator $\bm{D}$ in (\ref{eq:hand_eq}) -- (\ref{eq:beta_bounds}), and pass them to the following optimization steps.}

\subsubsection{Optimization of the Mechanically Realizable Torque Manifold}

In this step, we optimize the joint pulley radii (tendon moment arms) $r_{tp}$, $r_{td}$, $r_{fr}$, $r_{fp}$, $r_{fd}$, where the subscripts $t$ and $f$ represent thumb and finger, and $r$, $p$, and $d$ represent the roll, proximal and distal joints (\remindtext{1-4}{we consider the two fingers to be mirrored versions of each other}). These parameters are illustrated in Fig. \ref{fig:joint_params} (a).

The actuation scheme we designed is also shown in Fig. \ref{fig:hand_and_actuation1}, where each finger is actuated by one tendon, and all tendons are rigidly connected to the actuator. We note that the finger tendons wrap around the roll joint pulleys, after which the plane of routing rotates 90 degrees and the tendons travel to the proximal and distal joints in the fingers. 

The vector of generated net joint torque in (\ref{eq:actuation}) $\bm{t}_{net}^{gen} \in \mathbb{R}^8$, the vector $\bm{t}_{net} \in \mathbb{R}^3$ (each element represents the tension on one tendon), and the Actuation Matrix has the specific form of
\begin{equation} \label{eq:actuation_matrix1}
\bm{A} = \left[ \begin{smallmatrix}
r_{tp} & & \\
r_{td} & & \\
 &-r_{fr} & \\
 &r_{fp} & \\
 &r_{fd} & \\
 & &r_{fr} \\
 & &r_{fp} \\
 & &r_{fd} \\
\end{smallmatrix} \right]
\end{equation}

In global optimization (\ref{eq:force_optim_search}) - (\ref{eq:force_optim_save}), the range of pulley radii is set to 2 ~ 12 mm, such that the pulley is large enough to be manufacturable but small enough to be mounted into the joints. Using a \remindtext{1-11}{threshold of 0.1 (unitless normalized torque) for unbalanced torque}, one outlier grasp is excluded in this step. The convergence tolerance is set to $10^{-10}$ for inner layer QP and $10^{-6}$ for outer layer CMA-ES. The above conditions are set the same for this design case and Design Case II and III. We note again that in this step, there are non-unique solutions, so the pulley radii are not fully determined yet. The minimum objective function value is recorded for the next step.

The computation time on a commodity desktop computer (quad-core 3.4 GHz CPU) is 10 minutes, using the cvxopt (for QP) and pycma (for CMA-ES) packages.

\begin{table}[b!]
\caption{Optimized parameters of Design Case I (in mm, Nmm/rad, rad, respectively)} 
\label{table:parameters}
\vspace{-3mm}
\begin{center}
\begin{tabular}{c|ccccc}
Parameter&$r_{tp}$ & $r_{td}$ & $r_{fr}$ & $r_{fp}$ & $r_{fd}$\\
\hline
\\[-2.5mm]
Value&12.0 & 4.6 & 2.0 & 11.8 & 4.5\\
\end{tabular}
\end{center}

\begin{center}
\begin{tabular}{c|ccccc}
Parameter&$K_{tp}$ & $K_{td}$ & $K_{fr}$ & $K_{fp}$ & $K_{fd}$\\
\hline
\\[-2.5mm]
Value&5.94 & 2.25 & 3.60 & 19.25 & 7.56\\
\end{tabular}
\end{center}

\begin{center}
\begin{tabular}{c|cccccc}
Parameter&$\theta_{0tp}$ & $\theta_{0td}$ & $\theta_{0fr}$ & $\theta_{0fp}$ & $\theta_{0fd}$\\
\hline
\\[-2.5mm]
Value&4.71 & 3.93 & 4.34 & 4.71 & 3.78\\
\end{tabular}
\end{center}
\end{table}

\subsubsection{Optimization of the Mechanically Realizable Posture Manifold for Inter-tendon Kinematic Behaviors}

As discussed in the subsection \ref{sec:method:inter}, we need to pick a unique solution of pulley radii ($r_{tp}$, $r_{td}$, $r_{fr}$, $r_{fp}$, $r_{fd}$) from the non-unique solutions from the previous step. The goal is to minimize the differences of actual and required tendon travel for different tendons. 

The motor-tendon connection matrix $\bm{M}$ in (\ref{eq:tendon_travel}) has a specific form of (\ref{eq:motor_tendon_connection_matrix1}), and $r_{mot}$ is the motor pulley radius.
\begin{equation} \label{eq:motor_tendon_connection_matrix1}
\bm{M} = [r_{mot}, r_{mot}, r_{mot}]^T
\end{equation}

Besides, the tendon travel vector $\bm{s}$ has a specific form of (\ref{eq:tendon_travel_vector}), where $\bm{\theta}$ is the vector of joint angles in a desired grasp configuration.
\begin{equation} \label{eq:tendon_travel_vector}
\bm{s} = \bm{A}^T\bm{\theta}
\end{equation}

In the outer layer, the candidate solutions need to satisfy the constraint that the previous optimization takes its minimal value with a tolerance of $10^{-3}$, and the CMA-ES algorithm finds the optimal set of pulley radii with function value convergence tolerance of $10^{-3}$. No outliers are found in this step \remindtext{1-11}{using a threshold length error of 2 mm.} 

The resulting optimal pulley radii are shown in table \ref{table:parameters}. The computation time is 30 minutes.

\subsubsection{Optimization of the Mechanically Realizable Posture Manifold for Intra-tendon Kinematic Behaviors}
In this step, we optimize spring stiffnesses $K_{tp}$, $K_{td}$, $K_{fr}$, $K_{fp}$, $K_{fd}$, and spring preload angles $\theta_{0tp}$, $\theta_{0td}$, $\theta_{0fr}$, $\theta_{0fp}$, $\theta_{0fd}$, where the subscripts have the same meaning as previous. These parameters are also illustrated in Fig. \ref{fig:joint_params}.

In practice, since the pre-contact kinematic behaviors of each finger is independent of every other one, we can search for each finger separately, and thus reduce the search dimensionality.

The spring stiffnesses are limited by the physical dimensions allowed in the mounting area in the joints. Also, they can only take discrete numbers offered by the manufacturer. In this design, the available stiffnesses are 10 discrete values from 2.25 to 19.25 Nmm/rad.

In addition, the spring preload angles are limited between the maximum allowed torsional angles by the datasheet and the minimum torsion angles to provide enough restoring torques over the entire range of motion. In this design case, the preload angles range from $\pi/4$ to $7\pi/4$ radians for the roll joints, $\pi/4$ to $3\pi/2$ radians for the proximal joints, and $0$ to $3\pi/2$ radians for the distal joints. 

In this step, there are three outlier grasps excluded \remindtext{1-11}{using a threshold unbalanced spring torque of 2 Nmm for the thumb and 5 Nmm for fingers.}

The optimal spring parameters are shown in table \ref{table:parameters}. The computation time is 5 minutes.

\begin{table}[b]
\caption{Optimized parameters of Design case II (in mm, Nmm/rad, rad, respectively)} 
\label{table:parameters2}
\vspace{-3mm}
\begin{center}
\begin{tabular}{c|ccccc}
Parameter&$r_{tp}$ & $r_{td}$ & $r_{fr}$ & $r_{fp}$ & $r_{fd}$\\
\hline
\\[-2.5mm]
Value& 12.0 & 4.5 & arbitrary & 12.0 & 4.5\\
\end{tabular}
\end{center}

\begin{center}
\begin{tabular}{c|cccc}
Parameter&$K_{tp}$ & $K_{td}$ & $K_{fp}$ & $K_{fd}$\\
\hline
\\[-2.5mm]
Value& 5.94 & 2.25 & 5.94 & 2.25\\
\end{tabular}
\end{center}

\begin{center}
\begin{tabular}{c|cccc}
Parameter&$\theta_{0tp}$ & $\theta_{0td}$ & $\theta_{0fp}$ & $\theta_{0fd}$\\
\hline
\\[-2.5mm]
Value& 4.71 & 3.86 & 4.71 & 3.82\\
\end{tabular}
\end{center}
\end{table}

\begin{figure}[t!]
\centering
\begin{tabular}{cc}
\includegraphics[width=0.5\linewidth]{images/hand_dim1.pdf} \hspace{-6mm} & \includegraphics[width=0.5\linewidth]{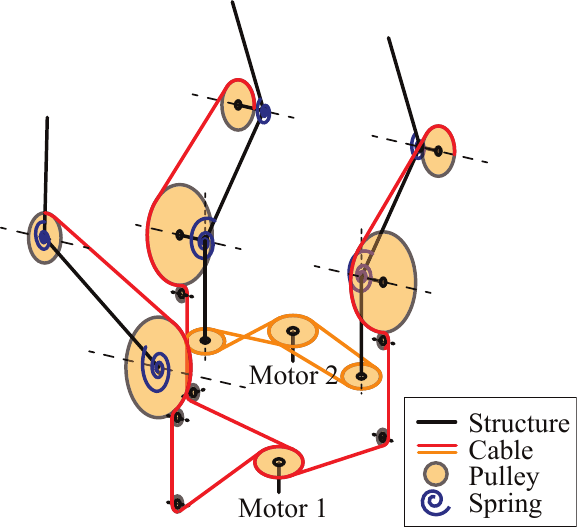} \\
(a) & (b)
\end{tabular}
\caption{(a) Hand model with pre-defined kinematics and (b) actuation scheme of Design Case II}
\label{fig:hand_and_actuation2}
\end{figure}


\subsection{Design Case II: Dual-motor Hand with Roll-pitch Fingers}
In the previous design, all eight joints are actuated by a single motor, but it is interesting to see the benefits of having an additional motor controlling part of the hand motion separately. Therefore, we propose a variation of the previous design: a dual-motor hand with roll-pitch finger configuration, where one motor is in charge of the flexion of all fingers, and the other motor is in charge of the finger rolling. The hand kinematic configuration is shown in Fig. \ref{fig:hand_and_actuation2}.

Here, the tendon tension vector in (\ref{eq:actuation}) $\bm{t}_{net} \in \mathbb{R}^5$ (the first three elements are the tensions on three tendons going to the thumb and fingers, the last two are the forces on the roll transmission connected to the second motor). The Actuation Matrix $\bm{A}$ in (\ref{eq:actuation}) has the specific form of
\begin{equation} \label{eq:actuation_matrix2}
\bm{A} = \left[ \begin{smallmatrix}
r_{tp} & & & & \\
r_{td} & & & & \\
 & & & -r_{fr}&\\
 & r_{fp} & & &\\
 & r_{fd} & & &\\
 & & & & r_{fr}\\
 & & r_{fp} & &\\
 & & r_{fd} & &\\
\end{smallmatrix} \right]
\end{equation}
The motor-tendon connection matrix $M$ in (\ref{eq:tendon_travel}) has a specific form of
\begin{equation} \label{eq:motor_tendon_connection_matrix2}
\bm{M} = \left[ \begin{smallmatrix}
r_{mot1} & \\
r_{mot1} & \\
r_{mot1} & \\
& r_{mot2} \\
& r_{mot2} \\
\end{smallmatrix} \right]
\end{equation}
and the tendon travel vector $\bm{s}$ in (\ref{eq:tendon_travel}) is also $\bm{A}^T\bm{\theta}$, where $\bm{\theta}$ is the vector of joint angles in a desired grasp configuration.

All design details are the same as in Design Case I, with the exception that the roll joints do not have springs, and thus not having associated spring parameters. There are altogether four grasps excluded using the same criteria as in Design Case I. The optimal pulley radii and spring parameters are shown in table \ref{table:parameters2}.

\begin{figure}[t!]
\centering
\begin{tabular}{cc}
\includegraphics[width=0.5\linewidth]{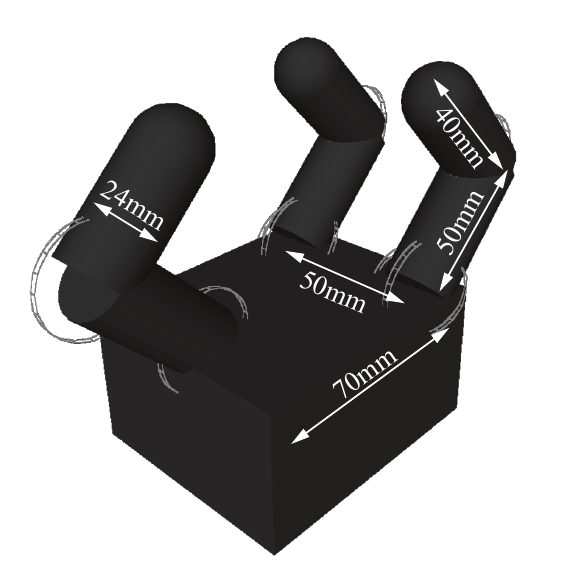} \hspace{-6mm} & \includegraphics[width=0.5\linewidth]{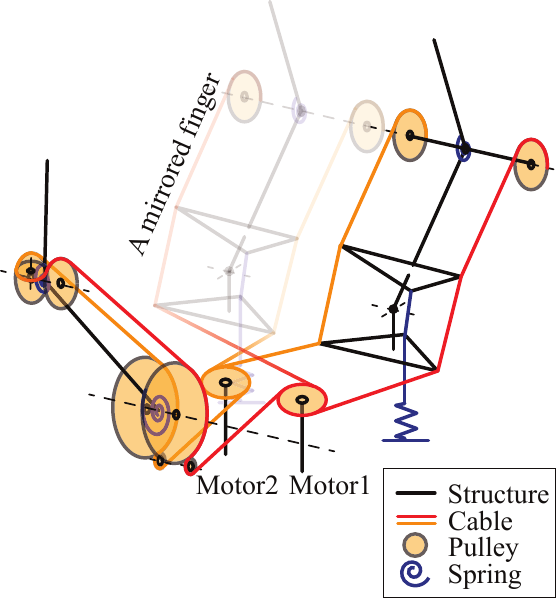} \\
(a) & (b)
\end{tabular}
\caption{(a) Hand model with pre-defined kinematics and (b) actuation scheme of Design Case III}
\label{fig:hand_and_actuation3}
\end{figure}

\begin{table}[b!]
\caption{Optimized parameters of Design case III (in mm, Nmm/rad, rad or mm, respectively)} 
\label{table:parameters3}
\vspace{-3mm}
\begin{center}
\begin{tabular}{c|ccccc}
Parameter &$r_{tp}$ & $r_{td}$ & $h_{fp}$ & $r_{fp}$ & $r_{fd}$\\
\hline
\\[-2.5mm]
Value &4.65 & 2.00 & 6.29 & 12.00 & 2.00\\
\end{tabular}
\end{center}

\begin{center}
\begin{tabular}{c|cccc}
Parameter&$K_{tp}$ & $K_{td}$ & $K_{fp}$ & $K_{fd}$\\
\hline
\\[-2.5mm]
Value& 5.94 & 2.25 & 0.18 & 19.25\\
\end{tabular}
\end{center}

\begin{center}
\begin{tabular}{c|cccc}
Parameter&$\theta_{0tp}$ & $\theta_{0td}$ & $l_{0fp}$ & $\theta_{0fd}$\\
\hline
\\[-2.5mm]
Value& 4.71 & 4.45 & 16.67 & 0.23\\
\end{tabular}
\end{center}
\end{table}

\begin{table*}[t!]
\caption{Optimization metrics (in unitless torque, mm, Nmm for column 1, 2 and 3)} 
\label{table:metrics}
\vspace{-3mm}
\begin{center}
\begin{tabular}{c|ccc}
Optimization of: &Mechanically Realizable   & Mechanically Realizable  & Mechanically Realizable \\
                    &Torque Manifold &Posture Manifold &Posture Manifold \\
                    &               &(inter-tendon)     &(intra-tendon) \\
\hline
\\[-2.5mm]

Design Case I & 0.13  & 5.82  & 8.22  \\
               &(1 grasp excluded) &(0 grasps excluded) &(3 grasps excluded) \\
\hline
Design Case II & 0.08  & 3.05 & 2.82  \\
               &(1 grasp excluded) &(0 grasps excluded) &(3 grasps excluded) \\
\hline

Design Case III & 0.33 & 6.02 & 14.06 \\
               &(0 grasps excluded) &(0 grasps excluded) &(14 grasps excluded) \\
\hline

\end{tabular}
\end{center}

\end{table*}

\begin{figure*}[t!]
\centering
\begin{tabular}{cccc}
&  \hspace{5mm} Design Case I & \hspace{5mm} Design Case II & \hspace{5mm} Design Case III \\

\rotatebox{90}{\hspace{17mm} Thumb} & \includegraphics[width=0.33\linewidth]{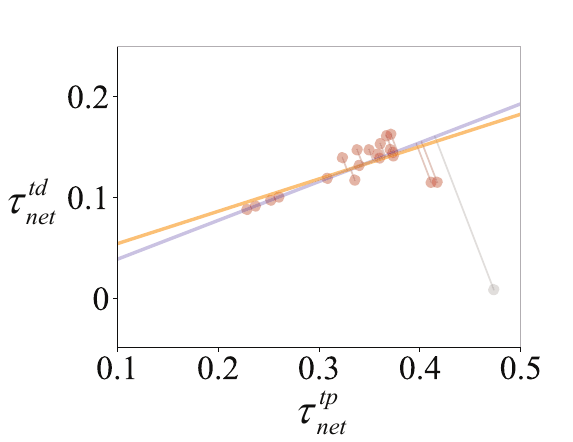} \hspace{-7mm} & \includegraphics[width=0.33\linewidth]{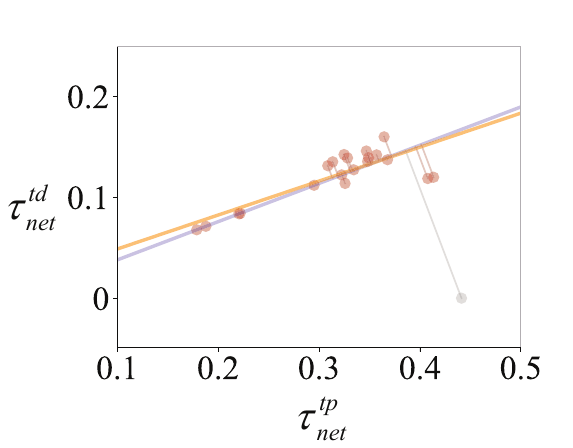} \hspace{-7mm} & \includegraphics[width=0.33\linewidth]{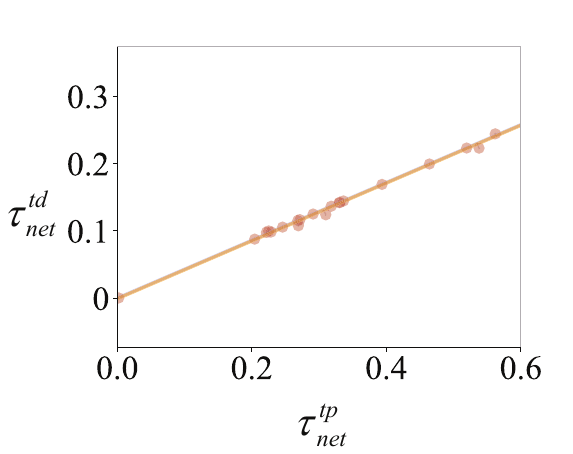} \\

& \hspace{8mm}(a) & \hspace{8mm}(b) & \hspace{8mm}(c) \\

\rotatebox{90}{\hspace{17mm} Finger} & \includegraphics[width=0.28\linewidth]{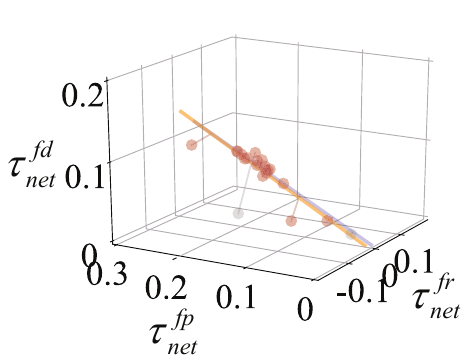} \hspace{-7mm} & \includegraphics[width=0.28\linewidth]{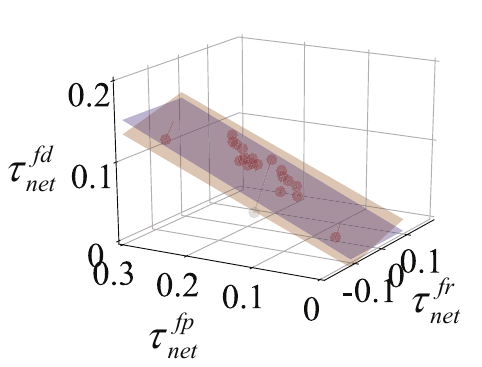} \hspace{-7mm} & \includegraphics[width=0.28\linewidth]{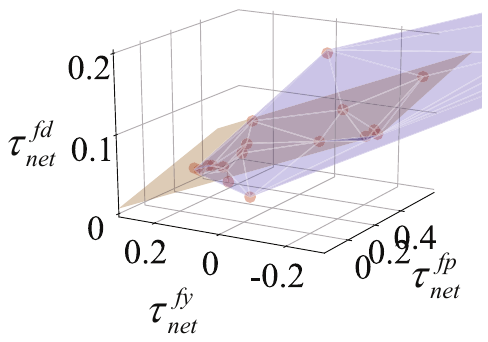} \\

& \hspace{8mm}(d) & \hspace{8mm}(e) & \hspace{8mm}(f) \\

& & \hspace{3mm} \includegraphics[width=0.3\linewidth]{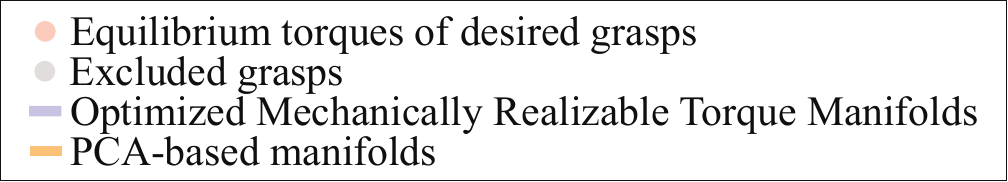} & \\

\end{tabular}

\caption{The visualization of the Mechanically Realizable Torque Manifolds. The first row shows 2D plots for the two joints in one thumb, and the second row shows 3D plots for the three joints in one finger. Different columns are different design cases. All torques are normalized unitless torques.}
\label{fig:mrtm_comparison}
\end{figure*}


\subsection{Design Case III: Dual-motor Hand with Pitch-yaw Fingers}
The third design case is an underactuated hand with pitch-yaw fingers. The pitch-yaw two-DoF proximal joint is realized by a universal joint with a three-tendon parallel mechanism. The back tendon of the joint is connected to a spring with preload. The front two tendons are actively controlled. Besides, they are not terminated in the proximal joint, but connected to distal joint pulleys. 

The two actively controlled tendons within a finger are connected to different motors. Meanwhile, the symmetric tendons on different fingers share a common motor. In the thumb, two tendons from the two motors are routed passing the proximal and distal joints and connected via an idler inside the fingertip. Fig. \ref{fig:hand_and_actuation3} illustrates the tendon routing scheme.

\begin{figure*}[t!]
\centering
\begin{tabular}{cccc}
&  \hspace{5mm} Design Case I & \hspace{5mm} Design Case II & \hspace{5mm} Design Case III \\

\rotatebox{90}{\hspace{17mm} Thumb} & \includegraphics[width=0.33\linewidth]{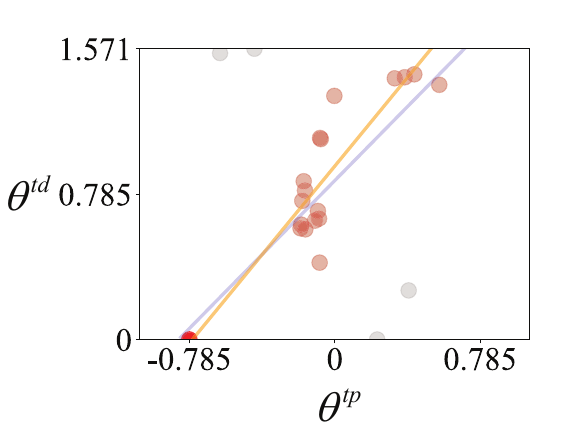} \hspace{-7mm} & \includegraphics[width=0.33\linewidth]{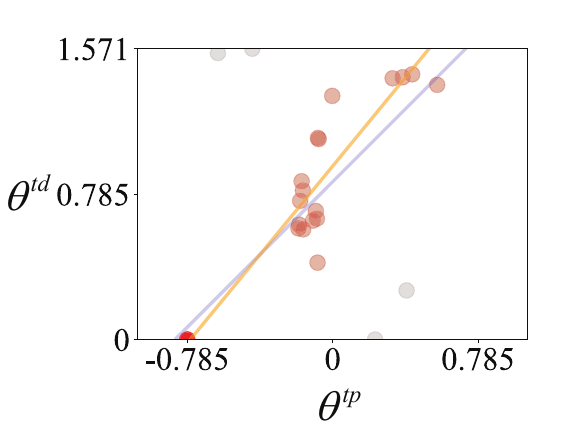} \hspace{-7mm} & \includegraphics[width=0.33\linewidth]{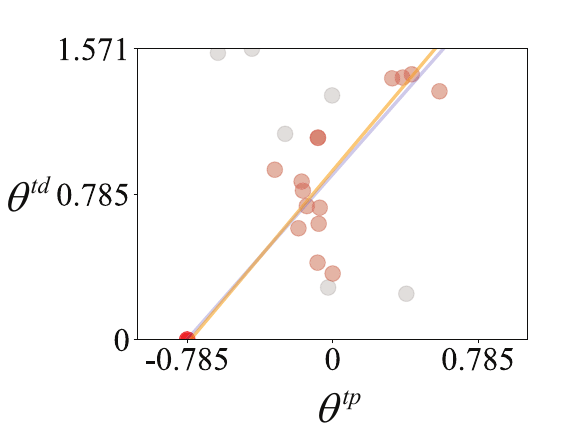} \\

& \hspace{8mm}(a) & \hspace{8mm}(b) & \hspace{8mm}(c) \\

\rotatebox{90}{\hspace{17mm} Finger} & \includegraphics[width=0.28\linewidth]{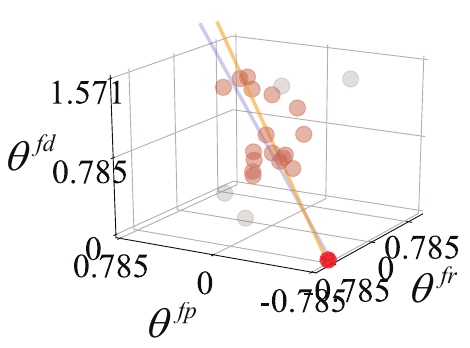} \hspace{-7mm} & \includegraphics[width=0.28\linewidth]{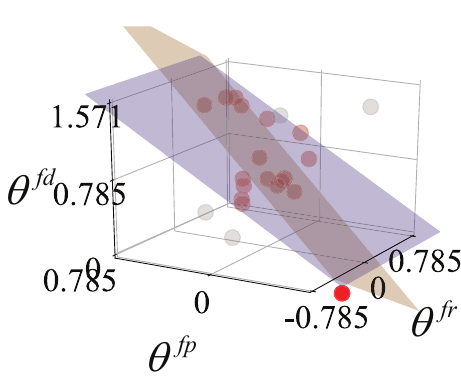} \hspace{-7mm} & \includegraphics[width=0.28\linewidth]{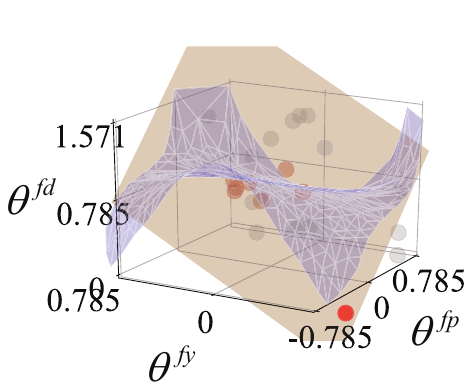} \\

& \hspace{8mm}(d) & \hspace{8mm}(e) & \hspace{8mm}(f) \\

& &\hspace{3mm} \includegraphics[width=0.3\linewidth]{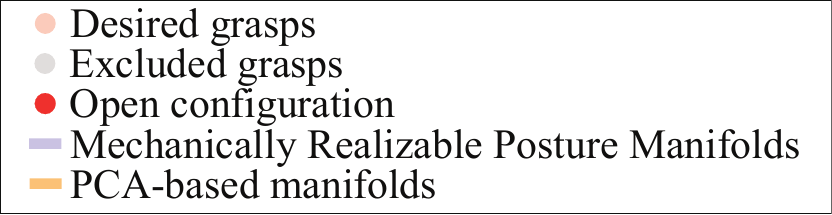} & \\

\end{tabular}

\caption{The visualization of the Mechanically Realizable Posture Manifolds. The first row shows 2D plots for the two joints in one thumb, and the second row shows 3D plots for the three joints in one finger. Different columns are different design cases. All angles are in radians.}
\label{fig:mrpm_comparison}
\end{figure*}

In this design case, the Actuation Matrix has the form of (\ref{eq:actuation_matrix3}), where $\rho$'s are the configuration-dependent moment arms of the tendon forces, which can be calculated via geometric relationships.  
The subscripts $p$ and $y$ mean pitch and yaw, the numbers in the subscripts are combinations of finger number and motor number.

\begin{equation} \label{eq:actuation_matrix3}
\bm{A} = \left[ 
\begin{smallmatrix}
&2r_{tp} & & & & \\
&2r_{td} & & & & \\
 & &\rho_{fpp11} &\rho_{fpp12} & & \\
 & &\rho_{fpy11} &\rho_{fpy12} & & \\
 & &r_{fd} &r_{rd} & & \\
 & & & &\rho_{fpp21} &\rho_{fpp22} \\
 & & & &\rho_{fpy21} &\rho_{fpy22} \\
 & & & &r_{fd} &r_{rd} \\
\end{smallmatrix} 
\right]
\end{equation}

The motor-tendon connection matrix $\bm{M}$ in (\ref{eq:tendon_travel}) has a specific form of
\begin{equation} \label{eq:motor_tendon_connection_matrix3}
\bm{M} = \left[
\begin{smallmatrix}
&r_{mot} &r_{mot} \\
&r_{mot} &0 \\
&0 &r_{mot} \\
&r_{mot} &0 \\
&0 &r_{mot} \\
\end{smallmatrix} 
\right]
\end{equation}

And the tendon travel vector $\bm{s}$ in (\ref{eq:tendon_travel}) is 
\begin{equation}
\label{eq:tendon_travel_vector3}
\bm{s} = \left[
\begin{smallmatrix}
&2(\theta_{tp}r_{tp} + \theta_{td}r_{td}) \\
&\theta_{fd}r_{fd} + \Delta l_{fp11} \\
&\theta_{fd}r_{fd} + \Delta l_{fp12} \\
&\theta_{fd}r_{fd} + \Delta l_{fp21} \\
&\theta_{fd}r_{fd} + \Delta l_{fp22} \\
\end{smallmatrix} 
\right]
\end{equation}
where the $\Delta l$'s are the tendon length changes between zero-configuration and grasp configuration. We present the details of the derivation of the above matrices in the Appendix.

There are 14 grasps excluded using the same criteria. The optimal design parameters are shown in Table \ref{table:parameters3}.


\subsection{Numerical Evaluation}
All the metrics provided by our optimization framework for all three design cases, and the number of excluded grasps in each design step within each case, are summarized in Table \ref{table:metrics}.

In addition to the numerical metrics, it can also be informative to visualize the Mechanically Realizable Torque Manifold and Posture Manifold. However, these manifolds are high-dimensional and cannot be visualized directly. Therefore, we plot the slices of the spaces: The thumb manifolds are shown in two-dimensional joint space and the finger manifolds are shown in three-dimensional joint space separately. In each plot, the dimensionality of the manifold equals the number of motors connected to that thumb or finger.

The Mechanically Realizable Torque Manifolds for all three design cases are shown as Fig. \ref{fig:mrtm_comparison}. Each column shows one design case, and each row shows the plots for a thumb or a finger. The red dots represent the considered grasps, while the gray ones represent the excluded ones. The blue lines or planes are the Mechanically Realizable Torque Manifolds, and the orange lines or planes are the least-square fittings of red dots based on PCA for comparison. We note that, for each fully-articulated desired grasp, there are infinitely many solutions for equilibrium torques; among these desired points in torque space, we chose to display on the plots (as a red or gray dot) the equilibrium torques \textit{closest} to the Mechanically Realizable Torque Manifold.

Similarly, the Mechanically Realizable Posture Manifolds for all the design cases are shown in Fig. \ref{fig:mrpm_comparison}. The red dots represent the considered grasps, while the gray ones represent the excluded grasps. The lines or surfaces in blue are the Mechanically Realizable Posture Manifolds, and the lines or planes in orange are the PCA-based manifolds.

\subsection{Construction of the Hands and Experimental Evaluations}
\begin{figure*}[t!]
\begin{tabular}{lcccc}
\multicolumn{5}{c}{\includegraphics[width=\linewidth]{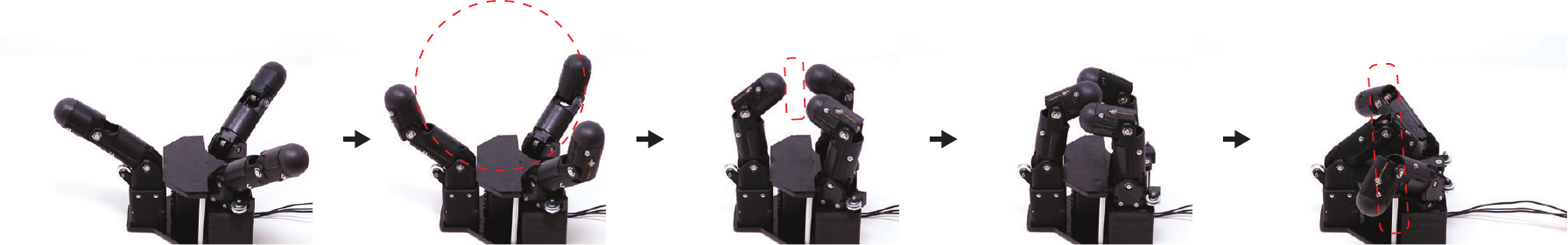}} \\
\hspace{10mm} (a) opening pose & \hspace{5mm} (b) spherical grasp & \hspace{3mm} (c) pinch grasp & \hspace{-3mm} (d) fingers passing each other & \hspace{-4mm} (e) enveloping grasp \\
\end{tabular}
\caption{Finger closing trajectory and the types of grasp along the trajectory of Design Case I. }
\label{fig:finger_traj}
\end{figure*}

\begin{figure}[t!]
\centering
\includegraphics[width=\linewidth]{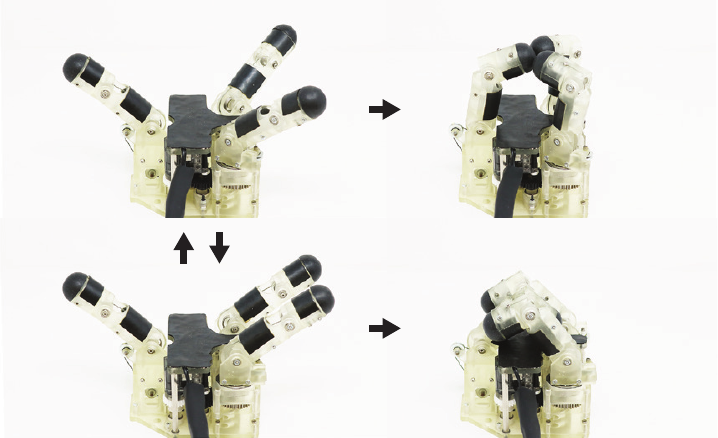}
\caption{Finger trajectory of Design Case II. The first row shows the non-parallel closing, resulting in a pinch grasp. The second row shows the parallel-finger closing, resulting in an enveloping grasp. Unlike in Design Case I, the finger roll angle can be actively controlled in Design Case II, which leads to the different finger trajectories.}
\label{fig:finger_traj_dual_motor}
\end{figure}

We constructed physical prototypes for Design Case I and II, shown in Fig. \ref{fig:finger_traj} and Fig. \ref{fig:finger_traj_dual_motor}.

\begin{figure}[t!]
\centering
\begin{tabular}{ccc}
\hspace{-5mm}\includegraphics[width=0.39\linewidth]{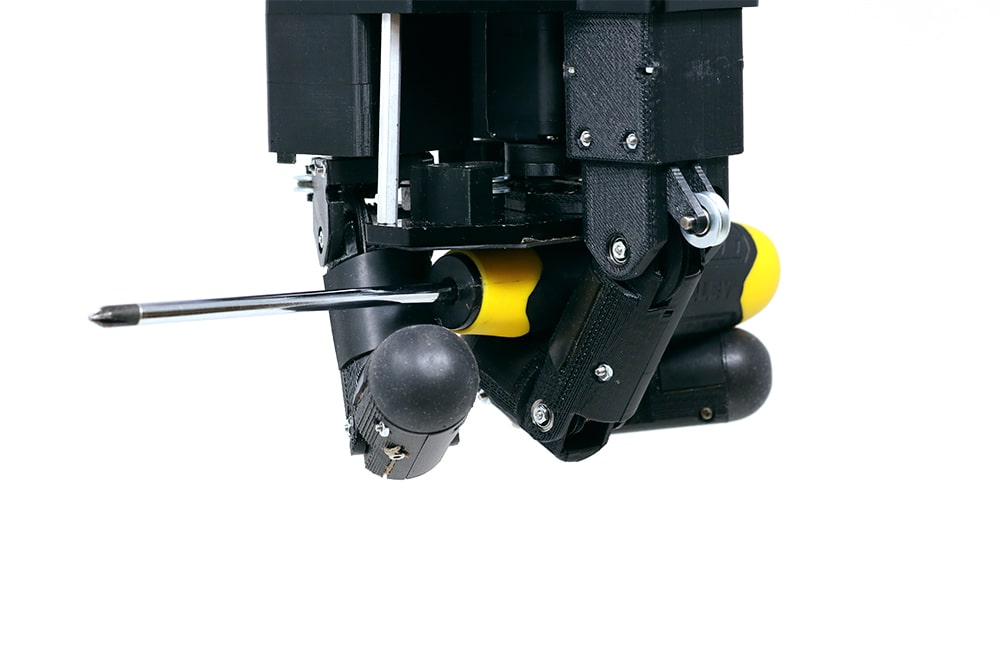}  & \hspace{-10mm} \includegraphics[width=0.39\linewidth]{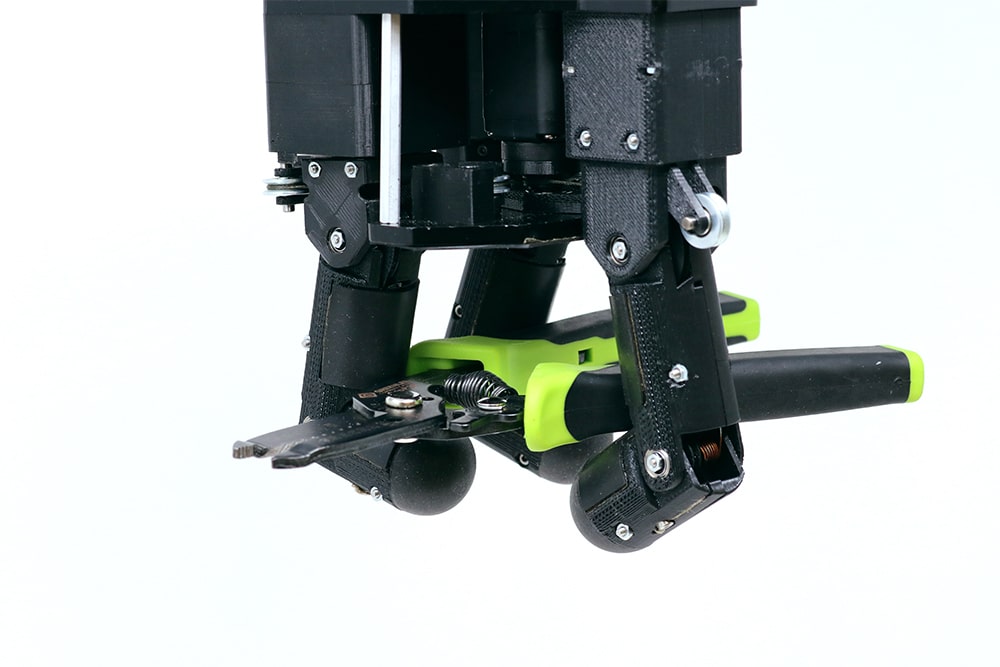}  & \hspace{-10mm} \includegraphics[width=0.39\linewidth]{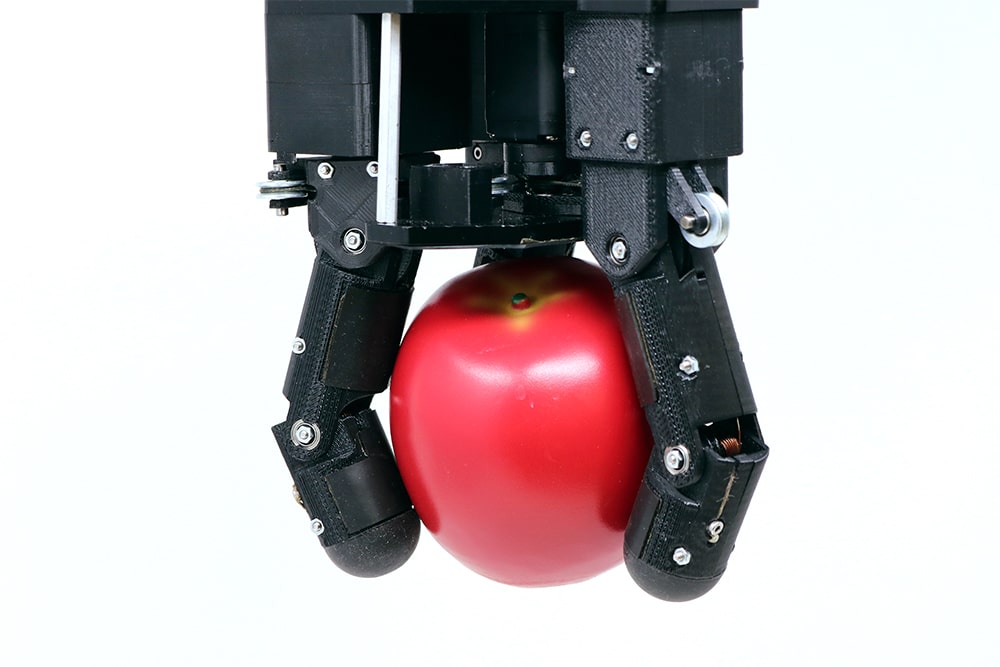}\\
\hspace{-5mm}
(a) Large screw driver & \hspace{-10mm}
(b) Pliers & \hspace{-10mm}
(c) Apple \vspace{2mm}\\
\hspace{-5mm}\includegraphics[width=0.39\linewidth]{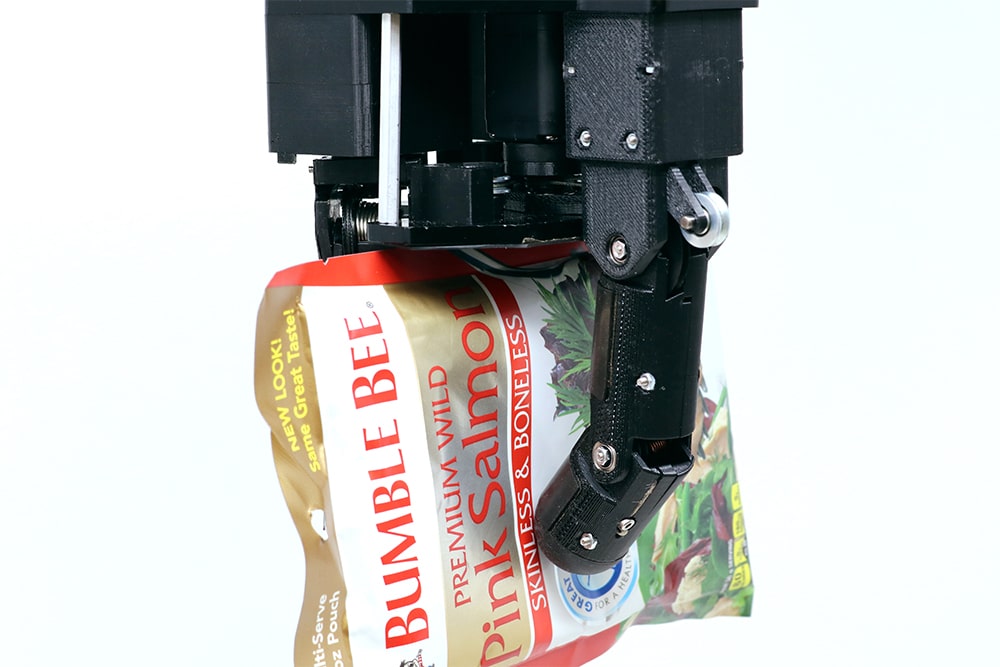}   & \hspace{-10mm} \includegraphics[width=0.39\linewidth]{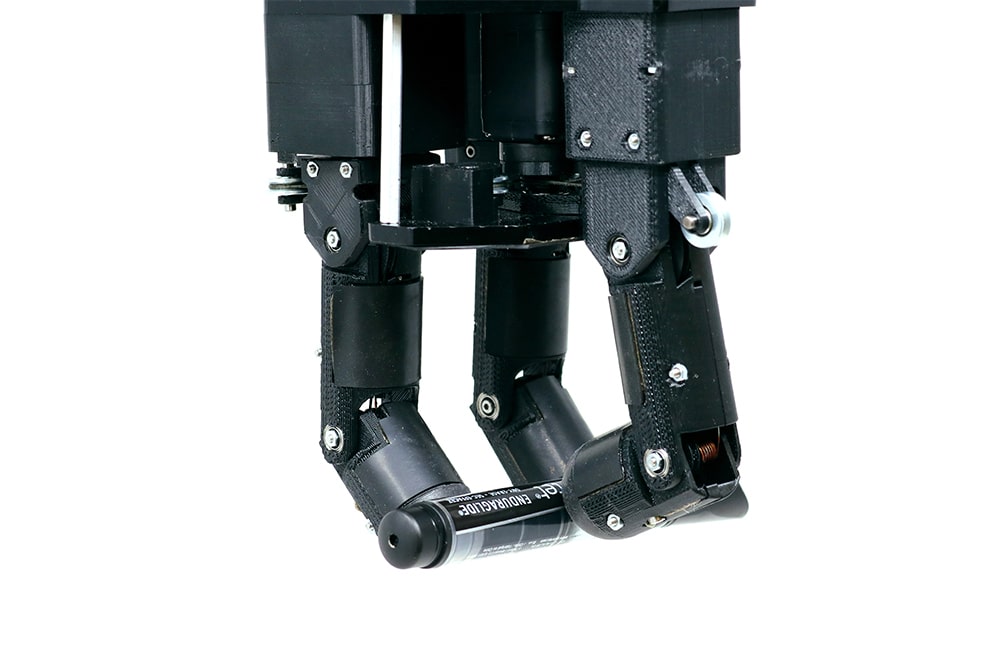}  & \hspace{-10mm} \includegraphics[width=0.39\linewidth]{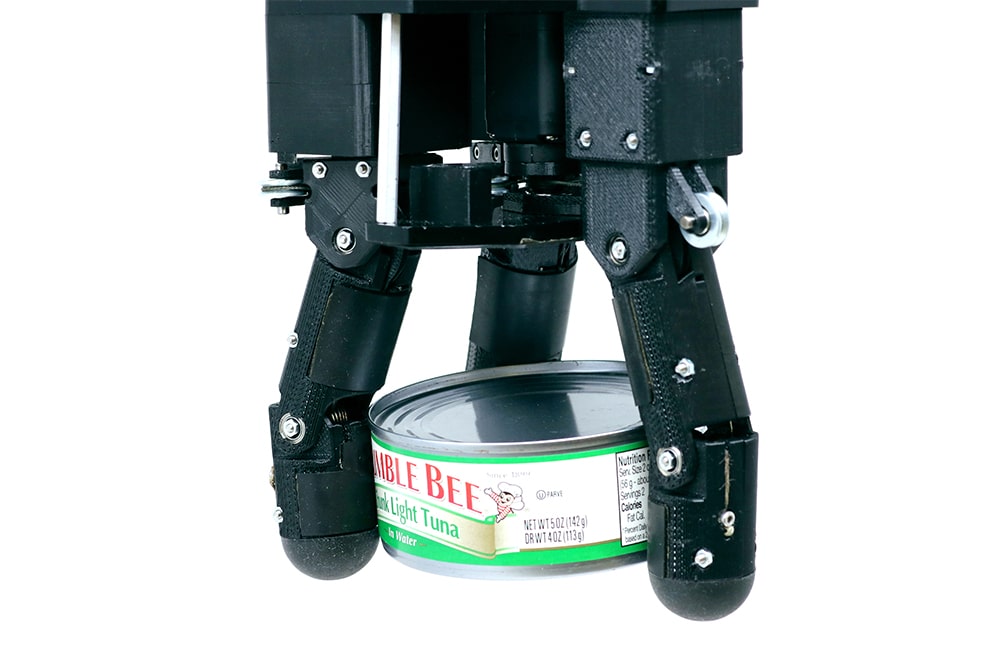}\\
\hspace{-5mm}
(d) Food bag & \hspace{-10mm}
(e) Pen & \hspace{-10mm}
(f) Food can \vspace{2mm}\\
\hspace{-5mm}\includegraphics[width=0.39\linewidth]{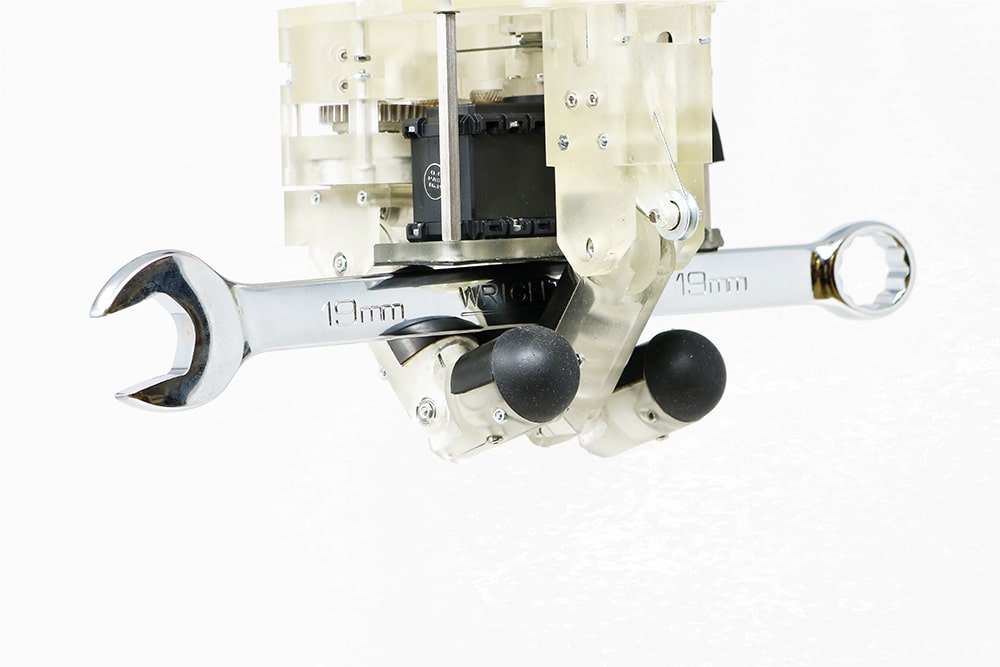}  & \hspace{-10mm} \includegraphics[width=0.39\linewidth]{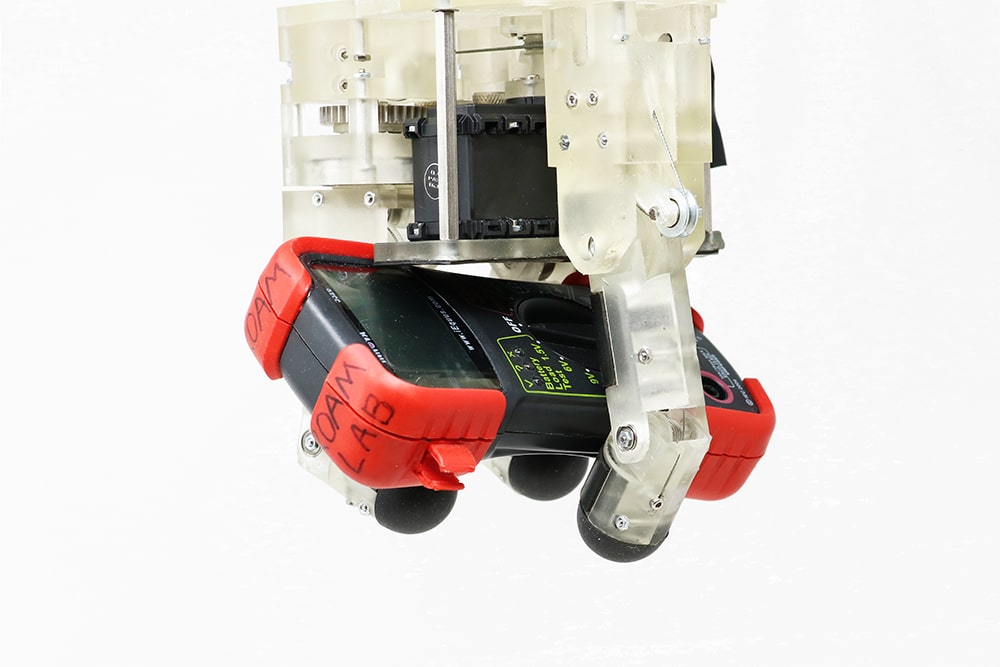}  & \hspace{-10mm} \includegraphics[width=0.39\linewidth]{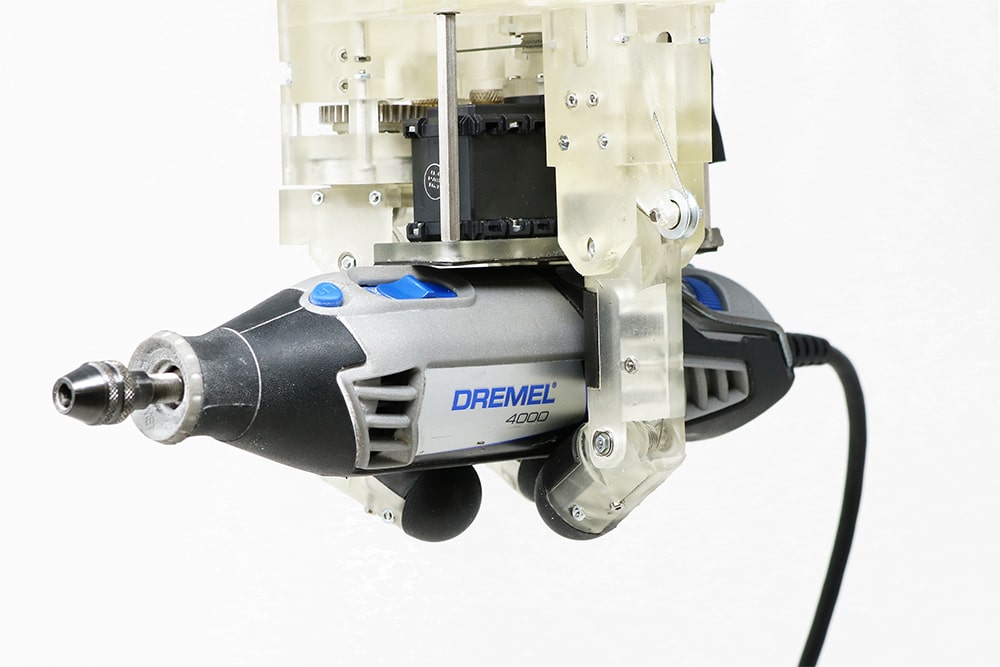}\\
\hspace{-5mm}
(g) Wrench & \hspace{-10mm}
(h) Multimeter & \hspace{-10mm}
(i) Dremel  \vspace{2mm}\\
\hspace{-5mm}\includegraphics[width=0.39\linewidth]{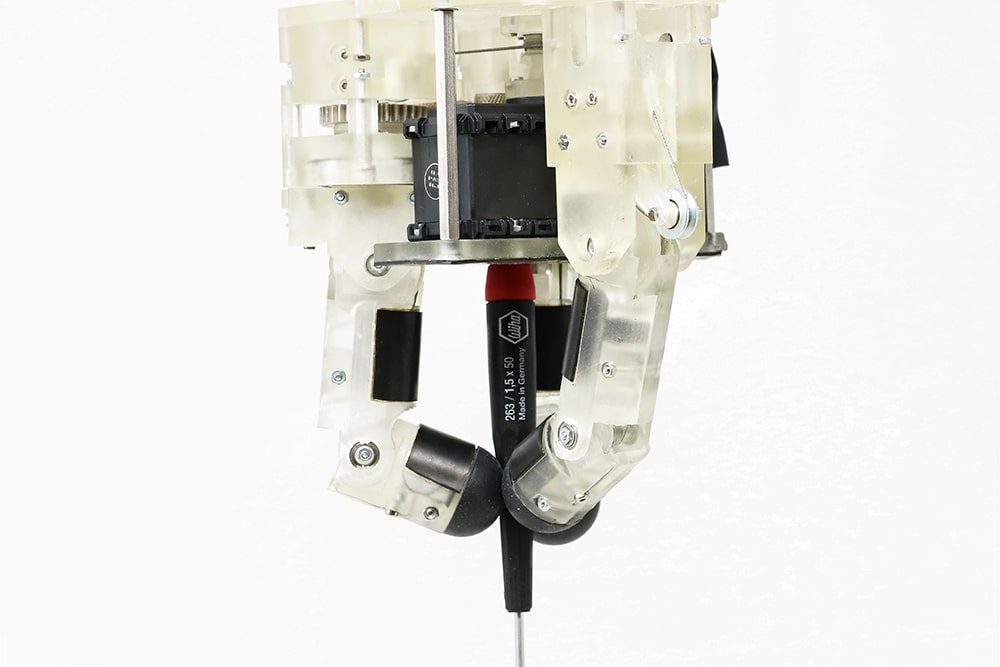}  & \hspace{-10mm} \includegraphics[width=0.39\linewidth]{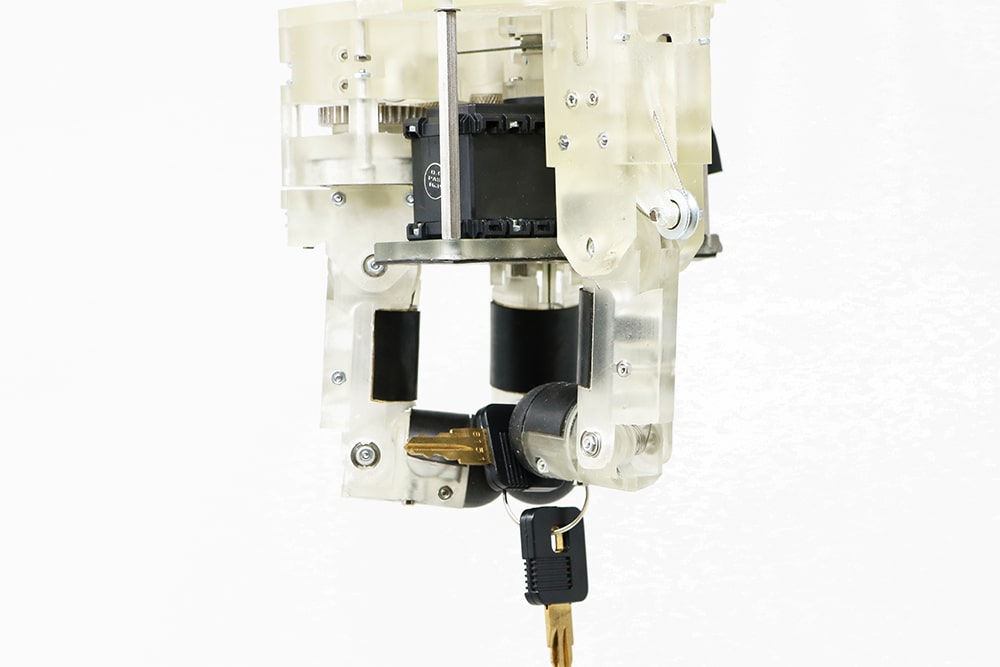}  & \hspace{-10mm} \includegraphics[width=0.39\linewidth]{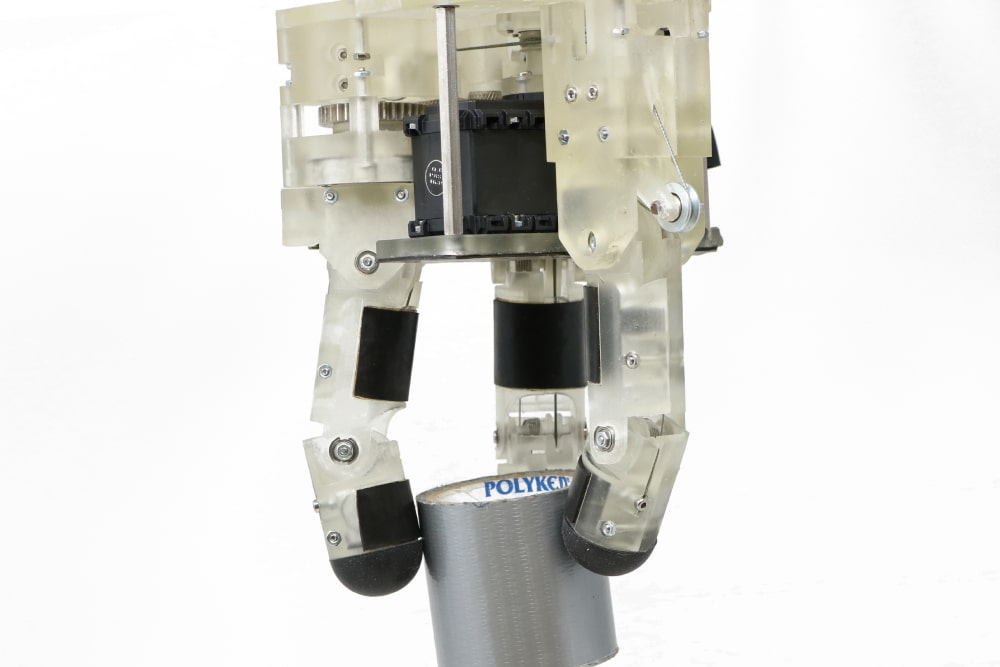}\\
\hspace{-5mm}
(j) Small screw driver & \hspace{-10mm}
(k) Key & \hspace{-10mm}
(l) Tape  \vspace{2mm}\\
\end{tabular}
\caption{Grasp examples using prototype hands. (a) -- (f): Design Case I, (g) -- (l): Design Case II.}
\label{fig:grasp_photos}
\end{figure}

In Design Case I, all joints are actuated by a single motor. Fig. \ref{fig:finger_traj} (and the accompanying multimedia attachment) demonstrates the finger trajectories, in which the hand first closes toward the center, making a spherical grasp posture and then a pinch grasp posture. Continuing to close, the fingertips do not collide but rather pass each other (due to motion in the roll degree of freedom). Finally, the hand creates an enveloping grasp. Fig. \ref{fig:grasp_photos} (a) - (f) show several grasps using this hand, displaying the versatility of the hand. We can see the hand can perform stable pinch grasps (d)(e), spherical grasps (c)(f), and power grasps (a)(b). 

In Design Case II, the tendons of three fingers are connected to the pulley of the main motor, and the roll joints are actuated by a smaller motor via gear transmission (which shares the same mathematical expression as the antagonistic tendon transmission in Fig. \ref{fig:hand_and_actuation2} (b)). Therefore, the closing and rolling motion are controlled separately. Shown in Fig. \ref{fig:finger_traj_dual_motor}, the hand can close the fingers towards the center to pinch small objects, or close the fingers in a parallel fashion so the finger can pass each other and make an enveloping grasp. Fig. \ref{fig:grasp_photos} (g) - (l) show some example grasps of the resulted hand of Design Case II. \remindtext{1-13}{We note that this design can perform pinch grasps of very small objects as shown in (j) and (k).} We also highlight the fact that Fig. \ref{fig:grasp_photos} also includes objects outside the desired grasp pool (e.g.(c)(h)(k)).

As for Design Case III, since the optimization result indicates that the performance is worse than the other two designs shown here, we did not build a physical prototype.

\section{Discussion}
\label{sec:discussion}
The results for all three design cases demonstrate that our method is effective: the proposed optimization framework can indeed shape the Mechanically Realizable Torque and Postural Manifolds to fit the desired grasps.

\addedtext{3-7}{Fig.\ref{fig:mrtm_comparison} and Fig.\ref{fig:mrpm_comparison} show that the optimized Mechanically Realizable Manifolds closely interpolate the data points defined by the desired grasps in both posture and torque spaces (except for the finger manifolds of Design Case III). In other words, the hands with corresponding underactuation designs can create stable grasps that are close to the desired ones in terms of both posture shaping and force generation.} \addedtext{1-2-2 \& A1-3}{In each plot in Fig.\ref{fig:mrtm_comparison}, the distance between the manifold and the desired grasp points (not considering excluded grasps) is either exactly or close to zero, indicative of similarly low unbalanced joint torques.}

The comparison with PCA results in Fig.\ref{fig:mrtm_comparison} and Fig.\ref{fig:mrpm_comparison} also illustrates the performance of our method. Assuming linear manifolds, the PCA-based manifolds explain the most variance (measured by Euclidean distance) of the desired grasps. However, due to the constraints, those theoretically optimal manifolds cannot be reached exactly. In contrast, the proposed method calculates manifolds which are close to the PCA results, by minimizing the aforementioned equilibrium-based metrics (instead of explicitly minimizing the Euclidean distances), and also considering physical constraints. In addition, the proposed method can also find nonlinear manifolds that make use of mechanism characteristics. For example, Design Case III shows the Mechanically Realizable Manifolds that are not limited to the linear domain.

In Table \ref{table:metrics}, the numeric values of objective functions, as well as the number of grasps excluded due to being beyond the resulted hand's capability, provide evaluations of a design. Even though the problem of initial kinematic design is out of the scope of our work (we require a pre-specified kinematic configuration), one can use these metrics calculated in our method to compare different kinematic designs, which provides insights for the choice of kinematics:

\begin{itemize}
\item Comparing the roll-pitch single motor (Design Case I) with dual-motor design (Design Case II), the results match our intuition: With the same kinematics, the dual-motor design \addedtext{1-5-1}{(having a two-dimensional Mechanically Realizable Posture Manifold as shown in Fig.\ref{fig:mrpm_comparison} (e)) can better span the joint posture space, i.e. has more versatility in terms of the postures (smaller numbers in column 2 and 3 in Table \ref{table:metrics}). For example, the corresponding hand can pick smaller objects as shown in Fig. \ref{fig:grasp_photos} (j) and (k)}. Also, Design Case II displays better force generation capability in order to create stability for different grasps \addedtext{1-5-1}{(smaller number in column 1 in Table \ref{table:metrics}, also displayed in Fig. \ref{fig:mrtm_comparison} (e))}, due to the decoupling between roll joints and the other joints. 

\item Comparing the roll-pitch configurations (Design Case I and II) with the pitch-yaw configuration (Design Case III), the pitch-yaw design is worse \addedtext{1-5-1}{(larger numbers in all columns in Table \ref{table:metrics})}, especially for the intra-tendon optimization. From this result, we can conclude that such a combination of pitch-yaw proximal joints with the certain tendon routing scheme in Fig. \ref{fig:hand_and_actuation3} has very limited capability, \addedtext{1-5-1}{as the parameterization of this manifold does not have enough capacity to express most of the desired grasps under the given mechanical constraints.}

\end{itemize}

The dual-layer optimization framework combining the non-convex stochastic global search in the outer-layer and the convex optimization in the inner-layer is a useful formulation. \modifiedtext{1-5}{Except for specific kinematic designs such as the floating pulley transmission assumed in \cite{ciocarlie2011constrained}}, it is generally not possible to formulate the parameter selection problem as a globally convex problem. In contrast, the dual-layer framework is capable of dealing with various kinematic configurations and actuation methods.

\addedtext{1-5-3}{Our framework explicitly optimizes underactuation designs for a given set of desired grasps. This is useful for cases where the range of objects a hand is expected to interact with is largely known in advance (as is the case for the use case of a free-flying robot on the ISS). However, our experiments show that the resulting hands can also create stable grasps outside the set that they are explicitly optimized for, as illustrated in Fig. \ref{fig:grasp_photos} (c)(h)(k). This suggests that, rather than ``overfitting'' to the desired grasps, the optimized manifolds fit between them in a useful fashion, extracting an inherent structure in posture and torque spaces that is useful for grasping common objects beyond the ones explicitly optimized for. However, we make no attempt to quantify this effect here.}

\addedtext{1-5-2}{There are still limitations of this work. The first limitation stems from the need to manually pre-specify hand kinematics and tendon connectivity patterns, which provides the parameterization needed to initialize our framework for a specific design. For example, in the three design cases discussed, the parameterization of the actuation matrix $\bm{A}$ and motor-tendon connection matrix $\bm{M}$ are set manually, which makes it impossible to fully automate the framework. In the future, if a unified parameterization of the kinematic design can be defined, or if a list of possible designs can be exhaustively enumerated, then such full design automation and optimization can be achieved. Second, our framework also starts from a list of manually defined desired grasps. The grasps we use here all have force closure, but are not optimized with respect to any other grasp quality metric (e.g.  Ferrari-Canny metric). Future work can investigate combining grasp planning and hand design in a larger-scale nested optimization addressing these limitations.} 

\section{Conclusion}
In this paper, we proposed the concept of Mechanically Realizable Manifolds, both in joint torque and joint angle space. We presented a data-driven approach that optimizes the Mechanically Realizable Manifolds to fit a set of desired grasp data points, in terms of posture shaping and force generation. We conducted three design cases using this theory, compared the results quantitatively among the cases, constructed prototypes of two cases that showed better capabilities and verified their performance in practice.

We believe that the synergistic behavior is part of the ``intelligence'' of robotic hands, and the idea of combining hardware-embedded and software-based intelligence is a promising path towards versatile grasping. Our method of Mechanically Realizable Manifolds is a step on this path, building a bridge between the dexterous hand behaviors and the underactuated hardware design. For future work, in addition to addressing the aforementioned limitations of this paper, we aim to progressively develop more general approaches to modeling both hardware-embedded hand behavior and software-based control strategy, which could then be used for deriving co-optimization frameworks. Ultimately, we aim to design robotic hands with an informed distribution of intelligence over hardware and software, resulting in more dexterity and less complexity.

\appendix[Derivation of the Matrices in Design Case III]

\begin{figure}[t!]
\centering
\includegraphics[width=\linewidth]{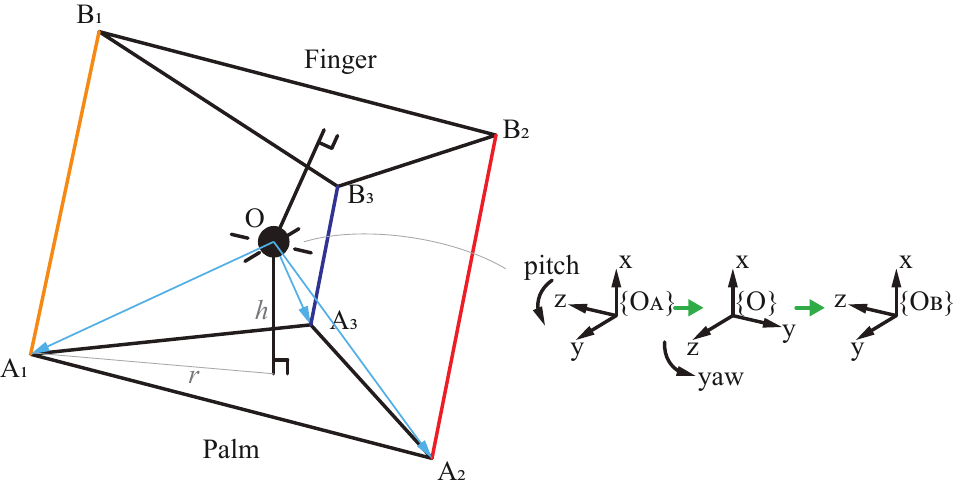}
\caption{The 2-DoF universal proximal joint in Design Case III, as well as the attached coordinate frames. The origin of all frames are located in \textit{O}, and frame ${O_A}$, ${O_B}$ are bonded to the lower and upper platform respectively.}
\label{fig:appendix_u_joint}
\end{figure}

First, the inverse kinematics (i.e. calculating the tendon length from angles) for two-DoF proximal joint need to be solved. Fig. \ref{fig:appendix_u_joint} shows the two-DoF joint and the attached coordinate frames. The tendon lengths can be solved using coordinate transforms. For example, the length of the tendon connecting $A_1$ and $B_1$ can be calculated as:
\vspace{-1mm}
\begin{equation} \label{eq:appendix_ik_l1}
\lVert  ^{O_A}{\vv{B_1 A_1}} \rVert = \lVert ^{O_A}{\vv{O A_1}} - ^{O_A}{R_O}  ^{O}{R_{O_B}}  ^{O_B}{\vv{O B_1}} \rVert
\end{equation}
where the prescripts are the coordinates the vector is described in, and the rotation matrix $^{O_A} R_O$ represents the transform from coordinate \{$O_A$\} to \{$O$\}. 

Next, we show the derivation of the Actuation Matrix, which relates the tendon forces and the net joint torques. Here we use the torque of pitch DoF $\tau_{p}$ as an example:
\begin{equation} \label{eq:appendix_trq}
\scalebox{.95}[1.0]{
$
\begin{gathered}
\scriptsize
\tau_{p} =   \begin{bmatrix}0 & 0 & 1\end{bmatrix} \cdot  \\
\normalsize
\begin{smallmatrix}
( ^{O_A}\vv{O A_1} \times \frac{^{O_A}\vv{B_1 A_1}}{\lVert ^{O_A}\vv{B_1 A_1} \rVert} t_1 \hspace{-1mm}
&+^{O_A}\vv{O A_2} \times \frac{^{O_A}\vv{B_2 A_2}}{\lVert ^{O_A}\vv{B_2 A_2} \rVert} t_2 \hspace{-1mm}
&+^{O_A}\vv{O A_3} \times \frac{^{O_A}\vv{B_3 A_3}}{\lVert ^{O_A}\vv{B_3 A_3} \rVert} t_3) \\
\end{smallmatrix}
\vspace{1mm}
\end{gathered}
$
}
\end{equation}

where $t_1$, $t_2$, $t_3$ are the magnitude of tendon forces. 

We denote:
\begin{equation} \label{eq:appendix_rho}
\begin{gathered}
\scriptsize
^{O_A}\hspace{-0.5mm}\rho =   \begin{bmatrix}0 & 0 & 1\end{bmatrix} \cdot  \\
\normalsize
\left[ \begin{smallmatrix} ^{O_A}\vv{O A_1} \times \frac{^{O_A}\vv{B_1 A_1}}{\lVert ^{O_A}\vv{B_1 A_1} \rVert} 
 & ^{O_A}\vv{O A_2} \times \frac{^{O_A}\vv{B_2 A_2}}{\lVert ^{O_A}\vv{B_2 A_2} \rVert} 
 &  ^{O_A}\vv{O A_3} \times \frac{^{O_A}\vv{B_3 A_3}}{\lVert ^{O_A}\vv{B_3 A_3} \rVert} \end{smallmatrix} \right]
\vspace{1mm}
\end{gathered}
\end{equation} 
and then:
\begin{equation} \label{eq:appendix_trq_expressed_by_rho}
\tau_{p} = ^{O_A}\hspace{-1mm}\rho \begin{bmatrix} t_1 \\ t_2 \\ t_3 \end{bmatrix} = ^{O_A}\hspace{-1mm}\rho_1 t_1 + ^{O_A}\hspace{-1mm}\rho_2 t_2 + ^{O_A}\hspace{-1mm}\rho_3 t_3
\end{equation} 

Following the requirement that the post-contact movement is negligible, and the reasoning shown in (\ref{eq:eq_before_touch}) (\ref{eq:eq_after_touch}) and (\ref{eq:net_trq}), the net joint torque can be expressed as:

\begin{equation} \label{eq:appendix_net_trq}
\tau_{p, net} =  ^{O_A}\hspace{-1mm}\rho_1 t_{1, net} + ^{O_A}\hspace{-1mm}\rho_2 t_{2, net}
\end{equation} 

The moment arms $^{O_A}\hspace{-0.5mm}\rho_1$ and $^{O_A}\hspace{-0.5mm}\rho_2$ in (\ref{eq:appendix_net_trq}) are the entries $\rho_{fpp11}$ and $\rho_{fpp12}$ of the Actuation Matrix $A$ shown in (\ref{eq:actuation_matrix3}). We can construct the entire Actuation Matrix in a similar way.

Finally, we give some details in the tendon shortening vector $\bm{s}$ in (\ref{eq:tendon_travel_vector3}). The $\Delta l$’s are the tendon length changes between zero-configuration and grasp configuration. For example:

\begin{equation} \label{eq:appendix_delta_l}
\Delta l_{fp11} = l_{1(zero)} - l_{1(grasp)}
\end{equation} 
where $l_{1(grasp)}$ and $l_{1(zero)}$ can be solved by the aforementioned joint inverse kinematics using the pitch and yaw angles.

\section*{Acknowledgment}
This work was supported in part by the NASA Early Stage Innovations (ESI) program through award NNX16AD13G and the Office of Naval Research (ONR) under grant N00014-16-1-2026.

The authors would like to thank Jinzhao Chang, Abhijeet Mishra and Tianyi Zhang for their contributions to the hardware design.


\bibliographystyle{bib/IEEEtran}
\bibliography{bib/sensing,bib/design,bib/control,bib/analysis,bib/planning,bib/simulation,bib/learning}

\begin{thebibliography}{10}
\providecommand{\url}[1]{#1}
\csname url@rmstyle\endcsname
\providecommand{\newblock}{\relax}
\providecommand{\bibinfo}[2]{#2}
\providecommand\BIBentrySTDinterwordspacing{\spaceskip=0pt\relax}
\providecommand\BIBentryALTinterwordstretchfactor{4}
\providecommand\BIBentryALTinterwordspacing{\spaceskip=\fontdimen2\font plus
\BIBentryALTinterwordstretchfactor\fontdimen3\font minus
  \fontdimen4\font\relax}
\providecommand\BIBforeignlanguage[2]{{%
\expandafter\ifx\csname l@#1\endcsname\relax
\typeout{** WARNING: IEEEtran.bst: No hyphenation pattern has been}%
\typeout{** loaded for the language `#1'. Using the pattern for}%
\typeout{** the default language instead.}%
\else
\language=\csname l@#1\endcsname
\fi
#2}}

\bibitem{santello1998postural}
M.~Santello, M.~Flanders, and J.~F. Soechting, ``Postural hand synergies for
  tool use,'' \emph{Journal of Neuroscience}, vol.~18, no.~23, pp.
  10\,105--10\,115, 1998.

\bibitem{rosell2011autonomous}
J.~Rosell, R.~Su{\'a}rez, C.~Rosales, and A.~P{\'e}rez, ``Autonomous motion
  planning of a hand-arm robotic system based on captured human-like hand
  postures,'' \emph{Autonomous Robots}, vol.~31, no.~1, p.~87, 2011.

\bibitem{ciocarlie2009hand}
M.~T. Ciocarlie and P.~K. Allen, ``Hand posture subspaces for dexterous robotic
  grasping,'' \emph{The International Journal of Robotics Research}, vol.~28,
  no.~7, pp. 851--867, 2009.

\bibitem{wimbock2011synergy}
T.~Wimb{\"o}ck, B.~Jahn, and G.~Hirzinger, ``Synergy level impedance control
  for multifingered hands,'' in \emph{IEEE/RSJ Intl. Conf. on Intelligent
  Robots and Systems}, 2011, pp. 973--979.

\bibitem{gioioso2013mapping}
G.~Gioioso, G.~Salvietti, M.~Malvezzi, and D.~Prattichizzo, ``Mapping synergies
  from human to robotic hands with dissimilar kinematics: an approach in the
  object domain,'' \emph{IEEE Transactions on Robotics}, vol.~29, no.~4, pp.
  825--837, 2013.

\bibitem{meeker2018intuitive}
C.~Meeker, T.~Rasmussen, and M.~Ciocarlie, ``Intuitive hand teleoperation by
  novice operators using a continuous teleoperation subspace,'' in \emph{IEEE
  Intl. Conf. on Robotics and Automation}, 2018.

\bibitem{matrone2010principal}
G.~C. Matrone, C.~Cipriani, E.~L. Secco, G.~Magenes, and M.~C. Carrozza,
  ``Principal components analysis based control of a multi-dof underactuated
  prosthetic hand,'' \emph{Journal of neuroengineering and rehabilitation},
  vol.~7, no.~1, p.~16, 2010.

\bibitem{matrone2012real}
G.~C. Matrone, C.~Cipriani, M.~C. Carrozza, and G.~Magenes, ``Real-time
  myoelectric control of a multi-fingered hand prosthesis using principal
  components analysis,'' \emph{Journal of neuroengineering and rehabilitation},
  vol.~9, no.~1, p.~40, 2012.

\bibitem{tsoli2010robot}
A.~Tsoli and O.~C. Jenkins, ``Robot grasping for prosthetic applications,'' in
  \emph{The International Journal of Robotics Research}, 2010, pp. 1--12.

\bibitem{brown2007inter}
C.~Y. Brown and H.~H. Asada, ``Inter-finger coordination and postural synergies
  in robot hands via mechanical implementation of principal components
  analysis,'' in \emph{IEEE/RSJ Intl. Conf. on Intelligent Robots and Systems},
  2007, pp. 2877--2882.

\bibitem{xu2014design}
K.~Xu, H.~Liu, Y.~Du, and X.~Zhu, ``Design of an underactuated anthropomorphic
  hand with mechanically implemented postural synergies,'' \emph{Advanced
  Robotics}, vol.~28, no.~21, pp. 1459--1474, 2014.

\bibitem{xu2019composed}
K.~Xu, Z.~Liu, B.~Zhao, H.~Liu, and X.~Zhu, ``Composed continuum mechanism for
  compliant mechanical postural synergy: An anthropomorphic hand design
  example,'' \emph{Mechanism and Machine Theory}, vol. 132, pp. 108--122, 2019.

\bibitem{li2014design}
S.~Li, X.~Sheng, H.~Liu, and X.~Zhu, ``Design of a myoelectric prosthetic hand
  implementing postural synergy mechanically,'' \emph{Industrial Robot},
  vol.~41, no.~5, pp. 447--455, 2014.

\bibitem{chen2015mechanical}
W.~Chen, C.~Xiong, and S.~Yue, ``Mechanical implementation of kinematic synergy
  for continual grasping generation of anthropomorphic hand,'' \emph{IEEE/ASME
  Transactions on Mechatronics}, vol.~20, no.~3, pp. 1249--1263, 2015.

\bibitem{xiong2016design}
C.~Xiong, W.~Chen, B.~Sun, M.~Liu, S.~Yue, and W.~Chen, ``Design and
  implementation of an anthropomorphic hand for replicating human grasping
  functions,'' \emph{IEEE Transactions on Robotics}, vol.~32, no.~3, pp.
  652--671, 2016.

\bibitem{gabiccini2011role}
M.~Gabiccini, A.~Bicchi, D.~Prattichizzo, and M.~Malvezzi, ``On the role of
  hand synergies in the optimal choice of grasping forces,'' \emph{Autonomous
  Robots}, vol.~31, no. 2-3, p. 235, 2011.

\bibitem{prattichizzo2013motion}
D.~Prattichizzo, M.~Malvezzi, M.~Gabiccini, and A.~Bicchi, ``On motion and
  force controllability of precision grasps with hands actuated by soft
  synergies,'' \emph{IEEE Transactions on Robotics}, vol.~29, no.~6, pp.
  1440--1456, 2013.

\bibitem{grioli2012adaptive}
G.~Grioli, M.~Catalano, E.~Silvestro, S.~Tono, and A.~Bicchi, ``Adaptive
  synergies: an approach to the design of under-actuated robotic hands,'' in
  \emph{IEEE/RSJ Intl. Conf. on Intelligent Robots and Systems}, 2012, pp.
  1251--1256.

\bibitem{catalano2014adaptive}
M.~G. Catalano, G.~Grioli, E.~Farnioli, A.~Serio, C.~Piazza, and A.~Bicchi,
  ``Adaptive synergies for the design and control of the pisa/iit softhand,''
  \emph{The International Journal of Robotics Research}, vol.~33, no.~5, pp.
  768--782, 2014.

\bibitem{birglen2007underactuated}
L.~Birglen, T.~Lalibert{\'e}, and C.~M. Gosselin, \emph{Underactuated robotic
  hands}.\hskip 1em plus 0.5em minus 0.4em\relax Springer, 2007, vol.~40.

\bibitem{dollar2011joint}
A.~M. Dollar and R.~D. Howe, ``Joint coupling design of underactuated hands for
  unstructured environments,'' \emph{The International Journal of Robotics
  Research}, vol.~30, no.~9, pp. 1157--1169, 2011.

\bibitem{ciocarlie2011constrained}
M.~Ciocarlie and P.~Allen, ``A constrained optimization framework for compliant
  underactuated grasping,'' \emph{Mechanical Sciences}, vol.~2, no.~1, pp.
  17--26, 2011.

\bibitem{saliba2016quasi}
M.~Saliba and C.~De~Silva, ``Quasi-dynamic analysis, design optimization, and
  evaluation of a two-finger underactuated hand,'' \emph{Mechatronics},
  vol.~33, pp. 93--107, 2016.

\bibitem{dong2018geometric}
H.~Dong, E.~Asadi, C.~Qiu, J.~Dai, and I.-M. Chen, ``Geometric design
  optimization of an under-actuated tendon-driven robotic gripper,''
  \emph{Robotics and Computer-Integrated Manufacturing}, vol.~50, pp. 80--89,
  2018.

\bibitem{ciocarlie2014velo}
M.~Ciocarlie, F.~M. Hicks, R.~Holmberg, J.~Hawke, M.~Schlicht, J.~Gee,
  S.~Stanford, and R.~Bahadur, ``The velo gripper: A versatile single-actuator
  design for enveloping, parallel and fingertip grasps,'' \emph{The
  International Journal of Robotics Research}, vol.~33, no.~5, pp. 753--767,
  2014.

\bibitem{chen2018underactuated}
T.~Chen, M.~Haas-Heger, and M.~Ciocarlie, ``Underactuated hand design using
  mechanically realizable manifolds,'' in \emph{IEEE Intl. Conf. on Robotics
  and Automation}, 2018, pp. 7392--7398.

\bibitem{prattichizzo2008grasping}
D.~Prattichizzo and J.~C. Trinkle, ``Grasping,'' in \emph{Springer handbook of
  robotics}, 2008, pp. 671--700.

\bibitem{hansen2001completely}
N.~Hansen and A.~Ostermeier, ``Completely derandomized self-adaptation in
  evolution strategies,'' \emph{Evolutionary computation}, vol.~9, no.~2, pp.
  159--195, 2001.

\bibitem{hansen2003reducing}
N.~Hansen, S.~D. M{\"u}ller, and P.~Koumoutsakos, ``Reducing the time
  complexity of the derandomized evolution strategy with covariance matrix
  adaptation (cma-es),'' \emph{Evolutionary computation}, vol.~11, no.~1, pp.
  1--18, 2003.

\bibitem{bualat2015astrobee}
M.~Bualat, J.~Barlow, T.~Fong, C.~Provencher, T.~Smith, and A.~Zuniga,
  ``Astrobee: Developing a free-flying robot for the international space
  station,'' in \emph{AIAA SPACE 2014 Conference and Exposition}, vol. 4643,
  2015.

\bibitem{miller2004graspit}
A.~T. Miller and P.~K. Allen, ``Graspit! a versatile simulator for robotic
  grasping,'' \emph{IEEE Robotics \& Automation Magazine}, vol.~11, no.~4, pp.
  110--122, 2004.

\bibitem{ferrari1992planning}
C.~Ferrari and J.~Canny, ``Planning optimal grasps,'' in \emph{IEEE Intl. Conf.
  on Robotics and Automation (ICRA)}, 1992, pp. 2290--2295.

\end{thebibliography}
\vspace{-20mm}

\begin{IEEEbiography}[{\includegraphics[width=1in,height=1.25in,clip, keepaspectratio]{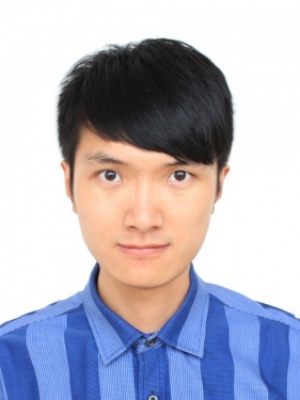}}]{Tianjian Chen}
received his B.E. degree in Mechanical Engineering from Huazhong University of Science and Technology, Wuhan, China, in 2013, and his M.S. degree in Mechanical Engineering from Carnegie Mellon University, Pittsburgh, PA, USA, in 2015. He is currently a Ph.D. candidate in the Robotic Manipulation and Mobility Lab at the Department of Mechanical Engineering, Columbia University, New York, NY, USA.

His research interests include the design, control and machine learning of robotic hands.
\end{IEEEbiography}

\vspace{-20mm}

\begin{IEEEbiography}[{\includegraphics[width=1in,height=1.25in, clip, keepaspectratio]{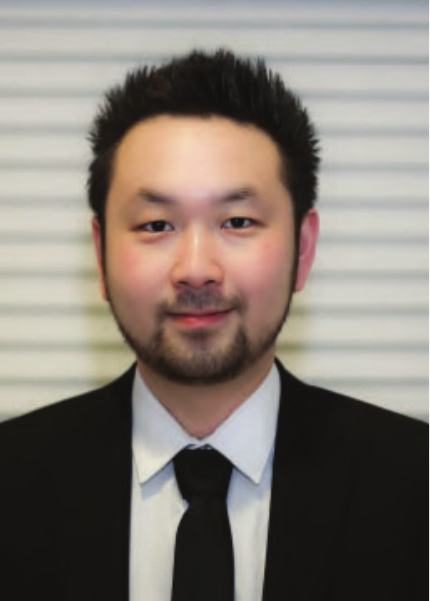}}]{Long Wang}
received his B.S. degree in Mechanical Engineering from Tsinghua University, Beijing, China, in 2010, and he received his M.S. degree in Mechanical Engineering from Columbia University, New York, NY, in 2012. He received his Ph.D. degree in Mechanical Engineering from Vanderbilt University, Nashville, TN, in 2019. He is currently a Postdoctoral Researcher at Columbia University Robotic Manipulation and Mobility Lab. 

His research interests include modeling and calibration of continuum robots, surgical robotics, telemanipulation, manipulation, and grasping.
\end{IEEEbiography}

\vspace{-20mm}

\begin{IEEEbiography}[{\includegraphics[width=1in,height=1.25in,clip,keepaspectratio]{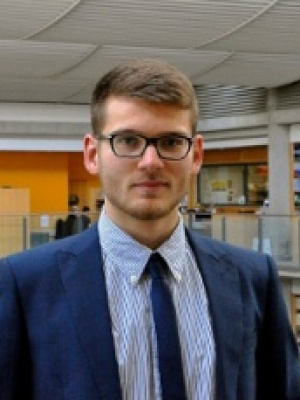}}]{Maximilan Haas-Heger}
received the MEng degree in Aeronautical Engineering from Imperial College London in 2015. Since 2015, he is a Ph.D. candidate in the Robotic Manipulation and Mobility Lab at Columbia University in New York. 

His research focuses on the theoretical foundations of robotic grasping, specifically applied to the development of accurate grasp models.
\end{IEEEbiography}

\vspace{-20mm}

\begin{IEEEbiography}[{\includegraphics[width=1in,height=1.25in,clip,keepaspectratio]{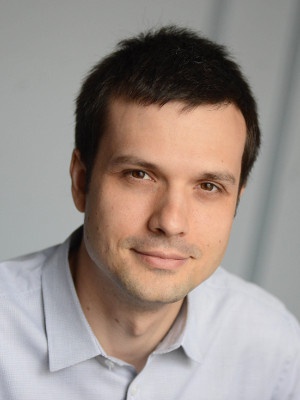}}]{Matei Ciocarlie}
is an Associate Professor in the Mechanical Engineering Department at Columbia University, with affiliated appointments in Computer Science and the Data Science Institute. He completed his Ph.D. at Columbia University, New York. He was awarded the Early Career Award by the IEEE Robotics and Automation Society, a Young Investigator Award by the Office of Naval Research, a CAREER Award by the National Science Foundation, and a Sloan Research Fellowship by the Alfred P. Sloan Foundation.

His research interest focuses on robot motor control, mechanism and sensor design, planning and learning, all aiming to demonstrate complex motor skills such as dexterous manipulation.
\end{IEEEbiography}

\end{document}